\documentclass{article}

\usepackage{microtype}
\usepackage{graphicx}
\usepackage{subcaption}
\usepackage{booktabs} % for professional tables

\usepackage{hyperref}
\PassOptionsToPackage{inline,shortlabels}{enumitem}
\usepackage{enumitem}

\usepackage[accepted]{icml2026}

\usepackage{amsmath}
\usepackage{amssymb}
\usepackage{mathtools}
\usepackage{amsthm}
\graphicspath{{figures/}}
\usepackage{algorithm}
\makeatletter
\@ifundefined{algorithmic}{}{%

}
\makeatother
\usepackage{multirow}
\usepackage{algpseudocode}
\algrenewcommand\alglinenumber[1]{} 
\usepackage[capitalize,noabbrev]{cleveref}
\usepackage{xcolor}
\definecolor{projectblue}{HTML}{1F5FAF}
\usepackage{url}
\usepackage{colortbl}  % 
\usepackage{tikz}
\usepackage[normalem]{ulem}
\usetikzlibrary{calc}
\newcommand{\second}[1]{\cellcolor{blue!6}#1}
\newcommand{\best}[1]{\cellcolor{blue!12}\textbf{#1}}
\newlength{\figpad}
\theoremstyle{plain}
\newtheorem{theorem}{Theorem}[section]
\newtheorem{proposition}[theorem]{Proposition}
\newtheorem{lemma}[theorem]{Lemma}
\newtheorem{corollary}[theorem]{Corollary}
\theoremstyle{definition}

\theoremstyle{remark}

\usepackage[disable, textsize=tiny]{todonotes}

\icmltitlerunning{RAMAC: Multimodal Risk-Aware Offline Reinforcement Learning}
\begin{document}

\twocolumn[
  \icmltitle{RAMAC: Multimodal Risk-Aware Offline Reinforcement Learning \protect\\ and the Role of Behavior Regularization}



  \icmlsetsymbol{equal}{*}

  \begin{icmlauthorlist}
    \icmlauthor{Kai Fukazawa}{sch1}
    \icmlauthor{Kunal Mundada}{sch1}
    \icmlauthor{Iman Soltani}{sch1}
  \end{icmlauthorlist}

  \icmlaffiliation{sch1}{University of California, Davis}
  % \icmlaffiliation{sch2}{Department of Computer Science, University of California, Davis, CA, USA}
\icmlcorrespondingauthor{Kai Fukazawa}{kfukazawa@ucdavis.edu}
  \icmlkeywords{Machine Learning, ICML}

  \vskip 0.3in
]



\printAffiliationsAndNotice{}  % no special notice (required even if empty)

\begin{abstract}
In safety-critical domains where online data collection is infeasible, offline reinforcement learning (RL) is attractive only if policies achieve high returns without catastrophic lower-tail risk. Prior work on risk-averse offline RL achieves safety at the cost of either (i) value/model-based pessimism or (ii) restricted policy classes that limit expressiveness, whereas diffusion/flow-based expressive generative policies have largely been used in risk-neutral settings.
We introduce \textbf{Risk-Aware Multimodal Actor-Critic (RAMAC)}, a simple, modular, model-free framework that couples an expressive generative actor (e.g., diffusion/flow) with a distributional critic and optimizes a composite objective that combines Conditional Value-at-Risk (CVaR) with behavioral cloning (BC), enabling risk-sensitive learning in complex multimodal scenarios. Since out-of-distribution (OOD) actions are a major driver of catastrophic failures in offline RL, we further provide an objective-level analysis showing that controlling behavior divergence via BC suppresses OOD actions and stabilizes CVaR. Instantiating RAMAC with a diffusion actor, we illustrate these insights on a 2-D risky bandit and evaluate on Stochastic-D4RL, observing consistent gains in $\mathrm{CVaR}_{0.1}$ while maintaining strong returns. The code and experimental results are available on the
\href{https://kaifukazawa.github.io/ramac-project/}
{\textcolor{projectblue}{project website}}.
\end{abstract}
\begin{figure}[t]
\centering
 \includegraphics[
  width=1.0\linewidth,
  trim={0.5cm 0.5cm 0.5cm 0.35cm},
  clip
]{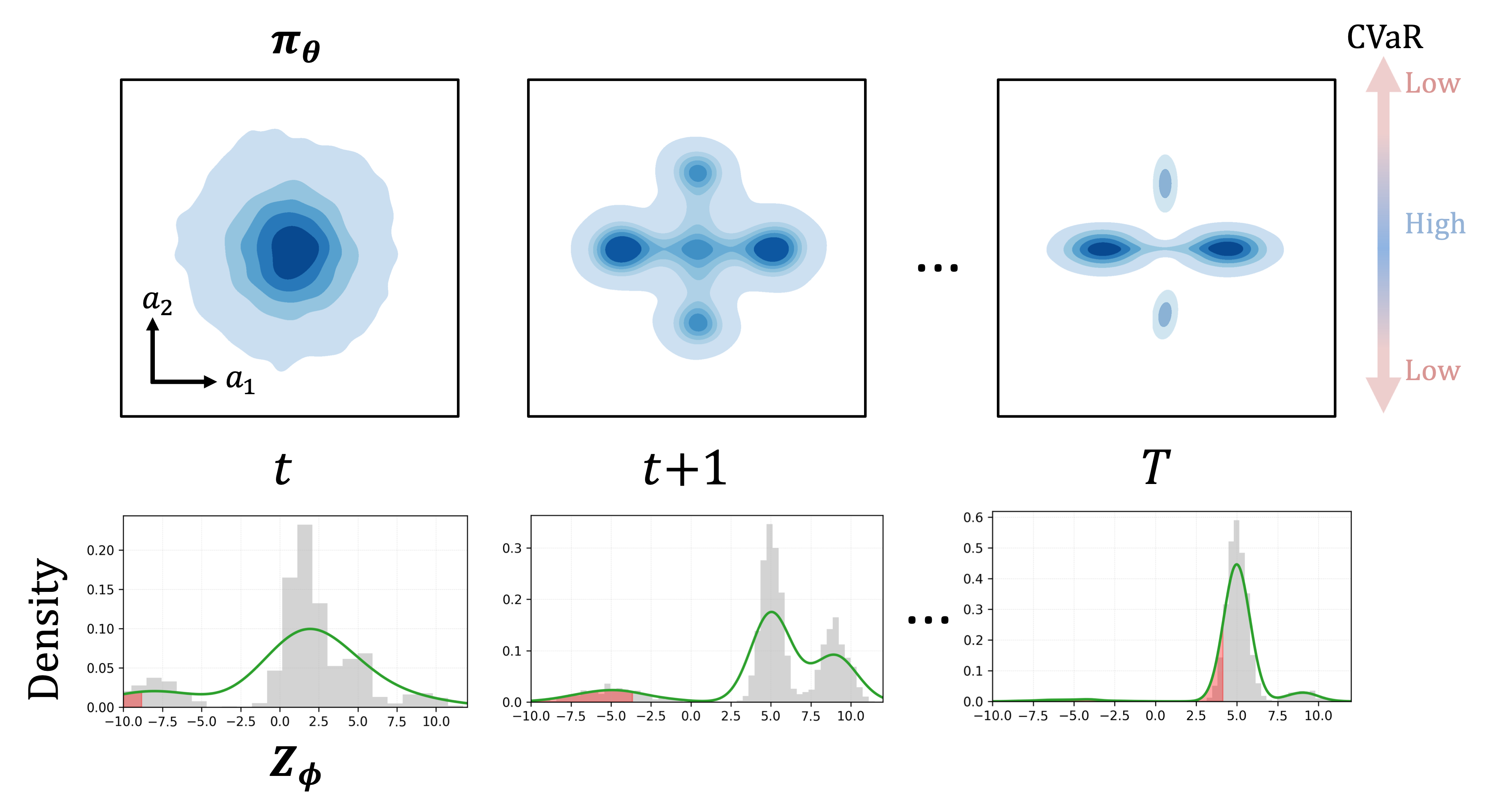}
  %\vspace{-0.4em}
%   \caption{
% \textbf{RAMAC learning dynamics (conceptual).}
% \emph{Top:} the policy density $\pi_\theta(a\!\mid\!s)$ induced by a reparameterized generative actor evolves during training.
% \emph{Bottom:} the critic models the return distribution $Z_\phi(s,a,\tau)$; CVaR focuses on lower tail highlighted (\textcolor[HTML]{FF6666}{red}). 
% The composite objective shifts probability mass away from low-tail regions while BC anchors the policy to dataset-supported modes, preserving multimodal high-reward behavior.
% }
\caption{
\textbf{RAMAC learning dynamics (conceptual).}
\emph{Top:} The generative policy $\pi_\theta(a\!\mid\!s)$ evolves during training; the desired outcome is to retain multiple dataset-supported high-CVaR modes while suppressing low-CVaR modes, without collapsing or moving off support.
\emph{Bottom:} The critic $Z_\phi(s,a,\tau)$ models the return distribution, whose lower tail, highlighted in \textcolor[HTML]{FF6666}{red}, defines CVaR. CVaR guides mode reweighting, while BC preserves data support.
}
  \label{fig:ramac_evolutions}
\end{figure}
\section{Introduction}
In high-stakes applications such as autonomous driving, robotics, finance, and healthcare, where real-world exploration can lead to catastrophic consequences, offline RL offers a safe approach for generating policies that not only maximize long-horizon returns but also \emph{tightly control risk} \citep{levine2020offline}. Recent expressive generative policies \citep{koirala2025flow,park2025flow,wang2022diffusion} can capture multimodal behavior and thus excel in achieving high expected return, yet their primary use has been limited to \emph{risk-neutral} settings. Conversely, existing risk-averse algorithms ensure safety by enforcing conservatism or restricted policy classes, which constrain policy expressiveness~\citep{kumar2020conservative,ma2021conservative,urpi2021risk}. This paper asks: \emph{Can we obtain safety without sacrificing expressiveness?}

We answer in the affirmative by proposing the \textbf{Risk-Aware Multimodal Actor-Critic (RAMAC)}, a simple and modular framework that couples an expressive generative actor with a distributional critic and optimizes a \emph{single composite objective} that combines BC regularization with distributional risk (instantiated with CVaR).
As illustrated conceptually in Fig.~\ref{fig:ramac_evolutions}, the CVaR term steers probability mass away from low-tail regions while BC keeps the generative actor anchored to dataset modes, enabling tail-risk control without sacrificing multimodal expressiveness.
This unifies high expressiveness with robust tail-risk control while directly constraining the generative policy to the data support, addressing two central safety concerns in offline RL: \emph{catastrophic tail outcomes} and \emph{out-of-distribution (OOD) actions}.

Prior offline-RL approaches can be organized by mechanism:

\textbf{(1) Policy regularization} constrains the policy to the data manifold via divergence minimization or policy priors, improving stability but often sacrificing policy expressiveness on complex tasks with risk-neutral examples such as \citep{fujimoto2021minimalist,fujimoto2019off,kumar2019stabilizing,wu2019behavior} and risk-aware methods with \emph{prior-anchored perturbation} designs such as \citep{chen2025diffusion,urpi2021risk}.

\textbf{(2) Value conservatism} reduces optimistic extrapolation, but can underestimate the value of infrequent yet high-return in-distribution modes due to global pessimism and data imbalance in both risk-neutral \citep{kumar2020conservative} and risk-aware instances \citep{ma2021conservative}.

\textbf{(3) Model-based pessimism} bounds transition uncertainty with ensembles and penalties, at the cost of compounding model errors at scale again under both risk-neutral \citep{rigter2022rambo,yu2020mopo,yu2021combo} and risk-aware \citep{rigter2023one} settings.

\textbf{(4) Expressive generative policies} faithfully clone multimodal behavior and achieve state-of-the-art mean returns, but have so far been used primarily in \emph{risk-neutral} applications \citep{hansen2023idql,kang2023efficient,koirala2025flow,park2025flow,wang2022diffusion} including closely related concurrent works that pair diffusion with distributional critics~\citep{liu2025distributional,zhang2025d2}.

Despite compelling results from these expressive generative policies in risk-neutral RL, their potential in offline risk-aware RL remains largely untapped. Among behavior-regularized risk-averse methods, the most prominent approaches that leverage expressive priors rely on \emph{prior-anchored perturbation}~\citep{chen2025diffusion,urpi2021risk}, which trade away much of the multimodal capacity and, as our analysis shows, can still incur OOD actions.

Here, we aim to leverage the advantages of expressive policies with tail-risk control without adding extra learned components (e.g., dynamics models or ensembles) or separate OOD pipelines (e.g., additional risk-averse policy training). To this end, inspired by the success of recent risk-neutral expressive policies such as~\citep{wang2022diffusion,park2025flow},
RAMAC couples an expressive generative actor with a distributional critic and \emph{differentiates a single composite objective (BC + CVaR) through the generative process}~\citep{chow2015risk,di2012policy}. Beyond providing a simple and modular recipe for risk-aware expressive policies, we give an objective-level analysis that links behavior divergence control to safety: BC regularization suppresses per-state OOD action probabilities and, more importantly, stabilizes CVaR of bounded return-related scores, explaining why this \emph{single composite objective} mitigates lower-tail blow-ups for expressive generative actors (Sec.~\ref{sec:risk_pitfalls}).
We also empirically show that RAMAC yields high expected return while minimizing risk on complex multimodal offline benchmarks. Our contributions can be summarized as:

1) Risk-aware expressive policy learning:
         We introduce \textbf{RAMAC}, a simple, modular, and model-free framework that enables risk-aware learning with expressive generative actors by optimizing a \emph{single composite objective} that combines BC regularization with distributional risk optimization. Our primary instantiation is a diffusion-based actor, \textbf{RADAC (Risk-Aware Diffusion Actor-Critic)}.

2) Objective-level and geometric analysis: We give simple bounds linking BC-induced behavior divergence control to (i) per-state OOD action events and (ii) the stability of $\mathrm{CVaR}_\alpha$ for bounded utility scores. We also provide a geometric perspective showing that prior-anchored perturbation can still allocate probability mass off-support even with expressive priors. 

3) Experimental evaluation:
        On Stochastic-D4RL benchmarks,
        \textbf{RADAC} outperforms baselines on CVaR while maintaining competitive mean return on most tasks. We additionally report a flow-matching variant, \textbf{RAFMAC (Risk-Aware Flow-Matching Actor-Critic)}, in App.~\ref{app:sd4rl_full}, which shows similar trends. Finally, we (i) visualize the geometric OOD mechanism in a 2-D risky bandit that contrasts \emph{prior-anchored perturbation} and \emph{expressive generative policies}, and (ii) use OOD-action detectors on Stochastic-D4RL to empirically validate the analysis.
\section{Preliminaries}
\label{sec:preliminaries}
%======================================================================
\paragraph{Offline RL.}
We consider a finite-horizon Markov Decision Process (MDP) $\mathcal{M} = (\mathcal{S}, \mathcal{A}, P, r, \gamma, H)$ with state space $\mathcal{S}$, action space $\mathcal{A}$, transition kernel 
$P(\cdot \mid s,a)$, reward function $r(s,a)$, discount factor $\gamma \in (0,1)$, and horizon $H \in \mathbb{N}$~\citep{sutton1998reinforcement}.
In offline RL, the learner is given only a static dataset
$\mathcal{D} = \{(s_i,a_i,r_i,s'_i)\}_{i=1}^N$ collected by some
unknown behavior policy $\beta$, and cannot further interact with the environment~\citep{prudencio2023survey}. Let $\operatorname{supp}(\mathcal D)$ denote the empirical support of the dataset. The objective is to learn a policy $\pi$ that maximizes the expected return $J(\pi) = \mathbb{E}_{\pi, P}[\sum_{t=0}^{H-1} \gamma^t r_t]$ without extra environment interaction. The central challenge is \emph{distributional shift} (OOD): When $\pi$ visits $(s,a)\notin \operatorname{supp}(\mathcal D)$,
value
estimates $Q(s,a)$ extrapolate and can become arbitrarily inaccurate~\citep{kumar2020conservative}.
Policies that place non-negligible mass on such OOD actions may therefore suffer catastrophic failures at deployment. Prior work alleviates this issue with behavior regularization,
conservative critics, or model-based pessimism.
\paragraph{Behavior–Regularized Actor–Critic (BRAC).} A large family of offline methods uses an actor–critic with an explicit proximity term to the behavior policy~\citep{fujimoto2021minimalist,kumar2019stabilizing, nair2020awac,wu2019behavior}. A representative actor–critic objective takes the form:
% \vspace{-0.3em}
% \begin{equation}
%   \mathcal L_{\text{Actor}}(\theta)
%   = \mathbb E_{
%       \substack{
%         s\sim\mathcal D,\\
%         a\sim\pi_\theta(\cdot\mid s)
%       }
%     }
%      \bigl[-\,Q_\phi(s,a)
%   - \alpha\log \pi_\theta(a\mid s)\bigr],
%   \label{eq:brac_actor_loss}
% \end{equation}
\begin{equation}
  \mathcal L_{\text{Actor}}(\theta)
  = \mathbb E_{
      \substack{
        (s,a)\sim\mathcal D,\\
        a^\pi\sim\pi_\theta(\cdot\mid s)
      }
    }
     \bigl[-\,Q_\phi(s,a^\pi)
  - \alpha\log \pi_\theta(a\mid s)\bigr],
  \label{eq:brac_actor_loss}
\end{equation}
% \vspace{-1.0em}
\begin{equation}
  \mathcal L_{\text{Critic}}(\phi)
  = \mathbb E_{
      \substack{
        (s,a,r,s')\sim\mathcal D,\\
        a'\sim\pi_\theta(\cdot\mid s')
      }
    }
    \bigl[Q_\phi(s,a) - r - \gamma Q_{\bar\phi}(s',a')\bigr]^{2}.
  \label{eq:brac_critic_loss}
\end{equation}

% Here the second term $-\alpha \log \pi_\theta(a\mid s)$ plays the role of a
% \emph{behavior regularizer}: it is typically instantiated as a behavioral-cloning (BC) term or other proximity/divergence penalties that keep $\pi_\theta(\cdot\mid s)$ close
% to the behavior policy $\beta(\cdot\mid s)$.
Here the second term $-\alpha \log \pi_\theta(a\mid s)$,
evaluated on dataset actions $a\sim\mathcal D$, plays the role of a
\emph{behavior regularizer}: it is typically instantiated as a
behavioral-cloning (BC) term or another proximity/divergence penalty
that keeps $\pi_\theta(\cdot\mid s)$ close to the behavior policy
$\beta(\cdot\mid s)$.
% ~\footnote{For diffusion/flow actors, $\log \pi_\theta(a\mid s)$ may be intractable; we therefore implement behavior regularization via BC-style surrogates that anchor the generative process to dataset actions (e.g., score-matching for diffusion; NLL for flows when available).} 
\footnote{Following~\citet{park2025flow}, we use the term
\emph{behavior-regularized actor--critic} broadly for offline
actor--critic methods that combine value improvement with behavior
regularization, rather than only for the original BRAC algorithm.
For expressive generative actors, the behavior-fidelity term is
implemented using the actor's native BC surrogate: denoising loss for
RADAC and flow-matching loss for RAFMAC.
}
Empirically, BRAC-style objectives have turned out to be surprisingly strong in offline RL ~\citep{tarasov2023revisiting}.
In this work, we extend this behavior-regularized pattern to a \emph{distributional} actor–critic
in which the critic is expanded into a \emph{distributional critic} that models the return distribution, enabling optimization of tail-sensitive objectives such as CVaR in place of the mean $Q$.

\paragraph{Distributional RL and Risk Measures.} Standard actor–critic methods including BRAC optimize the expected return by learning the mean action-value function $Q^\pi(s,a)=\mathbb{E}[Z^\pi(s,a)]$ as shown in Eq.~\ref{eq:brac_critic_loss}. Distributional RL instead models the entire \emph{return distribution} $Z^\pi(s,a)$~\citep{bellemare2017distributional}.
The distributional Bellman operator is:
\begin{equation}
  (\mathcal T^\pi Z)(s,a) \;\stackrel{d}{=}\; r(s,a) + \gamma\, Z(s',a').
    \label{eq:bellman_eq}
\end{equation}
where $s' \sim P(\cdot|s,a)$ and $a' \sim \pi(\cdot|s')$. A common parameterization uses an Implicit Quantile Network (IQN)~\citep{dabney2018implicit}
to approximate the inverse cumulative distribution function (CDF)
$Z_\phi(s,a;\tau) \approx F^{-1}_{Z^\pi(s,a)}(\tau)$ for quantile levels
$\tau \in (0,1)$.
% Access to quantiles enables coherent risk measures $\mathcal{D}(\cdot)$ that emphasize
% different parts of the return distribution.
% In this work we focus on the Conditional Value-at-Risk (CVaR), a widely used instantiation of $\mathcal{D}(\cdot)$
% as a risk-averse objective. 
Access to quantiles enables distortion-based risk objectives that emphasize
different parts of the return distribution.
Risk sensitivity in RL can be formulated in multiple ways; in this work, we focus
specifically on Conditional Value-at-Risk
(CVaR) as our primary instantiation because it directly targets catastrophic
lower-tail outcomes.
For a risk level $\alpha \in (0,1]$, the CVaR admits the integral form used for actor gradients:
\begin{equation}
  \mathrm{CVaR}_\alpha(X)=\frac{1}{\alpha}\int_0^\alpha F_X^{-1}(\tau)\,d\tau.
  \label{eq:cvar_integral}
\end{equation}

In our analysis, we will also later consider CVaR of bounded scalar cost/risk scores, for which simple stability bounds are available. Optimizing $\mathrm{CVaR}_\alpha$ encourages policies that trade some mean performance for
improved behavior in the worst $\alpha$-fraction of trajectories, which is crucial in
safety-critical settings.
In RAMAC, the distributional critic provides quantile estimates from which CVaR and its
gradients with respect to actions can be computed and backpropagated through the policy. 
\begin{figure*}[t]
    \centering
    \includegraphics[width=0.7\textwidth,trim={0cm 4.5cm 1.5cm 6cm}, clip]{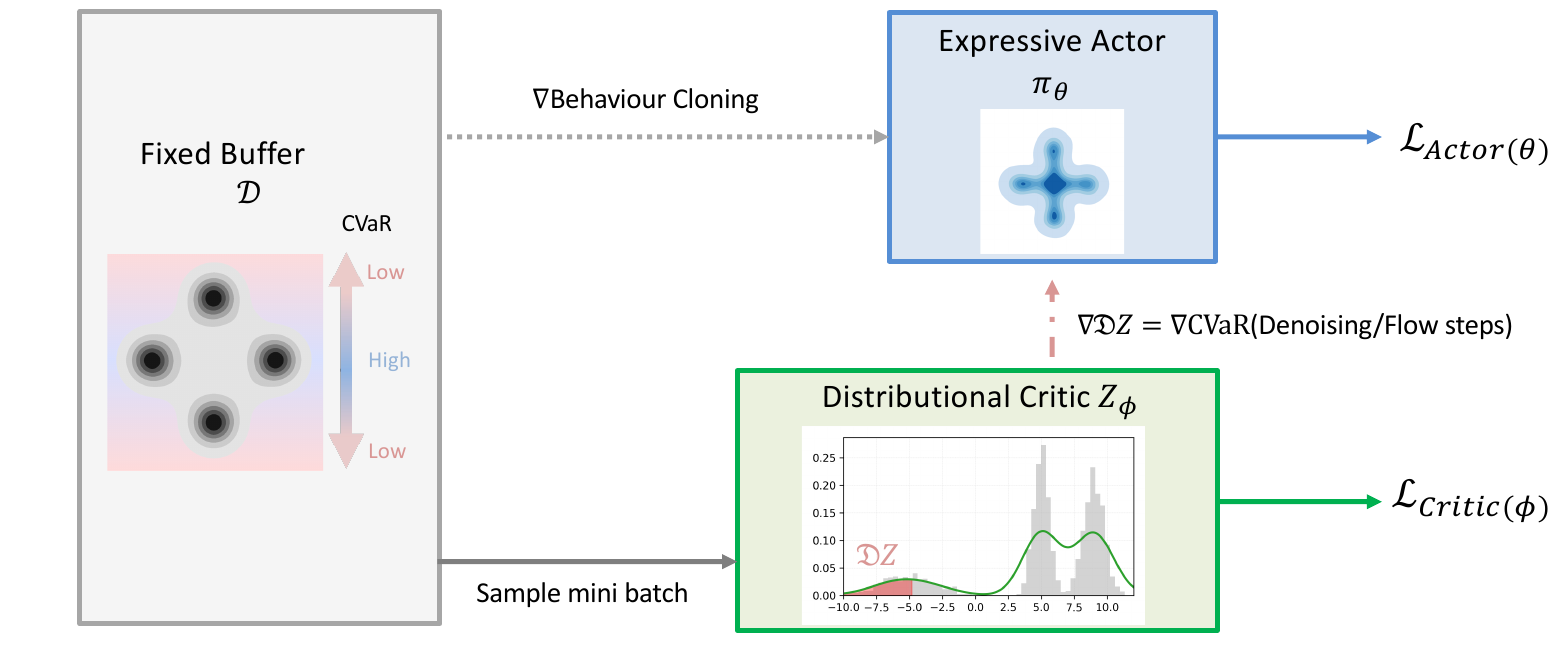}
    \caption{\textbf{RAMAC pipeline.}
From the offline buffer $\mathcal{D}$ \textcolor{gray!50!black}{(gray)}, the distributional critic $Z_\phi$ \textcolor{green!60!black}{(green)} fits the return law with a quantile loss and aggregates its lower tail into a CVaR signal.
That signal is differentiated through the generative path of the actor $\pi_\theta$ \textcolor{blue!70!black}{(blue; diffusion or flow)}, which is trained with the composite objective $\mathcal L_{\pi}=\mathcal L_{\mathrm{BC}}+\eta\,\mathcal L_{\mathrm{Risk}}$ to shift mass away from low-quantile regions while staying on-manifold, yielding stable lower-tail performance.}
    \label{fig:ramac_overview}
\end{figure*}
\paragraph{Expressive Generative Policies as Differentiable Trajectories.}
Recent offline RL methods stay within the behavior-regularized actor-critic
template of Eqs.~\ref{eq:brac_actor_loss} and \ref{eq:brac_critic_loss}, but replace
the simple parametric actor with an expressive conditional generative model ~\citep{hansen2023idql,kang2023efficient,koirala2025flow, park2025flow, wang2022diffusion}.
Given a state $s$ and latent $z\!\sim\!\mathcal N(0,I)$, the policy generates an
action $a = \psi_\theta(s,z)$ along a \emph{differentiable path}
~\citep{hansen2023idql,wang2022diffusion}, while an explicit behavior term keeps
$\psi_\theta$ close to the dataset actions. We focus on the two families:

(i) \emph{Diffusion policies} model a reverse-time stochastic differential equation (SDE) over actions~\citep{song2020score},
\begin{equation}
  \mathrm d\mathbf a_t = f_\theta(t,\mathbf a_t,s)\,\mathrm dt + g(t)\,\mathrm d\mathbf w_t,
  \label{eq:sde}
\end{equation}
where a forward noising process gradually corrupts dataset actions into near-Gaussian
noise, and the network $f_\theta$ learns to reverse this process conditioned on the state
$s$.

(ii) \emph{Flow-matching policies} solve a deterministic ODE~\citep{lipman2022flow},
\begin{equation}
  \frac{\mathrm d\mathbf a_t}{\mathrm dt} = v_\theta(t,\mathbf a_t,s).
  \label{eq:ode}
\end{equation}
where a neural vector field $v_\theta$ transports samples from a simple base distribution
to the data distribution along a continuous trajectory.
Integrating Eq.~\ref{eq:ode} from an initial noise sample yields an action conditioned
on $s$.

In both cases, the overall map $\psi_\theta : (s,z)\mapsto a$ is fully differentiable, and the behavior term encourages $\psi_\theta$ to approximate the behavior action
distribution itself; the critic then fine-tunes this expressive behavior model using
scalar signals.
Prior work typically uses expected-value or advantage-based signals from a mean-value critic to update the generative
policy under a BRAC template (as in Eq.~\ref{eq:brac_actor_loss}),
yielding a \emph{risk-neutral} generative actor-critic method. In contrast, RAMAC replaces the mean critic with a \emph{distributional} critic and uses tail-sensitive risk signals such as $\mathrm{CVaR}_\alpha(Z^\pi(s,a))$ (Eq.~\ref{eq:cvar_integral}) to shape the same generative policy under a \emph{single composite objective} that combines BC regularization and distributional risk; the
exact loss is introduced in Sec.~\ref{sec:method}.
\section{Method}
\label{sec:method}
We now introduce the \textbf{Risk-Aware Multimodal Actor-Critic (RAMAC)}. RAMAC is a simple, modular, model-free framework that couples (i) a \emph{distributional critic} learning the conditional return law and (ii) an \emph{expressive generative actor} (diffusion or flow) trained by a \emph{single composite objective}. As summarized in Fig.~\ref{fig:ramac_overview}, the critic provides a CVaR signal, whose gradients are backpropagated through the actor’s differentiable generative path, while a BC term regularizes the policy to the dataset support. Intuitively, CVaR updates steer probability mass away from low-quantile, catastrophic regions, whereas BC preserves multimodal high-reward behavior by constraining the behavior divergence. This BRAC-style regularizer not only mitigates OOD actions but also stabilizes tail objectives; our objective-level bounds in Sec.~\ref{sec:risk_pitfalls} formalize these effects.
%----------------------------------------------------------------------
\subsection{Distributional Critic}
\label{sec:critic}
Risk-sensitive objectives such as CVaR require access to the entire return distribution.
We therefore adopt a distributional critic $Z_\phi$ via IQN~\citep{dabney2018implicit}, building on the distributional Bellman operator
in Eq.~\ref{eq:bellman_eq}. Here $\delta_\phi$ is the \emph{quantile TD error} that matches the predicted quantile $Z_\phi(s,a;\tau)$ to the distributional Bellman target $r+\gamma Z_{\bar\phi}(s',a';\tau')$ with independently sampled quantile levels $\tau,\tau'$. We minimize a distributional Bellman residual:
{
\begin{align}
\delta_\phi
&:= r+\gamma Z_{\bar\phi}(s',a';\tau')-Z_\phi(s,a;\tau), \notag\\
\mathcal{L}_{\text{Critic}}(\phi)
&= \mathbb{E}_{
  \substack{
    (s,a,r,s')\sim\mathcal D,\\
    a'\sim\pi_\theta(\cdot|s'),\ \tau,\tau'\sim \mathcal U(0,1)
  }
}
\Big[ \mathcal L_\kappa\!\big(\delta_\phi\,;\,\tau\big)\Big].
\label{eq:critic_compact}
\end{align}
}
where $\mathcal L_\kappa(\delta;\tau)$ denotes the IQN quantile-Huber loss (with $\kappa\!=\!1$)~\citep{dabney2018implicit}; we provide its explicit form and implementation details in App.~\ref{app:impl_ramac}. This yields lower-tail quantiles that will directly drive the risk-aware actor update in Sec.~\ref{sec:risk_aware_actor}.
%----------------------------------------------------------------------
\subsection{Risk-Aware Generative Actor}
\label{sec:risk_aware_actor}
Given a state $s$ and latent noise $z\sim\mathcal N(0,I)$, the
generative actor produces an action $a=\psi_\theta(s,z)$.
We define CVaR at level $\alpha$ through the critic’s quantiles and use a Monte Carlo estimator:
{
% \setlength{\abovedisplayskip}{4pt}
%  \setlength{\belowdisplayskip}{4pt}
%  \setlength{\abovedisplayshortskip}{4pt}
%  \setlength{\belowdisplayshortskip}{4pt}
% An action is sampled as:
% {
% \setlength{\abovedisplayskip}{4pt}
%  \setlength{\belowdisplayskip}{4pt}
%  \setlength{\abovedisplayshortskip}{4pt}
%  \setlength{\belowdisplayshortskip}{4pt}
% \begin{equation}
% \label{eq:reparam_action}
% a \;=\; \psi_\theta(s,z), 
% \qquad z \sim \mathcal N(0,I).
% \end{equation}
% }
\begin{equation}
\label{eq:cvar_def_keep}
\begin{split}
\mathrm{CVaR}_\alpha\!\bigl(Z_\phi(s,a)\bigr)
&=\frac{1}{\alpha}\int_0^\alpha Z_\phi(s,a;\tau)\,d\tau \\
&\approx\; \frac{1}{K}\sum_{k=1}^{K} Z_\phi\bigl(s,a;\tau_k\bigr),
\end{split}
\end{equation}
}
where $\tau_k \sim \mathcal U(0,\alpha)$ for $k=1,\dots,K$.
The risk loss maximizes this quantity. This is equivalent to minimizing the negative CVaR~\footnote{This specific loss, instantiated with CVaR, is what we refer to as $\mathcal{L}_{\text{CVaR}}$ in our architectural diagrams for clarity.}:
{
\begin{equation}
\label{eq:risk_loss_keep}
\mathcal{L}_{\mathrm{Risk}}(\theta)
= -\,\mathbb{E}_{s\sim\mathcal D,\; a\sim \pi_\theta(\cdot\mid s)}
\bigl[\mathrm{CVaR}_\alpha\!\big(Z_\phi(s,a)\big)\bigr].
\end{equation}
}
\subsection{Behavior-Regularized Objective}
\label{subsec:behavior_regularized_loss}
The complete policy objective balances risk aversion with fidelity to the offline dataset. We instantiate fidelity through a standard behavior cloning (BC) term that encourages the policy to reproduce the behavior distribution. We define:
{
\begin{equation}
  \label{eq:bc_loss}
  \mathcal{L}_{\text{BC}}(\theta)
  = -\,\mathbb{E}_{(s,a)\sim \mathcal D}\big[\log \pi_\theta(a\mid s)\big],
\end{equation}
}
which corresponds to the BRAC-style behavior regularizer in Eq.~\ref{eq:brac_actor_loss}  up to a scaling of the coefficient.
It combines the risk term with a standard behavior cloning (BC) loss, $\mathcal{L}_{\text{BC}}(\theta)$:
{
\begin{equation}
  \mathcal{L}_{\pi}(\theta) = \underbrace{\mathcal{L}_{\text{BC}}(\theta)}_{\text{data fidelity}} + \eta\, \underbrace{\mathcal{L}_{\text{Risk}}(\theta)}_{\text{tail-risk aversion}}.
  \label{eq:policy_loss_full}
\end{equation}
}
where $\eta$ is a hyperparameter. Our primary instantiation is a diffusion policy (\textbf{RADAC}), while an additional flow-matching variant (\textbf{RAFMAC}) is reported in App.~\ref{app:sd4rl_full}. We show pseudocode for RAMAC in Algorithm~\ref{alg:ramac} and  describe the full implementation details in App.~\ref{app:impl_ramac}
\begin{algorithm}[t]
\caption{\textbf{Risk-Aware Multimodal Actor-Critic (RAMAC)}}
\label{alg:ramac}
\begin{algorithmic} 
\Statex \textbf{Initialize} policy network $\pi_{\theta}$, critic $Z_{\phi}$, target critic $Z_{\bar\phi}$;  mini-batch size $B$, risk level $\alpha$, critic-tail samples $K$, Exponential Moving Average (EMA) rate $\rho$. 
\vspace{0.2em}
\Repeat
  \State Sample a mini-batch $\{(s,a,r,s')\}_{b=1}^B \sim \mathcal D$.
  \State \textbf{Training Critic:} 
    \State Sample $z'\!\sim\!\mathcal N(0,I)$ and set $a'=\psi_\theta(s',z')$;
    % \textit{(Eq.~\ref{eq:reparam_action})}; 
    \State Sample $\tau,\tau'\!\sim\!\mathcal U(0,1)$
    \State Update $\phi$ by minimizing $\mathcal L_{\text{Critic}}(\phi)$ \textit{(Eq.~\ref{eq:critic_compact})}.
  \State \textbf{Training Actor:} 
    \State
    Sample $z\!\sim\!\mathcal N(0,I)$ and set $a=\psi_\theta(s,z)$;
    % \textit{(Eq.~\ref{eq:reparam_action})}; 
    \State
    Sample $\tau_1,\dots,\tau_K\!\sim\!\mathcal U(0,\alpha)$
    \State Update $\theta$ by minimizing $\mathcal L_{\pi}(\theta)$ \textit{(Eq.~\ref{eq:policy_loss_full})}.
  \State \textbf{Target update:} $\bar\phi \leftarrow \rho\,\bar\phi + (1-\rho)\,\phi$.
\Until{converged}
\end{algorithmic}
\end{algorithm}

\section{Behavior Regularization in Offline RL}
\label{sec:risk_pitfalls}

\noindent
Prior work has demonstrated the importance of behavior regularization in offline RL due to its ability to constrain the learned policy to the data manifold and curb value extrapolation~\citep{tarasov2023revisiting}. A commonly adopted regularization scheme in offline risk-aware RL is the prior-anchored perturbation method (e.g., ORAAC, UDAC)\footnote{For simplicity and consistency with our experiments, we will refer to UDAC as \emph{ORAAC–Diffusion}}~\citep{chen2025diffusion,urpi2021risk}, which uses a linear mixing of actions from a pretrained BC policy and from the RL actor (a bounded residual). Here, we first discuss a limitation of this regularization approach. We then highlight the advantages of our scheme, namely, the behavior-regularized objective in Eq.~\ref{eq:policy_loss_full}.

\subsection{Prior-Anchored Perturbation and Its Limitations} 
In this approach, policy output can be written as:
{
\begin{equation}
\label{eq:perturbation}
    a \;=\; b \;+\; \zeta_\psi(s,b),\ \ \|\zeta_\psi(s,b)\|\le \Phi,
\end{equation}
}
where $\ b\sim G_\phi(\cdot\mid s)$, $\zeta_\psi$ is a \emph{learned residual} (optimized to increase $Q$ or CVaR) and the norm bound $\Phi$ \emph{keeps updates close to the anchor}.
Define the anchor support $\mathcal S_G(s)$ (the region in action space where $G_\phi(\cdot\mid s)$ places mass), the full action space $\mathbb R^d$, and the $\Phi$-radius ball of $b$
{
\begin{equation}
   B_\Phi(b)\;=\;\{\,a\in\mathbb R^d:\ \|a-b\|_2\le \Phi\,\}. 
\end{equation}
}
 where any perturbed action $a=b+\zeta_\psi$ with $\|\zeta_\psi\|\le\Phi$ lies in $B_\Phi(b)$. 
Hence on-manifold deployment is guaranteed by the \emph{safety margin} condition
{
\begin{equation}
\label{eq:margin_condition}
\begin{aligned}
\mathrm{dist}\!\bigl(b,\ \mathbb R^d\!\setminus\!\mathcal S_G(s)\bigr) \;>\; \Phi
\Longrightarrow\ 
\begin{cases}
B_\Phi(b)\subseteq \mathcal S_G(s),\\
\forall\,\|\zeta_\psi\|\le \Phi:\ a\in\mathcal S_G(s).
\end{cases}
\end{aligned}
\end{equation}
}
where $\mathrm{dist}(x,A)\!\coloneqq\!\inf_{y\in A}\|x-y\|_2$ denotes Euclidean distance. 
OOD can still occur when this margin fails. This method provides a convenient \emph{local} improvement rule; however, prior work has observed that it suffers from \emph{poor mode coverage} in multimodal action spaces~\citep{wang2022diffusion}. In addition to the identified limitations, we show \emph{a distinct geometric weakness} that can occur even without multimodality; having multiple modes merely magnifies the effect.

% \begin{lemma}
% \label{lem:anchored_leakage}
% Fix $s$ and write $I_s = \mathcal S_G(s)$ and
% $O_s = \mathbb R^d \setminus I_s$.
% Suppose there exist an anchor $b^\star \in I_s$ and a radius
% $\Phi > 0$ such that
% $\lambda\!\big(B_\Phi(b^\star)\cap O_s\big) > 0$, and the policy
% $\pi_{\text{anch}}(\cdot\mid s)$ induced by Eq.~\ref{eq:perturbation}
% admits a density $p(\cdot\mid s)$ with
% \[
% p(a\mid s) \;\ge\; c > 0
% \quad\text{for all } a\in B_\Phi(b^\star).
% \]
% Then its per-state OOD probability
% \[
% \delta_s(\pi_{\text{anch}})
% \;\coloneqq\; \pi_{\text{anch}}(O_s\mid s)
% \]
% satisfies
% \[
% \delta_s(\pi_{\text{anch}})
% \;\ge\; c\cdot\lambda\bigl(B_\Phi(b^\star)\cap O_s\bigr)\;>\;0.
% \]
% In particular, as long as the density on $B_\Phi(b^\star)$ remains
% bounded below by $c>0$, further training of the residual cannot drive
% $\delta_s(\pi_{\text{anch}})$ to zero. Proof appears in App.~\ref{app:lemma1_proof}.
% \end{lemma}
\begin{lemma}
\label{lem:anchored_leakage}
Fix $s$ and write $I_s = \mathcal S_G(s)$ and
$O_s = \mathbb R^d \setminus I_s$.
Suppose there exist an anchor $b^\star \in I_s$, a radius
$\Phi > 0$, and a measurable leakage region
$A_s \subseteq B_\Phi(b^\star)\cap O_s$ such that
$\lambda(A_s)>0$.
If the policy $\pi_{\mathrm{anch}}(\cdot\mid s)$ induced by
Eq.~\ref{eq:perturbation} admits a density $p(\cdot\mid s)$ satisfying
\[
p(a\mid s) \;\ge\; c > 0
\quad\text{for all } a\in A_s,
\]
then its per-state OOD probability
\[
\delta_s(\pi_{\mathrm{anch}})
\;\coloneqq\; \pi_{\mathrm{anch}}(O_s\mid s)
\]
satisfies
\[
\delta_s(\pi_{\mathrm{anch}})
\;\ge\; c\cdot\lambda(A_s)\;>\;0.
\]
In particular, as long as the induced density remains bounded below by
$c>0$ on $A_s$ during residual training, further optimization of the
residual cannot drive $\delta_s(\pi_{\mathrm{anch}})$ to zero. Proof appears in App.~\ref{app:lemma1_proof}.
\end{lemma}
% \noindent
% Lemma~\ref{lem:anchored_leakage} captures how the geometry of the anchor
% support forces a strictly positive OOD probability once the anchor ball
% overlaps the complement $O_s$. Thin or nonconvex supports make such
% low-margin anchors $b^\star$ unavoidable, and gradients that are not
% constrained to lie tangent to the data manifold inevitably push some
% mass across the boundary (see App.~\ref{app:lemma_geometric} for a more
% detailed discussion).

\noindent
Lemma~\ref{lem:anchored_leakage} is a conditional mechanism statement: whenever an
anchor-centered perturbation policy assigns nontrivial probability density to
an off-support portion of its feasible perturbation region, residual training
alone cannot eliminate the associated OOD probability while that local mass
persists. Thin or nonconvex behavior supports can make such overlap more
likely, since bounded residual updates are not explicitly constrained to remain
on the data manifold (see App.~\ref{app:lemma_geometric} for a more
detailed discussion).
\begin{figure*}[t]
  \centering
  \begin{minipage}{0.8\textwidth} 
    \centering

    % --- row 1 (4 plots) ---
    \begin{subfigure}[b]{0.22\linewidth}
      \caption{Ground Truth}
      \fbox{\includegraphics[width=\linewidth]{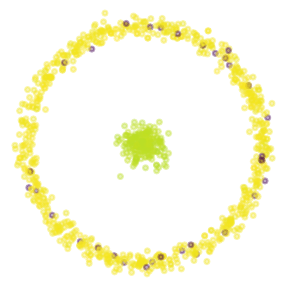}}
    \end{subfigure}\hfill
    \begin{subfigure}[b]{0.22\linewidth}
      \caption{CVAE-QL}
      \fbox{\includegraphics[width=\linewidth]{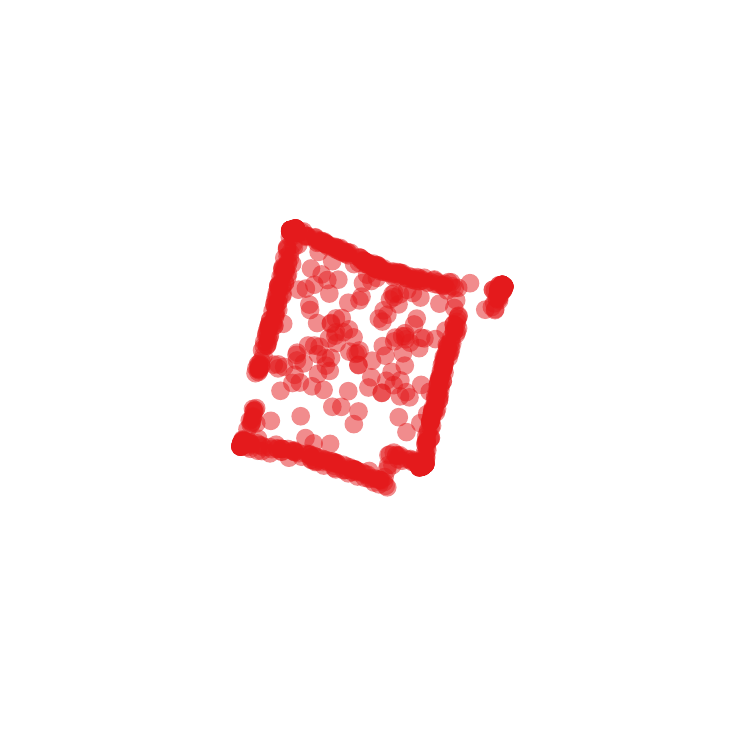}}
    \end{subfigure}\hfill
    \begin{subfigure}[b]{0.22\linewidth}
      \caption{DiffusionQL}
      \fbox{\includegraphics[width=\linewidth]{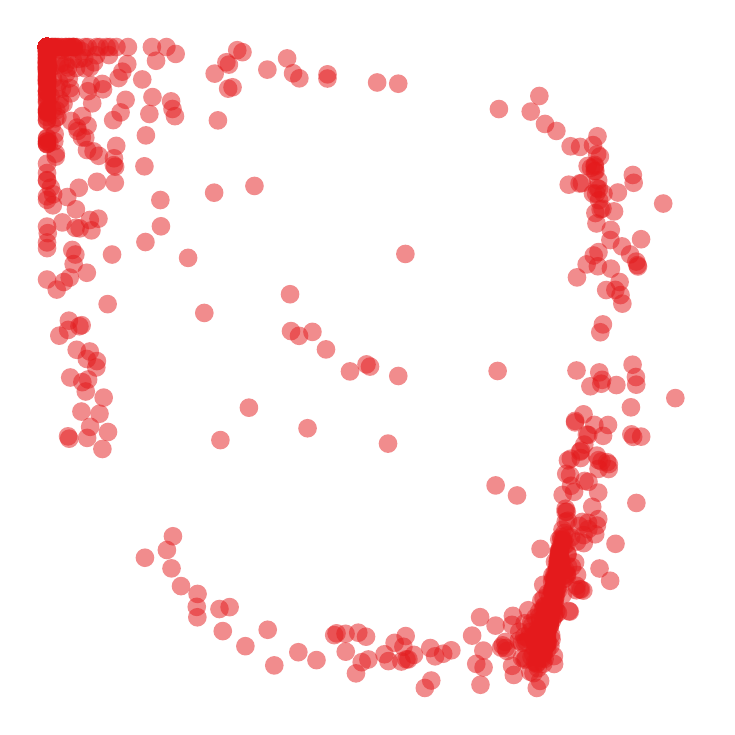}}
    \end{subfigure}\hfill
    \begin{subfigure}[b]{0.22\linewidth}
      \caption{FQL}
      \fbox{\includegraphics[width=\linewidth]{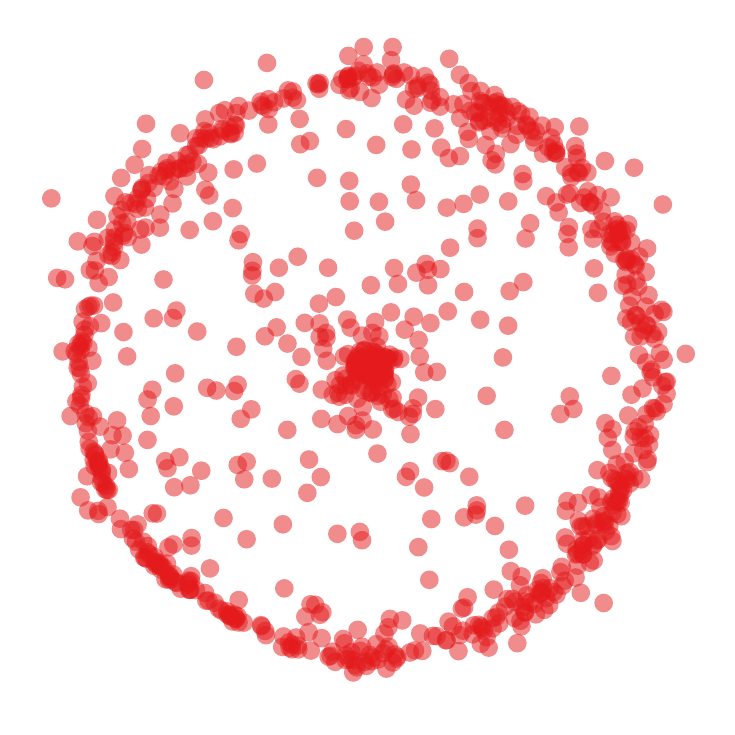}}
    \end{subfigure}

    \vspace{0.3em} 

    % --- row 2 (4 plots) ---
    \begin{subfigure}[b]{0.22\linewidth}
      \caption{ORAAC}
      \fbox{\includegraphics[width=\linewidth]{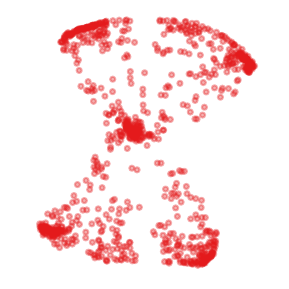}}
    \end{subfigure}\hfill
    \begin{subfigure}[b]{0.22\linewidth}
      \caption{ORAAC-Diff.}
      \fbox{\includegraphics[width=\linewidth]{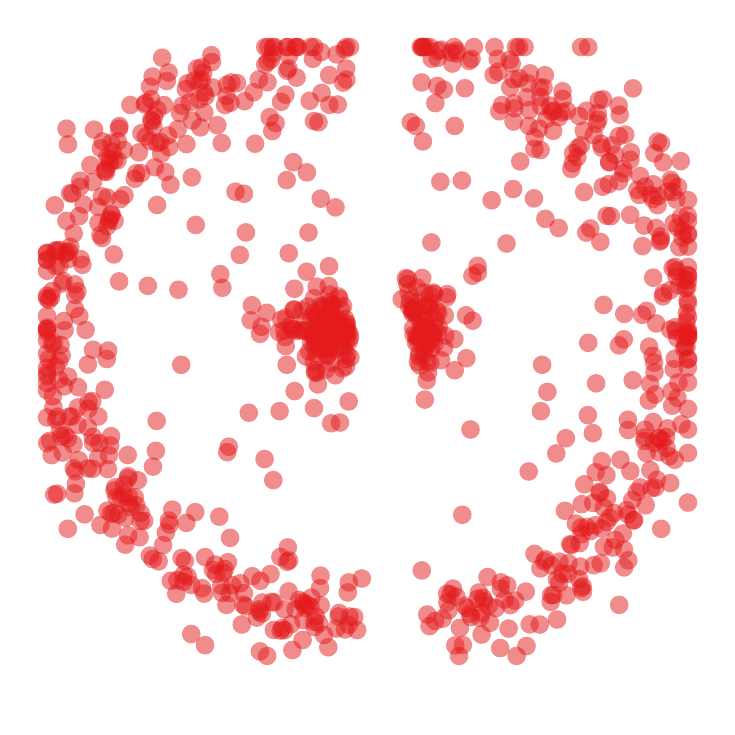}}
    \end{subfigure}\hfill
    \begin{subfigure}[b]{0.22\linewidth}
      \caption{ORAAC-Flow}
      \fbox{\includegraphics[width=\linewidth]{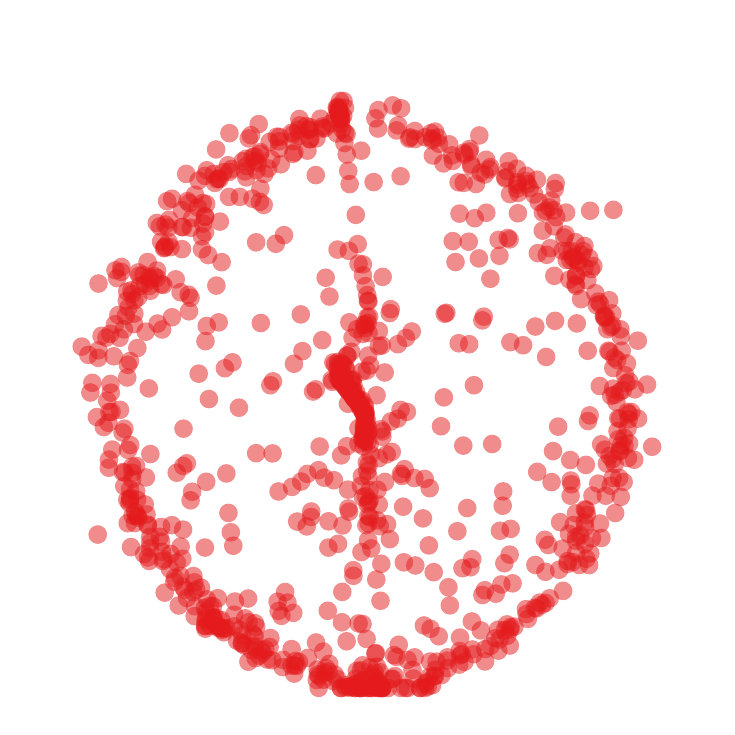}}
    \end{subfigure}\hfill
    \begin{subfigure}[b]{0.22\linewidth}
      \caption{RADAC}
      \fbox{\includegraphics[width=\linewidth]{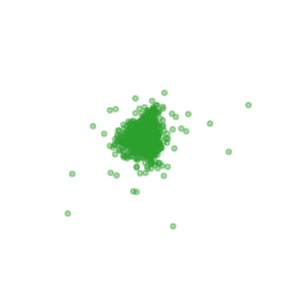}}
    \end{subfigure}

  \end{minipage}

  % \vspace{-0.6em}
  \caption{
    \textbf{Toy Risky Bandit Results}
    \emph{Top:} Ground truth consists of a safe center mode \textcolor[HTML]{AED581}{yellow-green} and a risky ring where high-reward samples \textcolor[rgb]{1.0,0.83,0.3}{yellow}
 are interspersed with catastrophic penalties (\textcolor[rgb]{0.61,0.15,0.69}{purple}). Risk-neutral generative baselines concentrate on the risky ring or collapse topology.
    \emph{Bottom:} Prior‑anchored perturbation methods produce samples in the low‑density inter‑mode region,
    exhibiting OOD leakage. RADAC concentrates near the safe center without collapsing into the low-density bridge / without filling the inter-mode gap. See App.~\ref{app:additional_toy} for more results.}
    \label{fig:toy_rl}
\end{figure*}

\subsection{Behavior‑Regularized Objective: Why It Works Better} In contrast to prior-anchored perturbation, RAMAC applies behavior regularization
\emph{directly} to the \emph{deployed} generative policy $\pi_\theta$ (Eq.~\ref{eq:policy_loss_full}). For explicit-likelihood actors, the BC term corresponds to the negative
log-likelihood and thus controls the forward KL
$D_{\mathrm{KL}}(\beta(\cdot\!\mid\!s)\Vert
\pi_\theta(\cdot\!\mid\!s))$ up to an additive constant.
For our diffusion and flow-matching instantiations, we instead use their
native BC-style surrogates, namely denoising and flow-matching losses,
which anchor the generative process to dataset actions
(see footnote in Sec.~\ref{sec:preliminaries}).
The following KL-based results therefore formalize the idealized
divergence-controlled mechanism, whose OOD implication we evaluate
empirically in Sec.~\ref{subsec:epsilon_act}.
% We therefore monitor the BC loss as a practical proxy for $D_{\mathrm{KL}}(\beta\Vert\pi_\theta).$

    \begin{lemma}
    \label{lem:kl_fence}
    For each state $s$, let $I_s=\{a:\beta(a\mid s)>0\}$ and $O_s=I_s^{c}$.
    Assume $\beta\!\ll\!\pi_\theta$ (absolute continuity on $I_s$).
    Then the per-state OOD probability
    $\delta_s(\pi_\theta)\coloneqq \pi_\theta(O_s\mid s)$ satisfies
    {
    \begin{equation}
    \delta_s(\pi_\theta)\ \le\ 1-\exp\!\Big\{-D_{\mathrm{KL}}\big(\beta(\cdot\mid s)\,\Vert\,\pi_\theta(\cdot\mid s)\big)\Big\}.
    \end{equation}
    }
    \end{lemma}
    This formalizes a simple point: shrinking the forward KL suppresses probability mass outside the behavior support. Proof appears in App.~\ref{app:lemm_kl_proof}.
We can extend this risk-agnostic mechanism to bound its impact
on tail risk objectives.
\begin{proposition}
\label{prop:cvar_stability}
Fix a state $s$. Let $X_s:\mathcal A\to[m,M]$ be any bounded
utility score, such as a clipped return or critic signal
(equivalently, the negative of a bounded cost). For $\alpha\in(0,1]$ and any two
action distributions $p(\cdot\mid s)$ and $q(\cdot\mid s)$, letting $A\sim p(\cdot\mid s)$
and $A'\sim q(\cdot\mid s)$,
{
\begin{align}
\notag
&\big|\mathrm{CVaR}_{\alpha}(X_s(A))-\mathrm{CVaR}_{\alpha}(X_s(A'))\big|\\
 &\le \frac{M-m}{\alpha}\,\mathrm{TV}\!\big(p(\cdot\mid s),q(\cdot\mid s)\big) \notag\\
&\le \frac{M-m}{\alpha}\sqrt{\tfrac12 D_{\mathrm{KL}}\!\big(q(\cdot\mid s)\,\|\,p(\cdot\mid s)\big)}.
\end{align}
}
\end{proposition}
\begin{corollary}
\label{cor:cvar_beta_pi}
Under the assumptions of Prop.~\ref{prop:cvar_stability}, taking
$q=\beta(\cdot\mid s)$ and $p=\pi_\theta(\cdot\mid s)$ yields
{
\begin{align}
\notag
&\big|\mathrm{CVaR}_{\alpha}^{\beta}-\mathrm{CVaR}_{\alpha}^{\pi_\theta}\big|\\
&\le \frac{M-m}{\alpha}\sqrt{\tfrac12
 D_{\mathrm{KL}}\!\big(\beta(\cdot\mid s)\,\|\,\pi_\theta(\cdot\mid s)\big)}.
\end{align}
}
\end{corollary}
\noindent
Proposition~\ref{prop:cvar_stability} and Corollary~\ref{cor:cvar_beta_pi} highlight that CVaR is amplified by $1/\alpha$ under distribution shift. This makes even small behavior mismatch potentially catastrophic in the lower tail, motivating explicit behavior regularization when optimizing $\mathrm{CVaR}_\alpha$.
 Proof is in App.~\ref{app:proof_cvar_tv}.
\begin{table*}[t]
  \centering
  \footnotesize
  \setlength{\tabcolsep}{6pt}
  \renewcommand{\arraystretch}{1.15}
  \caption{Stochastic–D4RL results over 5 seeds. We report Mean and CVaR$_{0.1}$; best in dark/ second in light shaded. Additional fixed-checkpoint results are provided in App.~\ref{app:sd4rl_full}.}
  \label{tab:d4rl_score}
  \begin{tabular}{llccccccc}
    \toprule
    \textbf{Dataset} & \textbf{Metric}
      & \textbf{CQL} & \textbf{CODAC} & \textbf{ORAAC} 
      & \textbf{ORAAC-Diff.} & \textbf{FQL} & \textbf{DiffusionQL} & \textbf{RADAC} \\
    \midrule
    \multirow{2}{*}{HalfCheetah-M-E}
      & Mean & $-66.66$ & $-0.12$ & $796.06$ & $650.70$ & \second{$844.14$} & $-20.71$  & \best{$\mathbf{916.64}$} \\
      & CVaR & $-135.39$ & $-0.11$ & $742.94$ & $455.40$ & \second{$754.44$} & $-76.39$ & \best{$\mathbf{805.25}$} \\[2pt]

    \multirow{2}{*}{Walker2d-M-E}
      & Mean & $-21.52$ & $23.96$ & $969.62$ & $823.20$ & \second{$1309.48$} & $-32.38$ & \best{$\mathbf{1708.68}$} \\
      & CVaR & $-64.88$ & $-43.88$ & $358.55$ & $317.92$ & \second{$468.15$} & $-116.19$ & \best{$\mathbf{573.22}$} \\[2pt]

    \multirow{2}{*}{Hopper-M-E}
      & Mean & $-25.87$ & $26.59$ & \second{$714.15$} & $577.73$ & $341.16$ & $-279.97$ & \best{$\mathbf{1277.74}$} \\
      & CVaR & $-111.37$ & $-150.92$ & \second{$374.63$} & $240.48$ & $-8.80$ & $-872.95$ & \best{$\mathbf{800.64}$} \\[2pt]

    \multirow{2}{*}{HalfCheetah-M-R}
      & Mean & $-66.21$ & $-0.11$ & $18.99$ & $220.00$ & \second{$434.33$} & $279.95$ & \best{$\mathbf{525.84}$} \\
      & CVaR & $-127.09$ & $-1.47$ & $-34.09$ & $44.93$ & \second{$224.73$} & $79.93$ & \best{$\mathbf{278.65}$} \\[2pt]

    \multirow{2}{*}{Walker2d-M-R}
      & Mean & $-16.90$ & $33.59$ & $126.94$ & $4.62$ & \second{$411.36$} & $96.88$  & \best{$\mathbf{615.94}$} \\
      & CVaR & $-51.49$ & $-52.63$ & $-203.64$ & $-113.20$ & $5.08$ & \second{$48.14$} & \best{$\mathbf{145.21}$} \\[2pt]

    \multirow{2}{*}{Hopper-M-R}
      & Mean & $-16.25$ & $-47.83$ & $-18.00$ & $25.68$ & \second{$373.16$} & $-2.79$ & \best{$\mathbf{385.58}$} \\
      & CVaR & $-118.70$ & $-160.08$ & $-129.25$ & $-131.91$ & $-62.24$ & \second{$-51.33$} & \best{$\mathbf{-8.16}$} \\
    \bottomrule
  \end{tabular}
\end{table*}
\subsection{Toy Risky Bandit Example} We design a 2‑D contextual bandit with two disjoint modes (\emph{Toy Risky Bandit})  to make the above geometric analysis concrete: The top-left panel in Fig.~\ref{fig:toy_rl} shows a ground truth that consists of a \emph{safe center}
(moderate reward, no catastrophic tail) and a \emph{risky ring}
(higher mean with rare large penalties). The task isolates multimodality and lower-tail hazards. Below we introduce our baselines.

\subsubsection{Expressive but Risk‑Neutral Controls}

As a common notation, let $G_\phi(\cdot \mid s)$ denote an expressive
conditional generator (diffusion model, flow, or conditional VAE) that
induces the policy $a \sim G_\phi(\cdot \mid s)$.
Our risk-neutral expressive baselines, DiffusionQL~\citep{wang2022diffusion},
FlowQL (FQL)~\citep{park2025flow}, and a conditional VAE-QL (CVAE-QL), all minimize
{
\begin{equation}
\label{eq:rn_objective}
\begin{array}{@{}l@{}}
\mathcal{L}_{\mathrm{RN}}(\phi)=\\
\lambda_{\mathrm{BC}}\,
\mathbb{E}_{(s,a)\sim\mathcal{D}}
\big[\ell\big(a, G_\phi(s)\big)\big]
-\mathbb{E}_{\substack{s\sim\mathcal{D},\\ a\sim G_\phi(\cdot \mid s)}}
\big[Q_\psi(s,a)\big].
\end{array}
\end{equation}
}
i.e., a standard BC loss plus a risk-neutral value-improvement term on
top of an expressive actor $G_\phi$ with scalar critic $Q_\psi$.
Each method simply instantiates $G_\phi$ (diffusion, flow, or CVAE) and
its optimization hyperparameters; full details are given in
App.~\ref{app:synthetic_details}.
\subsubsection{Prior‑Anchored Perturbation}
As risk-aware anchor–perturbation baselines we use ORAAC, ORAAC–Diffusion (ORAAC-Diff.)~\citep{chen2025diffusion,urpi2021risk},
and ORAAC–Flow, all instantiated
via Eq.~\ref{eq:perturbation} with their original coherent risk objectives.
Concretely, given a behavior anchor $b\sim\beta(\cdot\mid s)$, each method learns
a residual perturbation $f_\phi$ as in Sec.~\ref{sec:risk_pitfalls} and optimizes
its own risk functional (e.g., CVaR or distorted expectations) under this
prior-anchored geometry.

\subsubsection{Example Results}
The resulting policy distributions for various methods are shown in Fig.~\ref{fig:toy_rl}.

Risk‑neutral expressive controls (Fig.~\ref{fig:toy_rl} (b-d)): Overall, as expected, these methods are risk-blind and chase high‑$Q$ ridges ignoring the lower tail.
FQL often preserves both modes but does not suppress mass on the hazardous ring;
DiffusionQL drifts toward sparsely covered high‑$Q$ pockets on the ring, yielding risk exposure;
the CVAE variant collapses topology and fills low‑density bridges.

Prior‑anchored perturbation (Fig.~\ref{fig:toy_rl} (e-g)):
ORAAC and its diffusion/flow variants place substantial mass in the inter‑mode low‑density region, regardless of whether the BC prior is expressive or not. These controlled toy results illustrate the geometric leakage mechanism in
Lemma~\ref{lem:anchored_leakage}.

RADAC (Fig.~\ref{fig:toy_rl} (h)): By sending CVaR signals from a distributional critic through the diffusion trajectory
while regularizing with BC, RADAC concentrates probability near the safe center without filling
the gap.
Full configuration and additional plots are in App.~\ref{app:additional_toy} and~\ref{app:synthetic_details}. Overall, these patterns empirically support our analysis.

\section{Experiments}
\label{sec:experiments}
% =========================
In this section, we evaluate \textbf{RADAC} on the Stochastic-D4RL benchmarks to validate both \emph{risk awareness} and \emph{policy expressiveness}. We also quantify the OOD action rate $\varepsilon_{\mathrm{act}}$ (Sec.~\ref{subsec:epsilon_act}) to empirically validate the behavior-divergence mechanism in Sec.~\ref{sec:risk_pitfalls}.
Additional results, including the flow-matching instantiation \textbf{RAFMAC}, appear in App.~\ref{app:sd4rl_full}.
\paragraph{Tasks}
We augment standard D4RL locomotion tasks~\citep{fu2020d4rl} with rare heavy‑tailed penalties tied to forward velocity (HalfCheetah) or torso pitch angles (Hopper, Walker2d), together with early termination when the torso leaves a healthy range, following~\citep{urpi2021risk}. We evaluate on \textsc{Hopper}, \textsc{Walker2d}, and \textsc{HalfCheetah} using the \textsc{Medium–Expert (M-E)} and \textsc{Medium–Replay (M-R)} datasets, which are typically multimodal due to mixed data-collection policies. This lets us examine whether RAMAC learns \emph{risk‑aware} policies \emph{without sacrificing multimodality}. Full construction details appear in App.~\ref{app:sd4rl_details}.
% \textcolor{blue}{During training, all methods receive per-transition rewards relabeled with the same stochastic hazard model that is also used at evaluation time, so they observe identical logged return signals and never see the underlying hazard mask or triggers directly;}
\paragraph{Baselines}
We compare against representative offline‑RL methods covering value conservatism, distributional conservatism, anchor-perturbation risk aversion, and risk-neutral expressive generators:
CQL~\citep{kumar2020conservative}, CODAC~\citep{ma2021conservative}, ORAAC~\citep{urpi2021risk}, 
ORAAC-Diffusion (ORAAC-Diff.)~\citep{chen2025diffusion}, DiffusionQL~\citep{wang2022diffusion}, and FlowQL (FQL)~\citep{park2025flow}.
We defer detailed configurations to App.~\ref{app:baselines}.
% ----- Place this after \paragraph{Baselines} and before Results -----
\paragraph{Evaluation}
Following the protocols adopted in~\citep{urpi2021risk,wang2022diffusion}, we train for 2000 epochs, each with 1000 gradient steps. We evaluate methods at fixed intervals of gradient steps and report (i) raw returns averaged over 5 seeds and (ii) episodic \(\mathrm{CVaR}_{0.1}\)  computed over 50 rollouts (10 evaluation episodes per seed) to avoid score normalization artifacts. For ORAAC and CODAC, we adopt the authors’ risk-aware objectives (risk level \(\alpha{=}0.1\)). For the other baselines, we tune hyperparameters to ensure fairness and otherwise use authors’ recommended settings.
% For RADAC and RAFMAC, we apply early stopping and select the checkpoint with the lowest or second-lowest \(\mathcal{L}_{\mathrm{BC}}(\theta)\); for the baselines we follow the selection rule in their original papers.
Further protocol details appear in App.~\ref{app:sd4rl_details}. Runtime and inference-latency comparisons for RADAC with recent baselines are reported in App.~\ref{app:runtime}.
% Further protocol details appear in App.~\ref{app:experimental_details}. 
% For artifact reproducibility, following common practice, we provide full results with corresponding 1000-step evaluation in App.~\ref{app:additional_experimental_results}.

\subsection{Results and Analysis}
\label{sec:results}

Table~\ref{tab:d4rl_score} reports Mean and CVaR$_{0.1}$ for RADAC alongside baselines. Across six tasks, RADAC delivers \emph{strong lower tails with competitive or higher means}. 
FQL is often the strongest risk-neutral baseline, suggesting that flow-based generators can stabilize training
even without explicit tail-risk shaping, but its lower-tail returns remain below
RADAC. Further analysis is provided in the appendix, including an ablation on the tail-sample budget $K$ in the CVaR estimator (App.~\ref{app:cvar-k}), an objective ablation separating the BC and CVaR terms (App.~\ref{app:objective_ablation}), and a comparison of alternative risk distortions (App.~\ref{app:risk_ablation}).
% ORAAC regularizes toward a behavior anchor. It reliably handles sharp hazardous thresholds on tasks such as \textsc{Hopper‑Medium‑Expert} but may fail to exploit high-reward modes, and can place mass in low-density regions between modes when the anchor lies in a risky region.

%----------------------------------------------------------
\subsection{Qualitative Safety Analysis}
\label{subsec:safety_plots}
\begin{figure*}[t]
 \hspace*{0.8em}
  %================= Row 1 =================
  \begin{subfigure}[t]{.30\textwidth}
    \caption{\small HalfCheetah-Medium-Replay}
    \makebox[0pt][r]{%
      \raisebox{.25\linewidth}{\rotatebox{90}{\normalsize\text{Density}}}%
    }%
    \begin{tikzpicture}
      \node[inner sep=0pt, name=img]
        {\includegraphics[width=\linewidth]{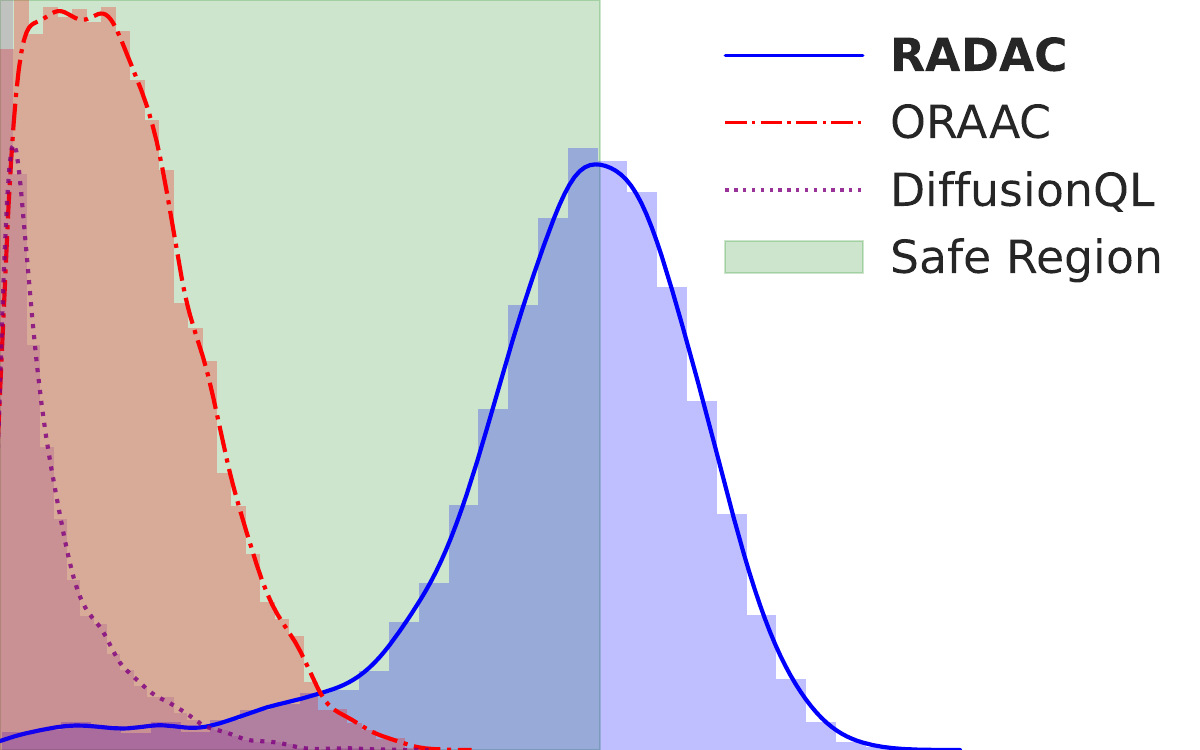}};
      \def\inset{0.6pt}
      \draw[line width=0.35pt]
        ($(img.south west)+(\inset,\inset)$) rectangle
        ($(img.north east)+(-\inset,-\inset)$);
      \coordinate (FSW) at ($(img.south west)+(\inset,\inset)$);
      \coordinate (FSE) at ($(img.south east)+(-\inset,\inset)$);
      \def\ticklen{2.0pt}
      \def\labelsep{0.7pt}
      \foreach \x/\lab in {0.007/0, 0.5007/5}{
        \coordinate (P) at ($(FSW)!\x!(FSE)$);
        \draw (P) -- ++(0,-\ticklen);
        \node[anchor=north, inner sep=0pt, font=\footnotesize\bfseries]
             at ($(P)+(0,-\ticklen-\labelsep)$) {\lab};
      }
    \end{tikzpicture}
  \end{subfigure}\hspace{0.65em}
  \begin{subfigure}[t]{.30\textwidth}
    \caption{\small Walker2d-Medium-Replay}
    \begin{minipage}[t]{\dimexpr\linewidth-\figpad\relax}
      \begin{tikzpicture}
        \node[inner sep=0pt, name=img]
          {\includegraphics[width=\linewidth]{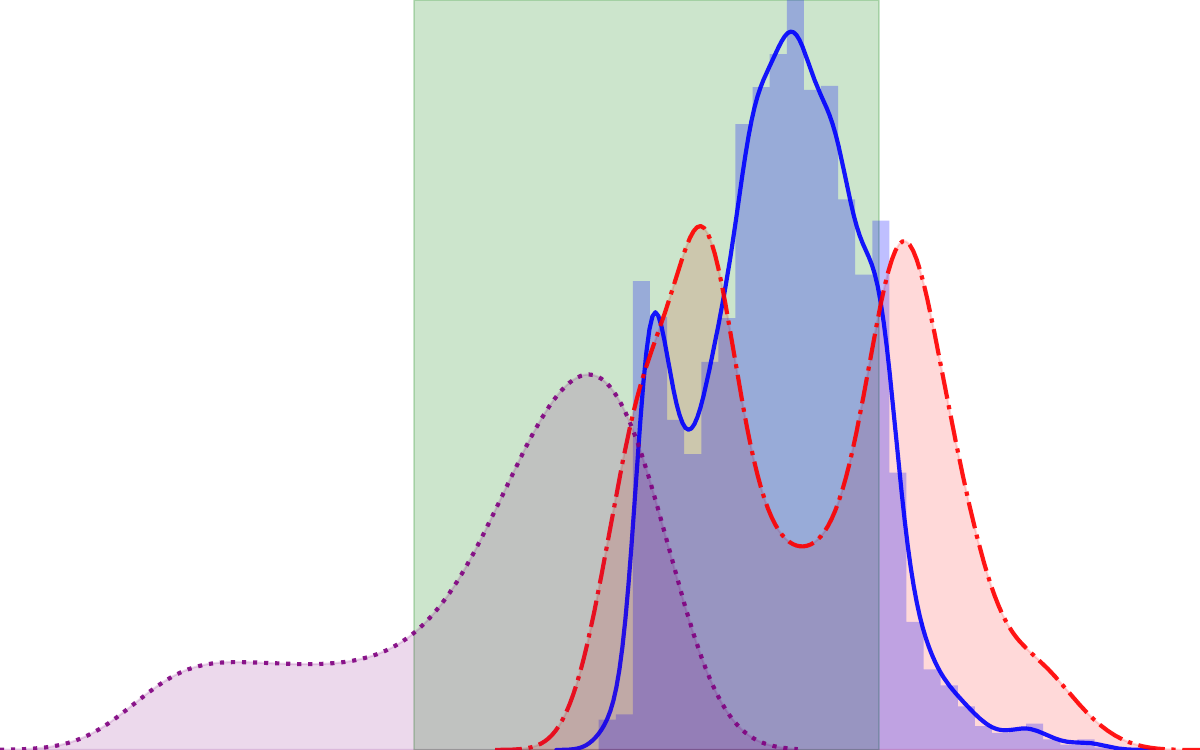}};
        \def\inset{0.6pt}
        \draw[line width=0.35pt]
          ($(img.south west)+(\inset,\inset)$) rectangle
          ($(img.north east)+(-\inset,-\inset)$);
        \coordinate (FSW) at ($(img.south west)+(\inset,\inset)$);
        \coordinate (FSE) at ($(img.south east)+(-\inset,\inset)$);
        \def\ticklen{2.0pt}\def\labelsep{0.7pt}
        \foreach \x/\lab in {0.35/-0.5, 0.555/0, 0.735/0.5}{
          \coordinate (P) at ($(FSW)!\x!(FSE)$);
          \draw (P) -- ++(0,-\ticklen);
          \node[anchor=north, inner sep=0pt, font=\footnotesize\bfseries]
               at ($(P)+(0,-\ticklen-\labelsep)$) {\lab};
        }
      \end{tikzpicture}
    \end{minipage}
  \end{subfigure}\hspace{0.55em}
  \begin{subfigure}[t]{.30\textwidth}
    \caption{\small Hopper-Medium-Replay}
    \begin{minipage}[t]{\dimexpr\linewidth-\figpad\relax}
      \begin{tikzpicture}
        \node[inner sep=0pt, name=img]
          {\includegraphics[width=\linewidth]{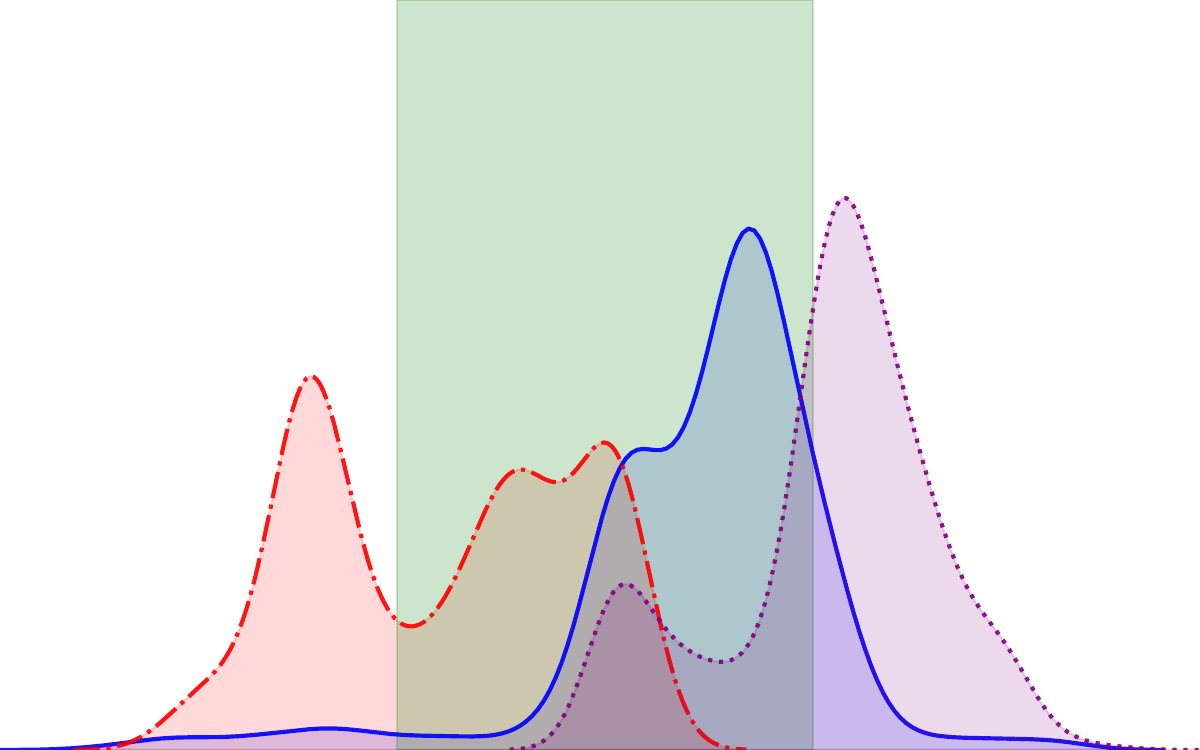}};
        \def\inset{0.6pt}
        \draw[line width=0.35pt]
          ($(img.south west)+(\inset,\inset)$) rectangle
          ($(img.north east)+(-\inset,-\inset)$);
        \coordinate (FSW) at ($(img.south west)+(\inset,\inset)$);
        \coordinate (FSE) at ($(img.south east)+(-\inset,\inset)$);
        \def\ticklen{2.0pt}\def\labelsep{0.7pt}
        \foreach \x/\lab in {0.34/-0.1, 0.51/0, 0.68/0.1}{
          \coordinate (P) at ($(FSW)!\x!(FSE)$);
          \draw (P) -- ++(0,-\ticklen);
          \node[anchor=north, inner sep=0pt, font=\footnotesize\bfseries]
               at ($(P)+(0,-\ticklen-\labelsep)$) {\lab};
        }
      \end{tikzpicture}
    \end{minipage}
  \end{subfigure}
  \caption{Policy distributions for RADAC, ORAAC, and DiffusionQL; shaded bands indicate safe operational ranges (HalfCheetah: $v\!\le\!5$; Hopper: $|\theta|\!\le\!0.1$; Walker2d: $|\theta|\!\le\!0.5$). RADAC reduces mass beyond thresholds.}
  \label{fig:safety_heatmaps}
\end{figure*}
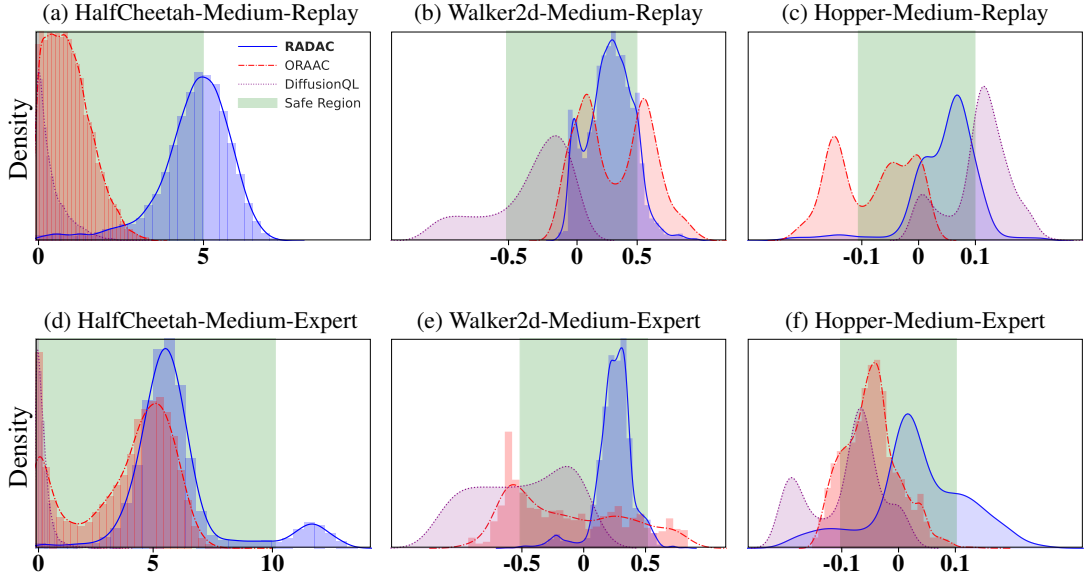

We further visualize a contrast among three representative methods: risk-aware expressive generator RADAC, risk-neutral expressive generator DiffusionQL, and the anchor-perturb risk-averse method ORAAC. Fig.~\ref{fig:safety_heatmaps} plots the monitored distribution of policies against safe regions. RADAC concentrates probability mass inside or near the risk-free boundary while \emph{actively reallocating probability to high‑return modes that lie within these regions}. DiffusionQL is tightly concentrated around zero because rare, high penalties depress bootstrapped values near the risk-free boundary. On the other hand, ORAAC regularizes toward a behavior anchor and thus can place nontrivial mass in low-density inter-mode regions due to its anchored-perturbation geometry. These features can be further observed in 
App.~\ref{app:pareto_frontiers} which plots Pareto frontiers of mean return versus per-episode safety-violation counts. Additionally, we show empirical return distributions and risk–return frontiers in App.~\ref{app:return-dists} and App.~\ref{app:risk_return_frontier}.
\subsection{Empirical Validation of the OOD Mechanism}
\label{subsec:epsilon_act}
% \begin{table}[!t]
%   \centering
%   \footnotesize
%   \setlength{\tabcolsep}{5pt}
%   \caption{State-conditioned OOD action rate ($\% \pm$ std over 3 seeds) on \textsc{medium–expert} ($\kappa{=}3$).}
%   \label{tab:ood_lb_compact}
%   \begin{tabular}{lccc}
%     \toprule
%     \textbf{Task} & \textbf{RADAC} & \textbf{ORAAC} & \textbf{ORAAC-Diff.}\\
%     \midrule
%     HalfCheetah & $1.22\!\pm\!0.21$ &
%     $29.27\!\pm\!12.76$ & $35.4\!\pm\!25.9$ \\
%     Walker2d    & $0.96\!\pm\!0.45$ & $5.57\!\pm\!3.31$ & $18.5\!\pm\!2.71$\\
%     Hopper      & $1.21\!\pm\!0.21$ & $11.60\!\pm\!3.95$ & $22.7\!\pm\!6.56$\\
%     \bottomrule
%   \end{tabular}
% \end{table}
We now provide measurements of OOD actions to validate the insights in Sec.~\ref{sec:risk_pitfalls}. For each policy, we report $\varepsilon_{\text{act}}$, the fraction of evaluation state-action pairs $(s_t,a_t)$ whose \emph{state-conditioned} 1-NN action distance exceeds $\kappa \times \mathrm{median}\, d_{\mathrm{NN}}$. Sec.~\ref{sec:risk_pitfalls} predicts that (i) the BC-regularized CVaR objective should reweight probability \emph{within} the data manifold, yielding low $\varepsilon_{\text{act}}$, and (ii) ORAAC variants, being based on anchor–perturbation, should exhibit \emph{higher} $\varepsilon_{\text{act}}$ than RADAC; 
Table~\ref{tab:ood_lb_compact} confirms this prediction: 
RADAC retains low OOD across tasks, whereas ORAAC variants are consistently higher, even with an expressive diffusion prior, matching the expected geometric leakage from anchor–perturbation. Similar RADAC $<$ ORAAC rankings hold under alternative OOD detectors (App.~\ref{app:ood_detectors}). RADAC achieves risk awareness and expressiveness simultaneously while maintaining low OOD rates.
\par\medskip
{\centering
\footnotesize
\captionof{table}{State-conditioned OOD action rate
($\% \pm$ std over 3 seeds) on \textsc{medium--expert} ($\kappa{=}3$).}
\label{tab:ood_lb_compact}
\setlength{\tabcolsep}{5pt}
\begin{tabular}{lccc}
  \toprule
  \textbf{Task} & \textbf{RADAC} & \textbf{ORAAC} & \textbf{ORAAC-Diff.}\\
  \midrule
  HalfCheetah & $1.22\!\pm\!0.21$ &
  $29.27\!\pm\!12.76$ & $35.4\!\pm\!25.9$ \\
  Walker2d    & $0.96\!\pm\!0.45$ &
  $5.57\!\pm\!3.31$ & $18.5\!\pm\!2.71$ \\
  Hopper      & $1.21\!\pm\!0.21$ &
  $11.60\!\pm\!3.95$ & $22.7\!\pm\!6.56$ \\
  \bottomrule
\end{tabular}
\par}

\section{Conclusions}
This paper introduces RAMAC, a model-free framework for risk-aware offline RL with expressive generative policies. RAMAC couples a distributional critic with a diffusion/flow actor and optimizes a single composite objective that combines behavioral cloning (BC) with CVaR. Our analysis highlights two complementary mechanisms: (i) prior-anchored perturbation can retain off-support leakage on thin or
nonconvex supports when its feasible perturbation region overlaps the
support complement and the induced policy maintains mass there, and (ii) controlling behavior divergence of the deployed policy suppresses OOD action probability and stabilizes tail objectives such as $\mathrm{CVaR}_\alpha$ under distribution shift. On Stochastic-D4RL benchmarks, RADAC achieves consistently stronger lower-tail returns (CVaR$_{0.1}$) while maintaining competitive mean performance and lower measured OOD action rates than representative anchor-perturbation risk-aware baselines.

\section*{Impact Statement}
This work advances risk-aware learning for decision-making policies, with the goal of improving reliability under distribution shift and rare adverse outcomes. Potential positive impacts include safer deployment of learned controllers in domains such as robotics and autonomous systems, and more informative evaluation of tail risks. 

As with most learning-based decision systems, misuse or misdeployment could lead to unsafe behavior if models are applied outside their intended operating conditions or without adequate testing. We mitigate this by explicitly evaluating under heavy-tailed penalties and reporting risk-sensitive metrics, and by encouraging practitioners to validate performance and safety in their target environments before deployment.


\bibliography{example_paper}
\bibliographystyle{icml2026}

\appendix
\onecolumn
% \section{Limitations and Future Directions}
% \label{app:limitations}
% Practical remedies include adding a distilled \emph{one-step} RL head to avoid recursive backprop, and using score- or energy–weighted objectives that reduce full-path backprop for diffusion/flow policies~\citep{koirala2025flow,park2025flow}; both are natural directions for future improvement. On the critic side,  modeling the \emph{return distribution} itself with diffusion/flow value networks is a natural extension that may improve tail calibration without sacrificing actor expressiveness \citep{agrawalla2025floq,zhang2025d2}. Our theory is deliberately scoped to proposition-level guidance (e.g., KL–OOD relations) and does not provide finite-sample, high-probability guarantees under function approximation or partial coverage. Strengthening these to non-asymptotic guarantees, exploring alternatives to BC ( $f$‑divergences or Wasserstein) to trade off mode‑covering vs.\ mode‑seeking; extending analysis to \emph{dynamic/spectral risk} and partial observability, and developing \emph{risk‑aware offline‑to‑online fine‑tuning} that preserves on‑manifold exploration are all promising directions for future work. 
\section{Limitations and Future Directions}
\label{app:limitations}

Practical remedies include adding a distilled \emph{one-step} RL head to avoid recursive backprop, and using score- or energy-weighted objectives that reduce full-path backprop for diffusion/flow policies~\citep{koirala2025flow,park2025flow}; both are natural directions for future improvement. On the critic side, modeling the \emph{return distribution} itself with diffusion/flow value networks is a natural extension that may improve tail calibration without sacrificing actor expressiveness~\citep{agrawalla2025floq,zhang2025d2}.

Our theory is deliberately scoped to proposition-level guidance (e.g., KL--OOD relations) and does not provide finite-sample, high-probability guarantees under function approximation or partial coverage. Moreover, our primary evaluation uses CVaR as a direct objective for offline lower-tail risk control. While RAMAC can incorporate alternative distortion-based objectives, as illustrated by the Wang and CPW results in App.~\ref{app:risk_ablation}, we do not study broader risk formulations such as mean--variance or entropic risk, nor explicit cost-constraint or barrier-based safety objectives; these constitute complementary directions to the setting considered here.

Promising future directions include strengthening the analysis to obtain non-asymptotic guarantees, exploring alternatives to BC (e.g., $f$-divergences or Wasserstein objectives) to trade off mode-covering versus mode-seeking behavior, extending the framework to dynamic or spectral risk measures and partial observability, connecting lower-tail risk optimization with explicit constrained-safety formulations, and developing \emph{risk-aware offline-to-online fine-tuning} that preserves on-manifold exploration.
\section{Proofs}
\label{app:proofs}
\subsection{Proof of Lemma~\ref{lem:anchored_leakage}}
\label{app:lemma1_proof}
% \begin{proof}
% By assumption, $\pi_{\text{anch}}(\cdot\mid s)$ has a density
% $p(a\mid s)$ on $B_\Phi(b^\star)$ with $p(a\mid s)\ge c>0$.
% Thus
% \[
% \delta_s(\pi_{\text{anch}})
% = \int_{O_s} p(a\mid s)\,da
% \;\ge\; \int_{B_\Phi(b^\star)\cap O_s} p(a\mid s)\,da
% \;\ge\; c\cdot\lambda\bigl(B_\Phi(b^\star)\cap O_s\bigr) > 0.
% \]
% The final claim follows as long as the density lower bound $c$ on
% $B_\Phi(b^\star)$ is maintained during training.
% \end{proof}
\begin{proof}
By assumption, $\pi_{\text{anch}}(\cdot\mid s)$ has a density
$p(a\mid s)$ satisfying $p(a\mid s)\ge c>0$ on the measurable
leakage region $A_s\subseteq O_s$. Thus
\[
\delta_s(\pi_{\text{anch}})
= \int_{O_s} p(a\mid s)\,da
\;\ge\; \int_{A_s} p(a\mid s)\,da
\;\ge\; c\cdot\lambda(A_s) > 0.
\]
\end{proof}
\subsubsection{Geometric Intuition Behind Lemma~\ref{lem:anchored_leakage}}
\label{app:lemma_geometric}
{Lemma~\ref{lem:anchored_leakage} formalizes the geometric intuition that
when the perturbation ball centered at $b$ overlaps the complement of
the support, a residual policy can assign off-support probability mass;
such leakage persists whenever the induced policy maintains nontrivial
density on the overlapping region.
The following points provide an intuitive view:
\begin{itemize}[leftmargin=1.15em,itemsep=2pt]
  \item \textbf{Thin support near $b$:}
        When the local margin $m(b)$ is small, any ball $B_\Phi(b)$ with
        $\Phi\gtrsim m(b)$ necessarily overlaps $O_s$, so some residual updates
        produce OOD actions even if $\|\zeta_\psi\|\le\Phi$.
  \item \textbf{Nonconvex support:}
        If $\mathcal S_G(s)$ is nonconvex (e.g., a ring/annulus),
        anchors can sit near holes or concavities, again yielding small $m(b)$ and
        forcing overlap between $B_\Phi(b)$ and $O_s$.
  \item \textbf{Gradients pushing off the data surface:}
        The residual $\zeta_\psi(s,b)$ is trained to increase $Q$ or CVaR and is not
        constrained to remain tangent to the manifold, so gradients can push along
        the normal direction toward the boundary of $B_\Phi(b)$, further amplifying
        leakage when $m(b)\le\Phi$.
\end{itemize}
Even when a coherent risk objective such as CVaR is used to train
$\zeta_\psi$, Lemma~\ref{lem:anchored_leakage} shows that OOD mass
persists whenever the induced policy maintains nontrivial density on
an off-support portion of its feasible perturbation region.
Thin or nonconvex supported regions make this condition more likely.}
\subsection{Proof of Lemma~\ref{lem:kl_fence}}
\label{app:lemm_kl_proof}
\begin{proof}
Fix a state $s$ and define the $\beta$--support
$I_s \coloneqq \{a : \beta(a\mid s)>0\}$ and its complement $O_s \coloneqq I_s^{c}$.
Assume $\beta\ll \pi_\theta$ on $I_s$ (so $\pi_\theta(I_s\mid s)>0$ and the forward KL is finite).

\medskip
Since $\beta(\cdot\mid s)$ has all its mass on $I_s$,
\begin{align*}
\mathrm{KL}(\beta\Vert \pi_\theta)
&= \int_{\mathcal A} \beta(a\mid s)\,\log\frac{\beta(a\mid s)}{\pi_\theta(a\mid s)}\,da
 \;=\; \int_{I_s} \beta(a\mid s)\,\log\frac{\beta(a\mid s)}{\pi_\theta(a\mid s)}\,da. \tag{1}
\end{align*}

Define the normalization of $\pi_\theta$ to $I_s$:
\[
\pi_I(a\mid s) \;\coloneqq\; \pi_\theta(a\mid s,\ a\in I_s)
\;=\; \frac{\pi_\theta(a\mid s)}{\pi_\theta(I_s\mid s)},\qquad a\in I_s,
\]
so that on $I_s$ we have the identity $\pi_\theta(a\mid s)=\pi_\theta(I_s\mid s)\,\pi_I(a\mid s)$.

\medskip
Substitute the above factorization into (1) and use $\log(xy)=\log x + \log y$:
\begin{align*}
\log\frac{\beta(a\mid s)}{\pi_\theta(a\mid s)}
&=\log\frac{\beta(a\mid s)}{\pi_\theta(I_s\mid s)\,\pi_I(a\mid s)}
 =\log\frac{\beta(a\mid s)}{\pi_I(a\mid s)}-\log \pi_\theta(I_s\mid s). \tag{2}
\end{align*}

Plug (2) into (1) and split the integral:
\begin{align*}
\mathrm{KL}(\beta\Vert \pi_\theta)
&=\int_{I_s}\beta(a\mid s)\,\log\frac{\beta(a\mid s)}{\pi_I(a\mid s)}\,da
 \;-\; \int_{I_s} \beta(a\mid s)\,\log \pi_\theta(I_s\mid s)\,da \\
&=\underbrace{\mathrm{KL}\big(\beta(\cdot\mid s)\,\Vert\,\pi_I(\cdot\mid s)\big)}_{\ge 0}
 \;-\; \log \pi_\theta(I_s\mid s)\,\underbrace{\int_{I_s}\beta(a\mid s)\,da}_{=1}. \tag{3}
\end{align*}
Here the last equality uses that $\log \pi_\theta(I_s\mid s)$ is constant in $a$, and that
$\beta$ places total mass $1$ on $I_s$.

\medskip
From (3) and nonnegativity of KL,
\[
\mathrm{KL}(\beta\Vert \pi_\theta)\;\ge\; -\,\log \pi_\theta(I_s\mid s).
\]
Exponentiating both sides gives
\[
e^{-\mathrm{KL}(\beta\Vert \pi_\theta)} \;\le\; \pi_\theta(I_s\mid s).
\]
Since $\pi_\theta(I_s\mid s)=1-\delta_s(\pi_\theta)$ with $\delta_s(\pi_\theta)\coloneqq \pi_\theta(O_s\mid s)$,
we obtain the per–state OOD bound
\[
\delta_s(\pi_\theta)\ \le\ 1-\exp\!\bigl\{-\mathrm{KL}(\beta(\cdot\mid s)\Vert \pi_\theta(\cdot\mid s))\bigr\}.
\qedhere
\]
\end{proof}
\subsection{Proof of Proposition~\ref{prop:cvar_stability} and Corollary~\ref{cor:cvar_beta_pi}}
\label{app:proof_cvar_tv}

We first recall two standard facts.
(i) \emph{Dual representation of lower-tail CVaR.}
For $\alpha\in(0,1]$ and any integrable random variable $X$,
\begin{equation}
\label{eq:cvar_dual}
\mathrm{CVaR}_{\alpha}(X)
=\inf_{w\in\mathcal W_{\alpha}}\; \mathbb E[wX],
\qquad
\mathcal W_{\alpha}\coloneqq\Bigl\{w:\ 0\le w\le \tfrac{1}{\alpha},\ \mathbb E[w]=1\Bigr\}.
\end{equation}
(See, e.g., standard references on coherent risk measures / expected shortfall.)

(ii) \emph{Coupling characterization of total variation.}
For distributions $p,q$ on $\mathcal A$,
\begin{equation}
\label{eq:tv_coupling}
\mathrm{TV}(p,q)=\inf_{\gamma\in\Gamma(p,q)}\ \mathbb P_{\gamma}(A\neq A'),
\end{equation}
where $\Gamma(p,q)$ denotes the set of all couplings $\gamma$ whose marginals are
$A\sim p$ and $A'\sim q$.

\begin{proof}[Proof of Proposition~\ref{prop:cvar_stability}]
Fix $s$ and let $A\sim p(\cdot\mid s)$ and $A'\sim q(\cdot\mid s)$ be coupled by an
arbitrary coupling $\gamma\in\Gamma(p,q)$.
Define
\[
X \coloneqq X_s(A),\qquad Y \coloneqq X_s(A').
\]
Since $X_s:\mathcal A\to[m,M]$, we have $X,Y\in[m,M]$ almost surely.

\paragraph{Step 1: CVaR is $(1/\alpha)$-Lipschitz in $L_1$.}
Let $w_Y\in\arg\min_{w\in\mathcal W_{\alpha}}\mathbb E[wY]$ be a minimizer in
\eqref{eq:cvar_dual} for $Y$ (if not attained, take an $\varepsilon$-minimizer and let
$\varepsilon\to 0$).
Then,
\begin{align}
\mathrm{CVaR}_{\alpha}(X)-\mathrm{CVaR}_{\alpha}(Y)
&= \inf_{w\in\mathcal W_{\alpha}}\mathbb E[wX]\;-\;\inf_{w\in\mathcal W_{\alpha}}\mathbb E[wY]\notag\\
&\le \mathbb E[w_Y X]-\mathbb E[w_Y Y]
= \mathbb E[w_Y(X-Y)]\notag\\
&\le \mathbb E[w_Y|X-Y|]
\le \tfrac{1}{\alpha}\,\mathbb E[|X-Y|]. \label{eq:cvar_l1_one_side}
\end{align}
Swapping the roles of $X$ and $Y$ yields the reverse inequality, hence
\begin{equation}
\label{eq:cvar_l1}
\bigl|\mathrm{CVaR}_{\alpha}(X)-\mathrm{CVaR}_{\alpha}(Y)\bigr|
\le \tfrac{1}{\alpha}\,\mathbb E[|X-Y|].
\end{equation}

\paragraph{Step 2: Bound $\mathbb E|X-Y|$ by TV via coupling.}
Since $X_s$ is deterministic given the action input, $A=A'$ implies $X_s(A)=X_s(A')$.
Thus,
\[
|X-Y|
=|X_s(A)-X_s(A')|
=|X_s(A)-X_s(A')|\mathbf 1_{\{A\neq A'\}}
\le (M-m)\mathbf 1_{\{A\neq A'\}},
\]
and taking expectation under the coupling $\gamma$ gives
\begin{equation}
\label{eq:l1_to_mismatch}
\mathbb E_{\gamma}[|X-Y|]\le (M-m)\,\mathbb P_{\gamma}(A\neq A').
\end{equation}
Combining \eqref{eq:cvar_l1} and \eqref{eq:l1_to_mismatch},
\[
\bigl|\mathrm{CVaR}_{\alpha}(X_s(A))-\mathrm{CVaR}_{\alpha}(X_s(A'))\bigr|
\le \tfrac{M-m}{\alpha}\,\mathbb P_{\gamma}(A\neq A').
\]
Since $\gamma$ was arbitrary, we can minimize over all couplings and apply
\eqref{eq:tv_coupling}:
\begin{equation}
\label{eq:cvar_tv}
\bigl|\mathrm{CVaR}_{\alpha}(X_s(A))-\mathrm{CVaR}_{\alpha}(X_s(A'))\bigr|
\le \tfrac{M-m}{\alpha}\,\mathrm{TV}\!\bigl(p(\cdot\mid s),q(\cdot\mid s)\bigr).
\end{equation}
This proves the first inequality.

\paragraph{Step 3: Pinsker.}
Using Pinsker's inequality and symmetry of TV,
\[
\mathrm{TV}(p,q)=\mathrm{TV}(q,p)
\le \sqrt{\tfrac12 D_{\mathrm{KL}}\!\bigl(q(\cdot\mid s)\,\|\,p(\cdot\mid s)\bigr)},
\]
and substituting into \eqref{eq:cvar_tv} yields the second inequality.
\end{proof}

\begin{proof}[Proof of Corollary~\ref{cor:cvar_beta_pi}]
Apply Proposition~\ref{prop:cvar_stability} with
$p(\cdot\mid s)=\pi_\theta(\cdot\mid s)$ and $q(\cdot\mid s)=\beta(\cdot\mid s)$.
Then \eqref{eq:cvar_tv} plus Pinsker gives
\[
\bigl|\mathrm{CVaR}_{\alpha}^{\beta}-\mathrm{CVaR}_{\alpha}^{\pi_\theta}\bigr|
\le \tfrac{M-m}{\alpha}\sqrt{\tfrac12
 D_{\mathrm{KL}}\!\bigl(\beta(\cdot\mid s)\,\|\,\pi_\theta(\cdot\mid s)\bigr)}.
\]
\end{proof}

\section{Related Work}
\label{app:related_works}
We review works most relevant to our \emph{risk-aware generative trajectory} view—policies that map noise to actions through a differentiable path and how safety is enforced therein—while avoiding repetition of the core background already covered in the main text. For a broad taxonomy of offline RL, see~\citep{prudencio2023survey}.

\paragraph{Expressive Generative Policies}
The main paper reviews diffusion and flow-matching policies (e.g., DiffusionQL, IDQL, FQL). Here we note complementary developments not detailed there: 
% (i) \emph{DDIM-style imitation learning} that accelerates inference while keeping diffusion’s expressiveness~\citep{song2021denoising};
(i) \emph{Diffusion-policy imitation learning with DDIM-style fast sampling}
that accelerates inference while retaining diffusion expressiveness in
visuomotor or robot control~\citep{ankile2024juicer,chi2025diffusion}; (ii) on the flow-matching side, recent generative
policies for robot manipulation replace diffusion with flow matching to achieve similar expressiveness
with faster, more stable inference~\citep{gao2025vita,yan2025maniflow,zhang2025flowpolicy}; (iii) \emph{transformer-based trajectory and policy models}, including
trajectory- and sequence-level generative decision-makers such as Decision Diffuser and Decision
Transformer, and hierarchical visuomotor transformers for bimanual or gaze-conditioned manipulation~\citep{ajay2022conditional,chen2021decision,chuang2025look, janner2022planning,lee2024interact}. These works bolster the case for expressive generative decision-making models but remain \emph{risk-neutral} in objective design.

\paragraph{Risk-Sensitive RL}
Beyond expectation-oriented objectives, risk-sensitive control formalizes tail-aware criteria via coherent/dynamic risk measures for MDPs~\citep{ruszczynski2010}. Among coherent measures, CVaR admits sampling- and policy-gradient formulations suitable for RL~\citep{tamar2015coherent,tamar2015optimizing}, and has been linked to robustness via CVaR–robust trade-offs~\citep{chow2015risk}. In the \emph{offline} regime, safety is often operationalized as high-confidence off-policy evaluation/improvement from fixed logs~\citep{laroche2019safe,thomas2015high}, which bound deployment risk yet do not address how \emph{expressive generators} should receive lower-tail gradients.

\paragraph{Mixture-Policy CVaR Optimization.} Luo et al.~\citep{luo2024simple} propose a mixture policy parameterization for
CVaR optimization in online RL, where a risk-neutral policy and an adjustable
component are combined to form a risk-averse policy and improve the sample
efficiency of CVaR policy gradients. While both their work and ours aim to
shape the lower tail of returns, their method uses standard parametric actors
and does not address offline data, generative policies, or explicit constraints
on out-of-distribution actions. By contrast, RAMAC targets offline risk-sensitive control with a
single expressive generative actor (diffusion/flow), a distributional
critic, a BC+CVaR objective, an analysis of per-state OOD-action
suppression, and hazard relabeling.

\paragraph{Closest Lines and Delineation}
Concurrent actor–critic lines that couple diffusion with value learning remain expectation-oriented:\citep{zhang2025d2} stabilizes \emph{online} diffusion actors with distributional critics and double-$Q$ but does not backpropagate CVaR along the denoising path;\citep{fang2025diffusion} formulates \emph{offline} constrained policy iteration as diffusion noise regression under KL/BC regularization; a distributional SAC variant~\citep{ma2025dsac} improves risk sensitivity via value-law estimation and distorted expectations
(e.g., CVaR), but operates with standard unimodal Gaussian policies in an
online setting, rather than shaping a multimodal diffusion/flow actor under
behavior-regularized offline constraints; the diffusion-policy instantiation~\citep{liu2025distributional} targets multi-modality but likewise reports no CVaR along the multi-step generation. Risk-averse offline methods relying on behavior priors (e.g.,~\citep{urpi2021risk}), and diffusion-prior (e.g.,~\citep{chen2025diffusion}) use anchor–perturb/mixing mechanisms, while~\citep{ma2021conservative} imposes conservative distributional critics (value pessimism). These approaches either (i) optimize expectation-oriented objectives with expressive generators or (ii) control risk via mixing or pessimism, in contrast to our distributional risk shaping without anchor mixing.

\section{Implementation Details}
\label{app:impl_ramac}

\paragraph{Actor Architecture}
RAMAC employs a reparameterized generative actor $a=\psi_\theta(s,z)$ so that gradients from the risk term flow through the entire generative trajectory.
RADAC instantiates $\psi_\theta$ as a denoising diffusion policy with VP schedule and $T{=}5$ denoising steps; the score network is an MLP (hidden 256–256, SiLU) following~\citep{wang2022diffusion}.
RAFMAC instantiates $\psi_\theta$ as a deterministic flow–matching ODE solved by Euler with \emph{flow\_steps} $K{=}10$; the velocity field is an MLP (hidden 512–512, SiLU) following~\citep{park2025flow}.
For both, the actor objective is
$\mathcal{L}_{\pi}=\mathcal{L}_{\mathrm{BC}}+\eta\,\mathcal{L}_{\mathrm{Risk}}$ (as shown in Eq.~\ref{eq:policy_loss_full}),
where $\mathcal{L}_{\mathrm{BC}}$ is the model's native BC loss
(denoising loss for diffusion and flow-matching loss for flow),
serving as the surrogate counterpart of Eq.~\ref{eq:bc_loss}, and
$\mathcal{L}_{\mathrm{Risk}}=-\mathbb{E}_{s,a\sim\pi_\theta}[\mathrm{CVaR}_\alpha(Z_\phi(s,a))]$ with $\alpha{=}0.1$ as shown in Eq.~\ref{eq:risk_loss_keep}.
\paragraph{Distributional Critic Architecture}
As described in Sec.~\ref{sec:critic}, both RAMAC (diffusion actor) and RAFMAC (flow actor) use a Double IQN critic $Z_\phi(s,a;\tau)$ that returns the $\tau$-quantile of the return distribution.
We instantiate two critics $Z_{\phi_1},Z_{\phi_2}$ and train them with the quantile Huber loss ($\kappa{=}1$), using a standard double-critic scheme where targets are formed with a minimum over a slowly updated target copy $Z_{\bar\phi}$ to curb overestimation.

For a batch $(s,a,r,s')\!\sim\!\mathcal D$, the bootstrap action is always produced by the same generative actor used in the policy loss:
\[
a' \;=\; \psi_\theta(s',z'),\qquad z' \sim \mathcal N(0,I),
\]
so that the critic learns return quantiles under the current actor and, in turn, supplies tail-sensitive gradients (via CVaR) to the risk-aware actor update (Step~2, Sec.~\ref{sec:risk_aware_actor}). Concretely, one can view the temporal-difference (TD) objective as approximating expectations over $\tau,\tau'\!\sim\!\mathcal U(0,1)$, while the actor-side CVaR objective uses $\tau\!\sim\!\mathcal U(0,\alpha)$.

In implementation, we replace stochastic quantile sampling with
midpoint quadrature on fixed uniform grids. This provides a
deterministic approximation to the relevant quantile integrals and
removes Monte Carlo sampling variance for a fixed critic. Instead of drawing $\tau$ at random, we use
\begin{equation}
\label{eq:det_iqn_grid}
\mathcal T_N \;=\; \Big\{\,\tau_i \;=\; \tfrac{i-\tfrac12}{N}\,\Big\}_{i=1}^{N},
\end{equation}
and approximate $\mathrm{CVaR}_\alpha$ by averaging the lowest-$\alpha$ fraction of quantiles. For CVaR at level $\alpha$, let $m \!=\! \lfloor \alpha N\rfloor$; then
\begin{equation}
\label{eq:det_cvar}
\mathrm{CVaR}_\alpha\!\big(Z_\phi(s,a)\big)
\;\approx\; \frac{1}{m}\sum_{i=1}^{m} Z_\phi\!\big(s,a;\tau_i\big),
\qquad \tau_i\in \mathcal T_N.
\end{equation}
This is equivalent in expectation to drawing $\tau\!\sim\!\mathcal U(0,\alpha)$ (Eq.~\ref{eq:cvar_def_keep}), but the deterministic grid substantially reduces variance in both the critic loss and the actor’s CVaR objective.

For the TD targets we use another grid $\mathcal T_{N'}$ and define the residual for each pair of source and target quantiles as
\[
\delta_{\tau_i,\tau'_j}
\;=\;
r + \gamma\,Z_{\bar\phi}(s',a';\tau'_j)\;-\;Z_\phi(s,a;\tau_i),
\qquad
\tau_i\!\in\!\mathcal T_{N},\ \tau'_j\!\in\!\mathcal T_{N'}.
\]
The critic minimizes the standard quantile Huber loss
\citep{dabney2018implicit,rowland2019statistics}
\begin{equation}
\label{eq:huber_loss}
\mathcal{L}_\kappa(\delta;\tau)
  = \bigl|\tau-\mathbf{1}_{\{\delta<0\}}\bigr|
    \times
    \begin{cases}
      \tfrac{\delta^{2}}{2\kappa}, & |\delta|\le\kappa,\\[4pt]
      |\delta|-\tfrac{\kappa}{2},   & \text{otherwise},
    \end{cases}
\end{equation}
with $\kappa\!=\!1$.
Averaging over all $N\times N'$ quantile pairs yields the final critic objective
\begin{equation}
  \mathcal{L}_{\text{critic}}(\phi)=
  \mathbb{E}_{(s,a,r,s'),\,a'}\!
  \biggl[
    \frac{1}{NN'}\!
    \sum_{i=1}^{N}\sum_{j=1}^{N'}
       \mathcal{L}_\kappa\bigl(\delta_{\tau_i,\tau'_j};\,\tau_i\bigr)
  \biggr].
  \label{eq:critic_loss_final}
\end{equation}
Thus Eq.~\ref{eq:critic_loss_final} is exactly the compact critic objective in Sec.~\ref{sec:critic}, with deterministic grids $(\tau_i,\tau'_j)$ used in place of stochastic $(\tau,\tau')$.
Optimising this loss yields an estimate of the return distribution; its lower tail is then aggregated into $\mathrm{CVaR}_\alpha$ to provide the tail-aware gradients used in the RAMAC actor loss (Step~2).

\paragraph{Hyperparameters}
Unless noted, we use Adam for all networks (default $3{\times}10^{-4}$), batch size 256, discount $\gamma{=}0.99$, soft target update $\tau_{\text{target}}{=}0.005$, and no LR decay.
RAMAC’s specific hyperparameters (critic LR, IQN size, $\eta$, gradient–norm clipping, optional $Q$–target clipping, etc.) are listed in Table~\ref{tab:ramac_key_hparams}.
\paragraph{RAFMAC Risk-Weight Tuning} For RAFMAC, we swept $\eta\in\{1,10,50,100,300,1000\}$ and \emph{unified to $\eta{=}1000$} for all datasets; critic settings are fixed ($lr_{\text{critic}}{=}3{\times}10^{-4}$, \texttt{emb\_dim}$=128$, \texttt{n\_quantiles}$=32$) (Table~\ref{tab:ramac_key_hparams}).
\paragraph{Critic–Target Clipping}
Where specified, target returns are clipped ($[-300,300]$ or $[-150,150]$) to dampen rare outliers without affecting on–manifold learning.

\section{Additional Experimental Results}
\label{app:additional_experimental_results} 
\subsection{More 2D Synthetic Task Results}
\label{app:additional_toy}
\begin{figure}[t]
  \centering
  \begin{subfigure}[b]{.21\textwidth}
    \caption{Ground Truth}
      \fbox{\includegraphics[width=\linewidth]{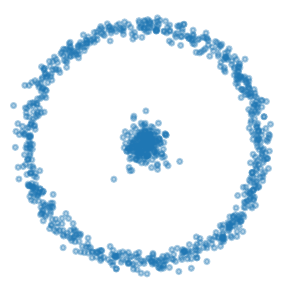}}
  \end{subfigure}\hfill
  \begin{subfigure}[b]{.21\textwidth}
        \caption{CVAE-BC}
      \fbox{\includegraphics[width=\linewidth]{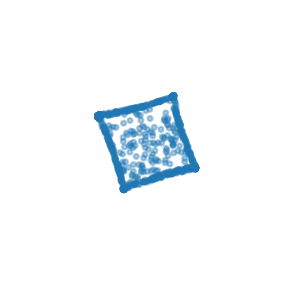}}
  \end{subfigure}\hfill
  \begin{subfigure}[b]{.21\textwidth}
    \caption{Diffusion-BC}
      \fbox{\includegraphics[width=\linewidth]{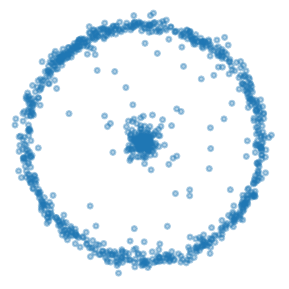}}
  \end{subfigure}\hfill
  \begin{subfigure}[b]{.21\textwidth}
        \caption{Flow Matching-BC}
      \fbox{\includegraphics[width=\linewidth]{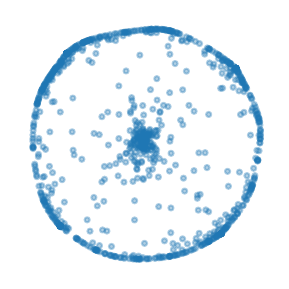}} 
  \end{subfigure}

  \vspace{-0.6em}
  % 1) Behavior Cloning (CVAE / Diffusion / Flow)
\caption{\textbf{Behavior cloning on the Toy Risky Bandit dataset.}
Each panel shows i.i.d.\ samples from the BC Policy.
CVAE‑BC mixes modes and places points in the low‑density gap; Diffusion‑BC reproduces both the outer ring and the central cluster; Flow‑Matching BC yields a crisp ring but assigns less mass to the center.}

  \label{fig:bc_results}
\end{figure}

\begin{figure}[t]
  \centering
  
  \begin{subfigure}[b]{.16\textwidth}
    \caption{epoch=0}
    \fbox{\includegraphics[width=\linewidth]{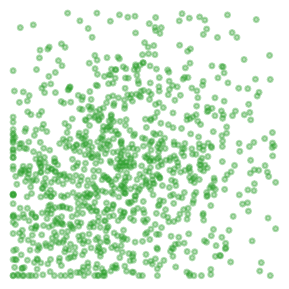}}
  \end{subfigure}\hfill
  \begin{subfigure}[b]{.16\textwidth}
    \caption{epoch=200}  \fbox{\includegraphics[width=\linewidth]{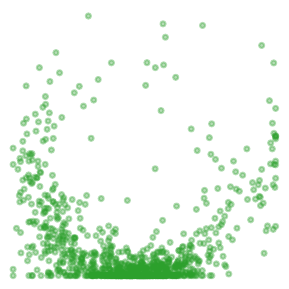}}
  \end{subfigure}\hfill
  \begin{subfigure}[b]{.16\textwidth}
        \caption{epoch =400}\fbox{\includegraphics[width=\linewidth]{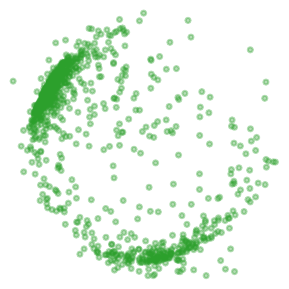}} 
  \end{subfigure}\hfill
  \begin{subfigure}[b]{.16\textwidth}
        \caption{epoch=600}
        \fbox{\includegraphics[width=\linewidth]{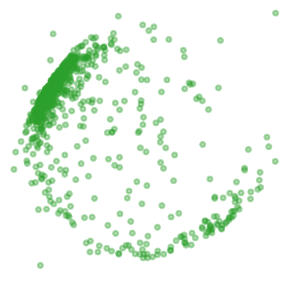}} 
  \end{subfigure}\hfill
  \begin{subfigure}[b]{.16\textwidth}
        \caption{epoch=800}
      \fbox{\includegraphics[width=\linewidth]{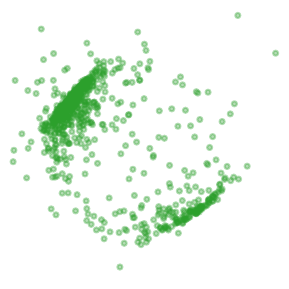}}
  \end{subfigure}\hfill
  \begin{subfigure}[b]{.16\textwidth}
        \caption{epoch=950}
      \fbox{\includegraphics[width=\linewidth]{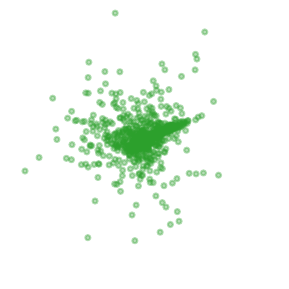}}
  \end{subfigure}\hfill
  \vspace{-0.6em}
\caption{\textbf{RADAC dynamics on the toy task.}
Mass gradually moves from the risky ring to the safe center: the ring thins (400–800 epochs) and the central cluster grows, ending with most mass at the center (~950 epochs). BC keeps the policy on-manifold while CVaR reduces lower-tail risk.}
  \label{fig:radac_transitions}
\end{figure}

\paragraph{Behavior cloning task Fig.~\ref{fig:bc_results}}
On the 2D bandit dataset, three BC models show generator-specific patterns.
CVAE-BC collapses topology and places probability in the low‑density gap.
Diffusion-BC most faithfully reproduces both the ring and the inner cluster with appropriate thickness.
Flow-Matching BC draws a sharp ring but allocates less mass to the center and shows edges spread slightly outward. These baselines confirm that a suitably trained generative model can
represent the full multimodal support of the dataset.
\paragraph{RADAC dynamics over training Fig.~\ref{fig:radac_transitions}}
RADAC starts from the same diffusion generator that is first fit to the
behavior data, and then applies CVaR-based policy updates on top of the
BC objective.
In Fig.~\ref{fig:radac_transitions} we fix the RL weight $\eta$ to the
value used in Fig.~\ref{fig:toy_rl} and visualize how the induced
policy evolves over training.

Early in training $\sim$200 epochs, the critic has not yet identified the outer ring as
hazardous: high-mean returns on the ring dominate the low-quantile
signal, so the CVaR-guided updates still allocate substantial mass to
the high-return ring while also covering the safe central cluster.
As learning proceeds and low-quantile returns on the ring are observed,
CVaR increasingly penalizes ring actions; between about 400 and
800 epochs the ring thins and probability mass shifts inward as the
central cluster grows.
By roughly 950 epochs most mass concentrates near the safe center,
matching the final snapshot in Fig.~\ref{fig:toy_rl}.

\begin{figure}[t]
  \centering
  \includegraphics[width=\linewidth]{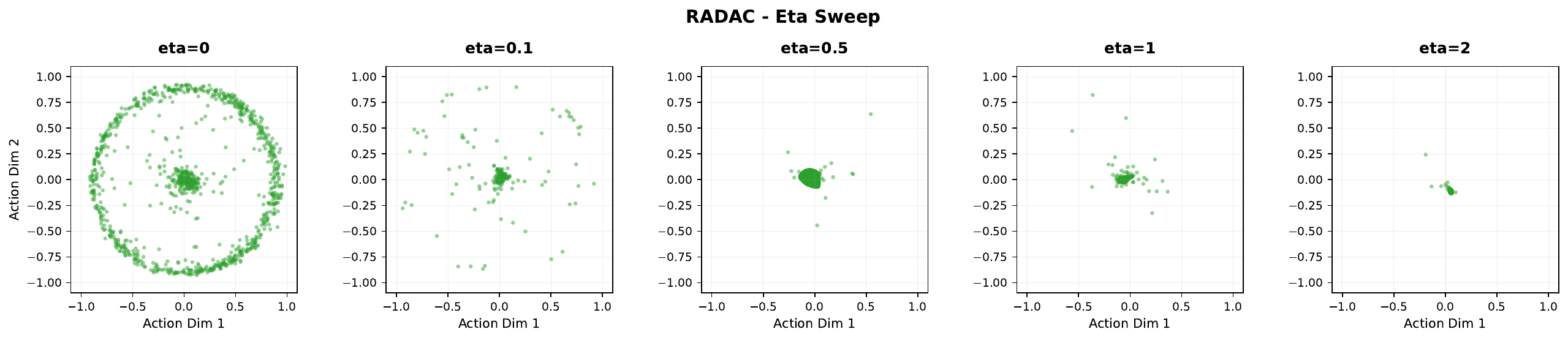}
  \caption{\textbf{Effect of RL weight $\eta$ in RADAC.}
  As $\eta$ increases from $0$ (pure BC), probability mass smoothly moves from the risky outer ring to the safe central cluster, without creating off-support modes.}
  \label{fig:radac_eta_sweep}
\end{figure}

This trajectory highlights that RADAC uses the expressive diffusion
policy to first capture the multimodal behavior distribution and then
\emph{reallocate} mass toward the high-CVaR central mode once its tail
risk is correctly estimated, rather than collapsing due to limited
capacity.
The explicit BC-risk trade-off as the RL weight $\eta$ varies is
further analyzed in Fig.~\ref{fig:radac_eta_sweep}.
\paragraph{On conservatism and hyperparameters in prior baselines.}
The same Risky–Bandit geometry also helps explain why conservative
tuning of prior methods (DiffusionQL, FQL, ORAAC-style
anchor-perturbation) does not eliminate their structural failure
modes.
As defined in App.~\ref{app:synthetic_details}, behavior actions lie on
a thin, high-mean but heavy-tailed outer ring and a lower-mean but
light-tailed central Gaussian cluster.
Geometrically, each behavior anchor $b$ lies on a narrow manifold with
margin $m(b)$ to the low-density gap and to regions that are unobserved
in the dataset.
Once a perturbation ball $\mathbb{B}_\varepsilon(b)$ has
$\varepsilon \gtrsim m(b)$, it necessarily crosses into these
off-support regions.
In this regime, residual or anchor-perturbation policies are not
constrained to stay tangent to the data manifold, so even small
residuals can systematically produce OOD actions, which in turn leads
to unstable Q-estimates and higher exposure to the trap-heavy ring.
This is precisely the behavior illustrated in Fig.~\ref{fig:toy_rl}(f-g).

\begin{figure}[t]
  \centering
  \subfloat[DiffusionQL: sweep over $\eta$]{
    \includegraphics[width=\linewidth]{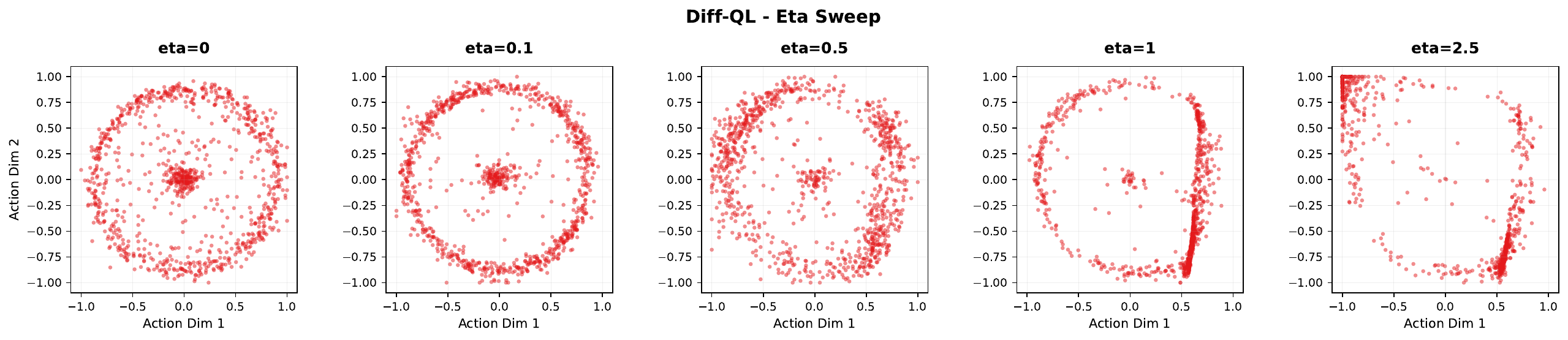}
  }\hfill
  \subfloat[FQL: sweep over $\eta$]{
    \includegraphics[width=\linewidth]{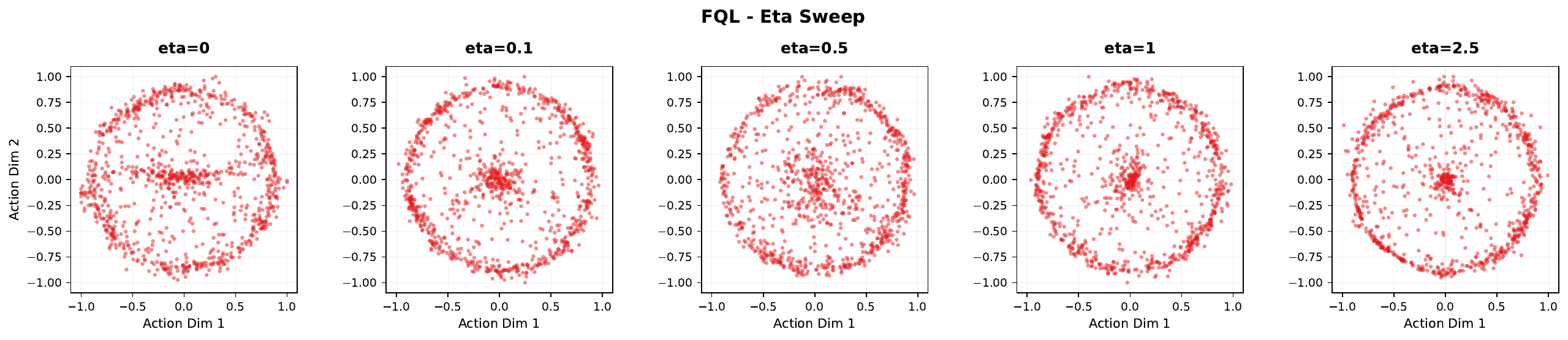}
  }
  \caption{\textbf{Effect of RL weight $\eta$ in Toy Risky Bandit.}
  Each panel shows a policy trained on the 2D Toy Risky Bandit with a different
  RL scaling coefficient $\eta$.
  As $\eta \to 0$, especially DiffusionQL reverts to BC-like behavior and
  recovers the multimodal behavior density.
  As $\eta$ increases, probability mass concentrates on the high-mean outer
  ring despite its heavy lower tail, illustrating the risk-neutral tendency
  discussed in Sec.~\ref{sec:method} and App.~\ref{app:additional_experimental_results}.}
  \label{fig:toy_eta_sweep}
\end{figure}

For DiffusionQL and FQL, the RL term is always driven by a
risk-neutral Q-function.
Sweeping the RL scaling coefficient $\eta$ (as shown in Fig.~\ref{fig:toy_eta_sweep}) shows the
expected dichotomy:
as $\eta \to 0$ the policies revert to BC-like behavior, recovering
the multimodal density of the behavior data;
as $\eta$ increases, mass increasingly concentrates on the high-mean
ring, despite its heavy lower tail.
FQL’s flow-matching prior retains a bit more mass near the center,
but the dominant mode still tracks the risky ring.
No choice of $\eta$ can simultaneously prevent this tendency and retain
a nontrivial RL component, because the objective is fundamentally
risk-neutral.

\begin{figure}[t]
  \centering
  \subfloat[ORAAC: sweep over $\lambda$]{
    \includegraphics[width=\linewidth]{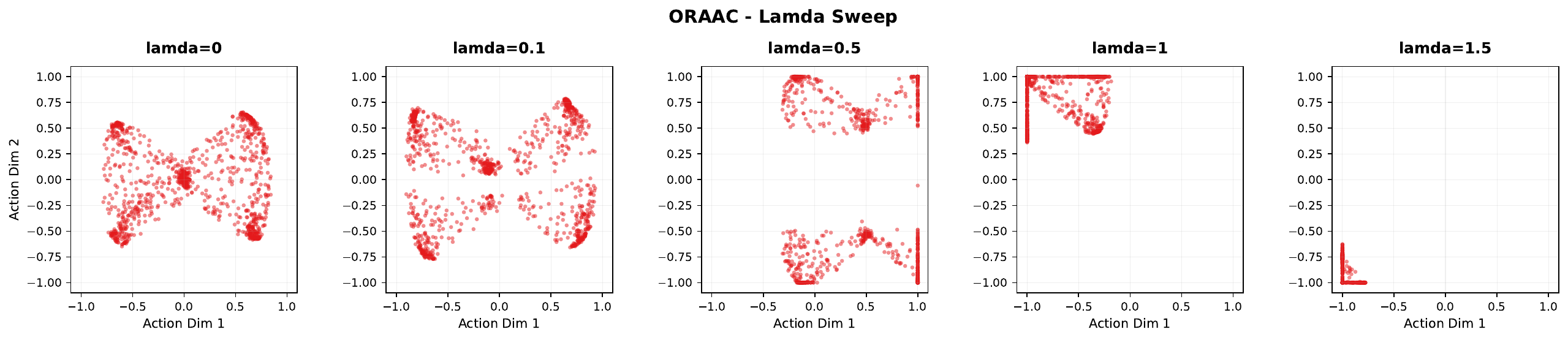}
  }\hfill
  \subfloat[ORAAC-Diff.: sweep over $\lambda$]{
    \includegraphics[width=\linewidth]{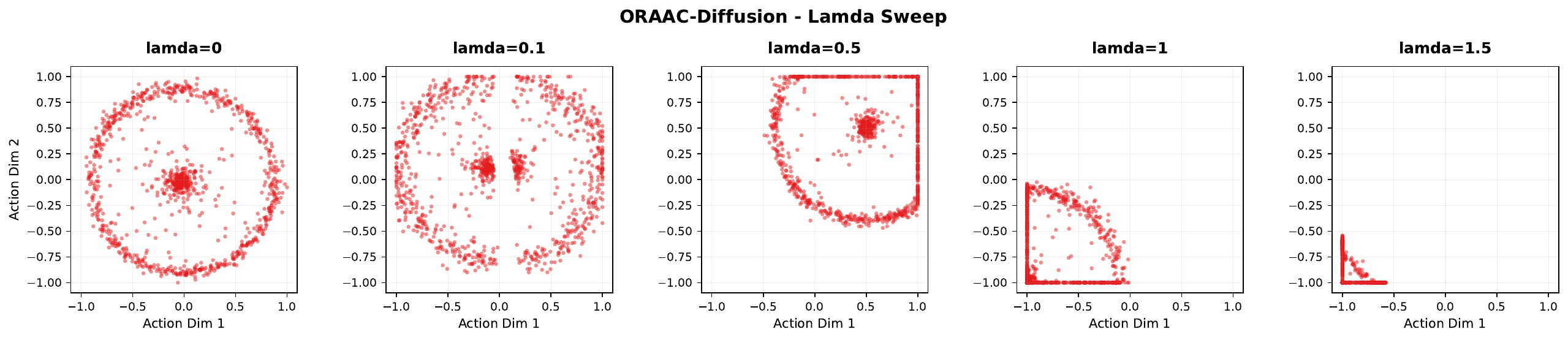}
  }\hfill
  \subfloat[ORAAC--Flow: sweep over $\lambda$]{
    \includegraphics[width=\linewidth]{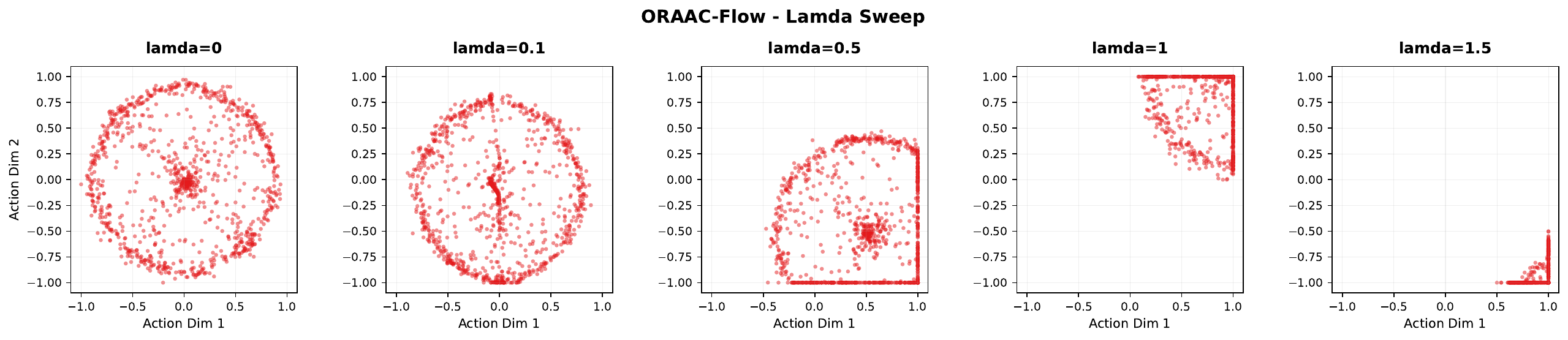}
  }
  \caption{\textbf{Effect of risk weight $\lambda$ in anchor-perturbation baselines.}
  Each panel visualizes policies on the 2D Toy Risky Bandit as the
  risk-distortion weight $\lambda$ is swept.
  For $\lambda=0$, all ORAAC variants reduce to BC on the behavior manifold.
  As $\lambda$ increases, residual updates push probability mass away from the
  BC solution: modes can split or leak through low-density gaps, and for
  larger $\lambda$ the policy may jump toward regions far from any behavior
  support, consistent with the analysis in Sec.~\ref{sec:risk_pitfalls}.}
  \label{fig:toy_lambda_sweep}
\end{figure}

{Fig.~\ref{fig:toy_lambda_sweep} shows ORAAC-style anchor-perturbation exhibits an analogous tradeoff.
When the risk-distortion weight $\lambda$ is set to zero, the method
again collapses to BC on the behavior manifold.
As $\lambda$ increases, the residual pushes probability mass away from
the BC solution: in the Risky–Bandit geometry this either splits modes
or routes probability through low-density regions between them, and for
larger $\lambda$ the policy can “jump’’ toward arbitrary corners of the
action space, far from any behavior support (as predicted by the
analysis in Sec.~\ref{sec:risk_pitfalls}).}

In other words, the observed failures of Fig.~\ref{fig:toy_rl}(c-g) are not artifacts
of insufficient hyperparameter tuning but a structural consequence of
combining risk-neutral or anchor-perturbation objectives with thin,
non-convex data supports.
RADAC replaces this with an on-manifold, risk-aware update that
reduces tail risk without relying on off-support residuals.

For completeness, we also sweep the RL weight $\eta$ in RADAC on the same Risky–Bandit (Fig.~\ref{fig:radac_eta_sweep}).
When $\eta \to 0$, the objective reduces to pure BC and the diffusion policy faithfully matches the multimodal behavior distribution, reproducing both the outer ring and the central cluster.
As $\eta$ increases, the CVaR-guided term gradually reallocates probability mass away from the high-mean but heavy-tailed ring toward the light-tailed central mode while staying on-support: the ring thins and eventually disappears, leaving a compact cluster around the safe center.
Unlike DiffusionQL/FQL and ORAAC-style anchor–perturbation, this sweep does not create spurious off-support modes or low-density bridges; instead, it yields a smooth trade-off between ring coverage and tail-risk reduction, consistent with the on-manifold, risk-aware behavior-regularized objective in Eq.~\ref{eq:policy_loss_full}.

% \begin{figure}[t]
%   \centering
%   \includegraphics[width=0.9\linewidth]{toy_comparison.png}
% \caption{\textbf{More toy results.}
% The qualitative pattern is unchanged across seeds; we show another seed.
% RAFMAC’s thinner ring and inward shift arise from CVaR‑shaped flow transport under BC.}
%   \label{fig:toy_complete}
% \end{figure}

\begin{table}[t]
  \centering
  \scriptsize
  \setlength{\tabcolsep}{2pt}
  \renewcommand{\arraystretch}{1.0}
  \caption{Stochastic-D4RL (1000-step evaluation): Mean and CVaR$_{0.1}\!\pm$ s.e.\ over 5 seeds.}
  \label{tab:sd4rl_results_1000_by_dataset}

  \begin{tabular}{llcc}
    \toprule
    \textbf{Environment, Dataset} & \textbf{Algorithm} & \textbf{Mean} & \textbf{CVaR} \\
    \midrule
    \multirow{7}{*}{HalfCheetah-m-e}
      & CQL         & $-0.97\!\pm\!0.24$      & $-2.24\!\pm\!0.43$ \\
      & CODAC       & $-0.12\!\pm\!0.08$      & $-1.48\!\pm\!0.27$ \\
      & ORAAC       & $4106.25\!\pm\!177.48$  & $3692.79\!\pm\!466.31$ \\
      & FQL      & $4695.46\!\pm\!65.97$   & $4025.12\!\pm\!230.08$ \\
      & DiffusionQL & $-118.72\!\pm\!64.53$   & $-198.01\!\pm\!76.76$ \\
      & RAFMAC      & $5084.12\!\pm\!230.43$  & $3735.37\!\pm\!827.60$ \\
      & RADAC       & $5659.40\!\pm\!131.94$  & $4667.96\!\pm\!42.59$ \\
    \midrule
    \multirow{7}{*}{Walker2d-m-e}
      & CQL         & $-10.32\!\pm\!6.27$     & $-73.38\!\pm\!9.02$ \\
      & CODAC       & $27.56\!\pm\!6.26$      & $-35.30\!\pm\!15.36$ \\
      & ORAAC       & $663.23\!\pm\!181.31$   & $205.21\!\pm\!65.45$ \\
      & FQL      & $2457.68\!\pm\!208.80$  & $448.48\!\pm\!208.81$ \\
      & DiffusionQL & $-32.33\!\pm\!4.59$     & $-68.43\!\pm\!11.28$ \\
      & RAFMAC      & $3567.89\!\pm\!206.63$  & $356.20\!\pm\!987.34$ \\
      & RADAC       & $2760.21\!\pm\!689.32$  & $322.76\!\pm\!757.44$ \\
    \midrule
    \multirow{7}{*}{Hopper-m-e}
      & CQL         & $43.22\!\pm\!29.48$     & $-65.90\!\pm\!36.42$ \\
      & CODAC       & $31.59\!\pm\!28.74$     & $-77.88\!\pm\!34.33$ \\
      & ORAAC       & $660.07\!\pm\!157.55$   & $400.84\!\pm\!142.60$ \\
      & FQL      & $393.64\!\pm\!27.75$    & $77.93\!\pm\!60.53$ \\
      & DiffusionQL & $-38.75\!\pm\!27.68$    & $-212.49\!\pm\!91.99$ \\
      & RAFMAC      & $370.11\!\pm\!39.95$    & $-120.09\!\pm\!56.34$ \\
      & RADAC       & $1513.27\!\pm\!101.71$  & $967.21\!\pm\!160.68$ \\
    \midrule
    \multirow{7}{*}{HalfCheetah-m-r}
      & CQL         & $-38.85\!\pm\!38.44$    & $-40.23\!\pm\!38.44$ \\
      & CODAC       & $-0.12\!\pm\!0.08$      & $-1.48\!\pm\!0.26$ \\
      & ORAAC       & $315.87\!\pm\!69.27$    & $161.54\!\pm\!68.76$ \\
      & FQL      & $1909.57\!\pm\!395.55$  & $568.43\!\pm\!256.85$ \\
      & DiffusionQL & $2261.16\!\pm\!531.18$  & $1439.77\!\pm\!461.28$ \\
      & RAFMAC      & $2696.61\!\pm\!110.68$  & $1499.80\!\pm\!394.08$ \\
      & RADAC       & $2674.72\!\pm\!51.76$   & $1401.03\!\pm\!199.08$ \\
    \midrule
    \multirow{7}{*}{Walker2d-m-r}
      & CQL         & $-14.68\!\pm\!5.52$     & $-95.30\!\pm\!18.50$ \\
      & CODAC       & $26.39\!\pm\!7.97$      & $-36.56\!\pm\!12.92$ \\
      & ORAAC       & $160.23\!\pm\!147.55$   & $-359.49\!\pm\!302.72$ \\
      & FQL      & $647.33\!\pm\!166.12$   & $-29.64\!\pm\!110.73$ \\
      & DiffusionQL & $-23.50\!\pm\!4.44$     & $-53.55\!\pm\!12.30$ \\
      & RAFMAC      & $778.00\!\pm\!130.03$   & $7.92\!\pm\!35.77$ \\
      & RADAC       & $383.87\!\pm\!288.95$   & $-309.70\!\pm\!246.62$ \\
    \midrule
    \multirow{7}{*}{Hopper-m-r}
      & CQL         & $2.28\!\pm\!42.17$      & $-130.48\!\pm\!53.25$ \\
      & CODAC       & $3.61\!\pm\!18.41$      & $-105.41\!\pm\!19.86$ \\
      & ORAAC       & $-30.00\!\pm\!32.77$    & $-179.92\!\pm\!61.46$ \\
      & FQL      & $448.26\!\pm\!70.39$    & $-33.21\!\pm\!43.38$ \\
      & DiffusionQL & $-22.15\!\pm\!24.93$    & $-163.82\!\pm\!59.18$ \\
      & RAFMAC      & $350.36\!\pm\!33.05$    & $-36.69\!\pm\!28.35$ \\
      & RADAC       & $453.64\!\pm\!68.46$    & $-87.04\!\pm\!123.96$ \\
    \bottomrule
  \end{tabular}
\end{table}
\begin{figure}[t]

  \centering
  % ---------- 1) RADAC ----------
  \begin{subfigure}[t]{.18\textwidth}
    \begin{tikzpicture}
      \node[inner sep=0pt] (img)
            {\includegraphics[trim={0cm 0cm 0 0cm}, clip,width=1.1\linewidth]{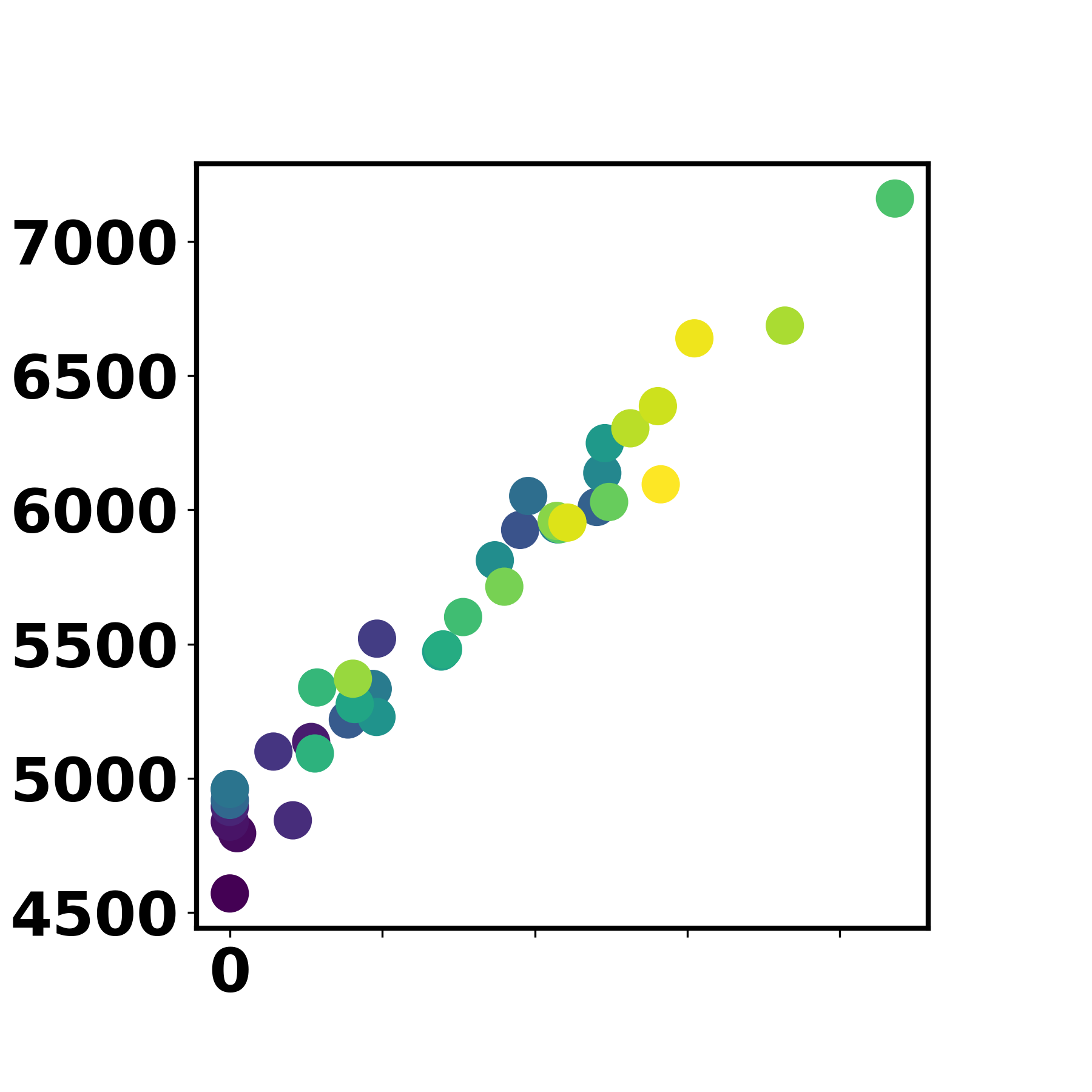}};
      %  (Mean Reward)
      \node[rotate=90, anchor=center]
            at ($(img.south west)!0.5!(img.north west)+(-8pt,0)$)
            {\footnotesize Mean Reward};
    \node[font=\scriptsize,
          anchor=north,        
          inner sep=0pt, outer sep=0pt,
          text height=1ex, text depth=0pt]
      at ($(img.north)+(0,-1.2ex)$) {HalfCheetah-m-e};
    \end{tikzpicture}
  \end{subfigure}
  \hspace{0.035\textwidth} 
  % ---------- 2) Diffusion QL ----------
  \begin{subfigure}[t]{.18\textwidth}
    \includegraphics[trim={0cm 0cm 0 0cm}, clip,width=1.1\linewidth]{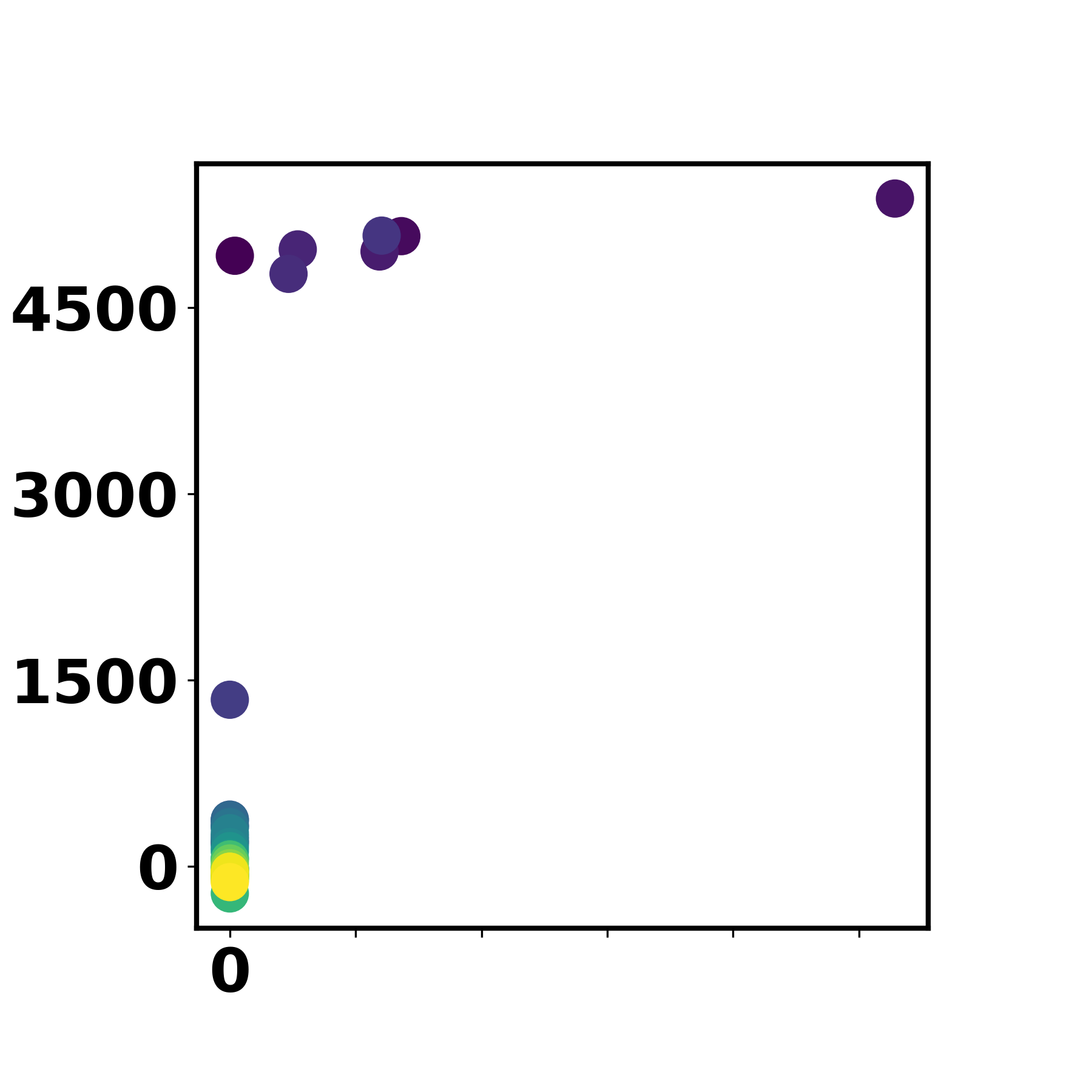}
  \end{subfigure}
  \begin{subfigure}[t]{.18\textwidth}
    \includegraphics[trim={0cm 0cm 0 0cm}, clip,width=1.1\linewidth]{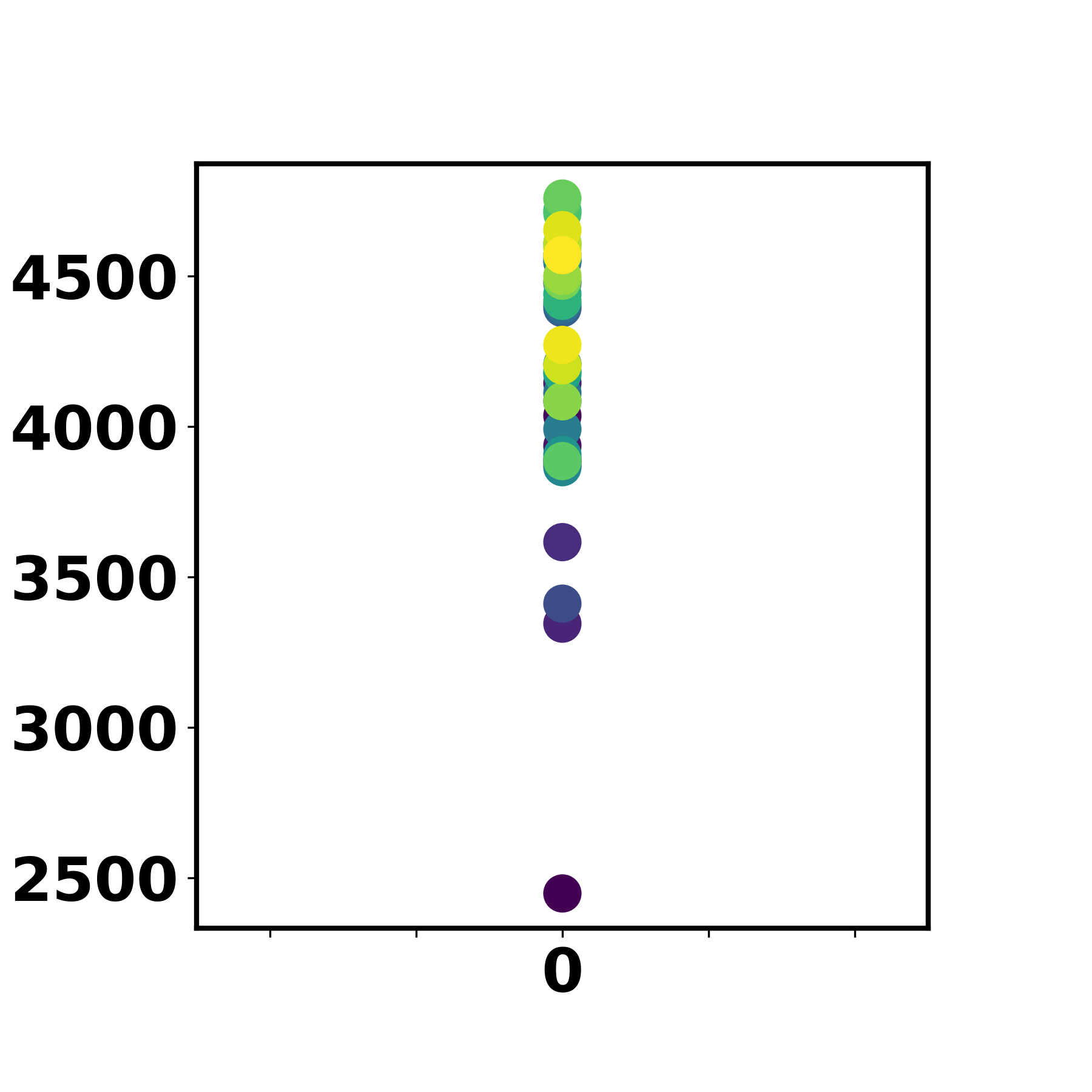}
    \end{subfigure}
  \begin{subfigure}[t]{.18\textwidth}
    \includegraphics[width=1.1\linewidth]{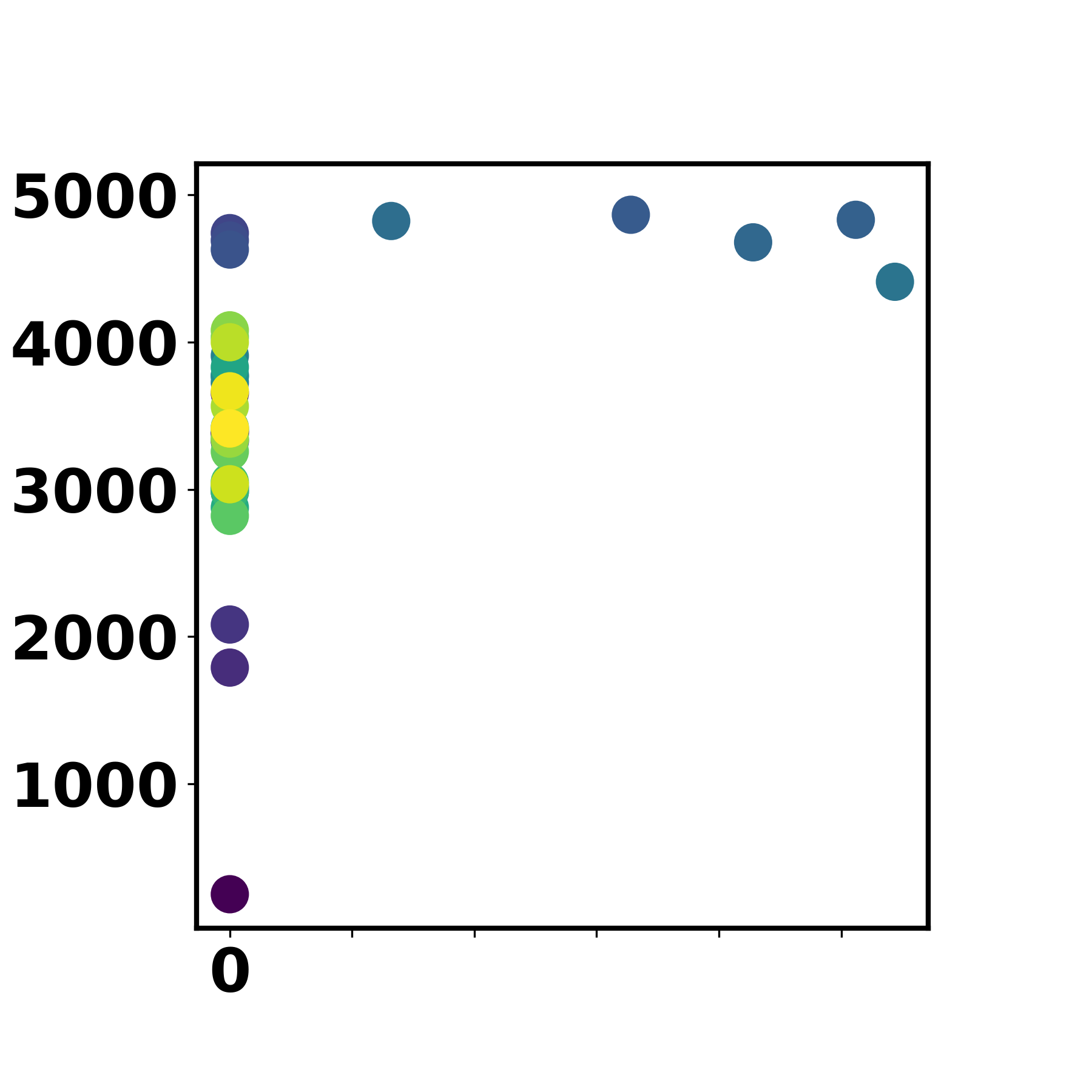}
  \end{subfigure}
  \begin{subfigure}[t]{.18\textwidth}
    \includegraphics[trim={2cm 0cm 0 0cm}, clip,width=1.5\linewidth]{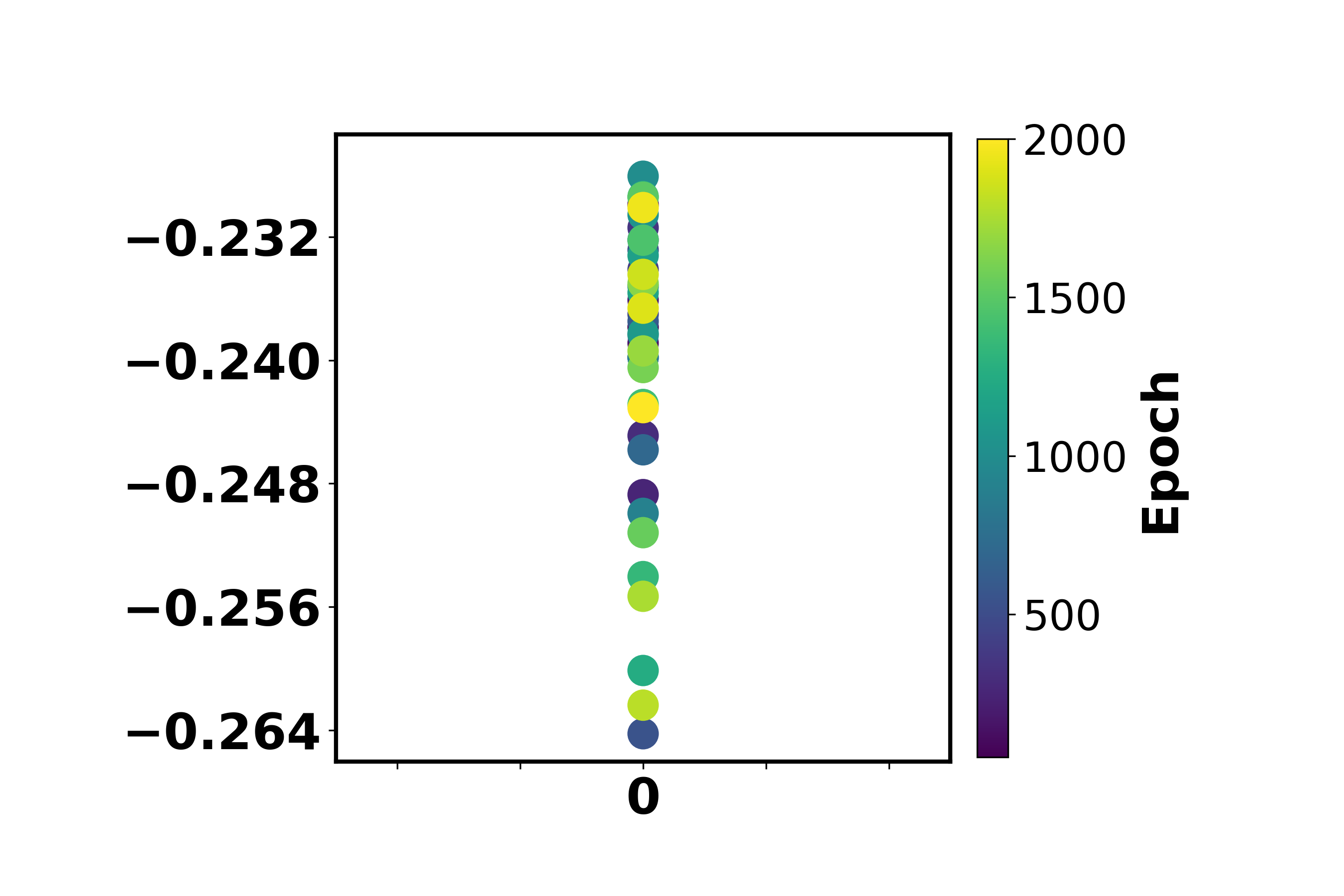}
  \end{subfigure}

    \hspace{1.2em}
    \begin{subfigure}[t]{.18\textwidth}
  \begin{tikzpicture}
    \node[inner sep=0pt, outer sep=0pt] (img)
      {\includegraphics[trim={0cm 0cm 0cm 2cm}, clip, width=1.1\linewidth]{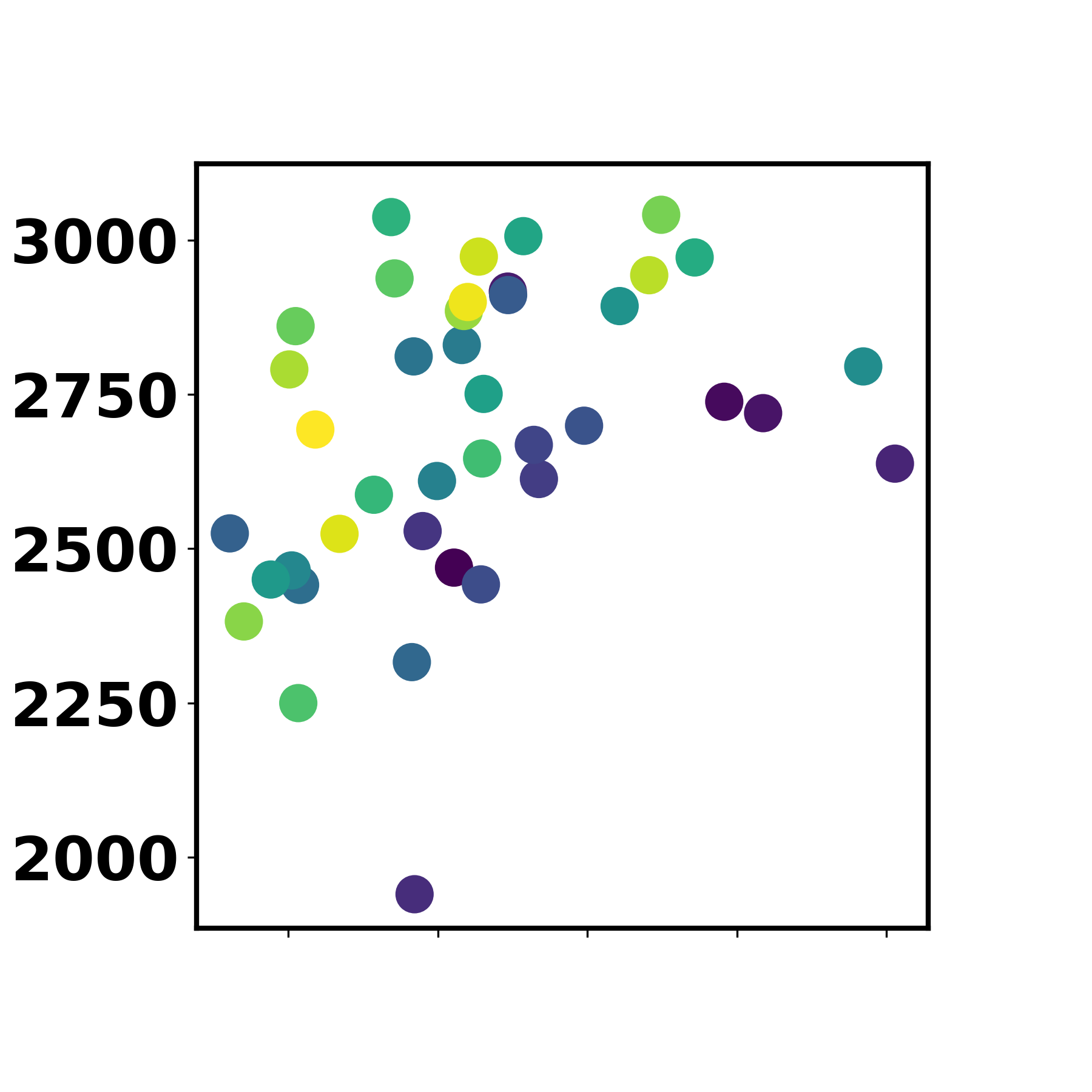}};
    \node[font=\scriptsize, anchor=north, inner sep=0pt, outer sep=0pt,
          text height=1ex, text depth=0pt]
      at ($(img.north)+(0,+1.0ex)$) {HalfCheetah-m-r};
  \end{tikzpicture}
\end{subfigure}
  \begin{subfigure}[t]{.18\textwidth}
    \includegraphics[trim={0cm 0cm 0cm 2cm}, clip,width=1.1\linewidth]{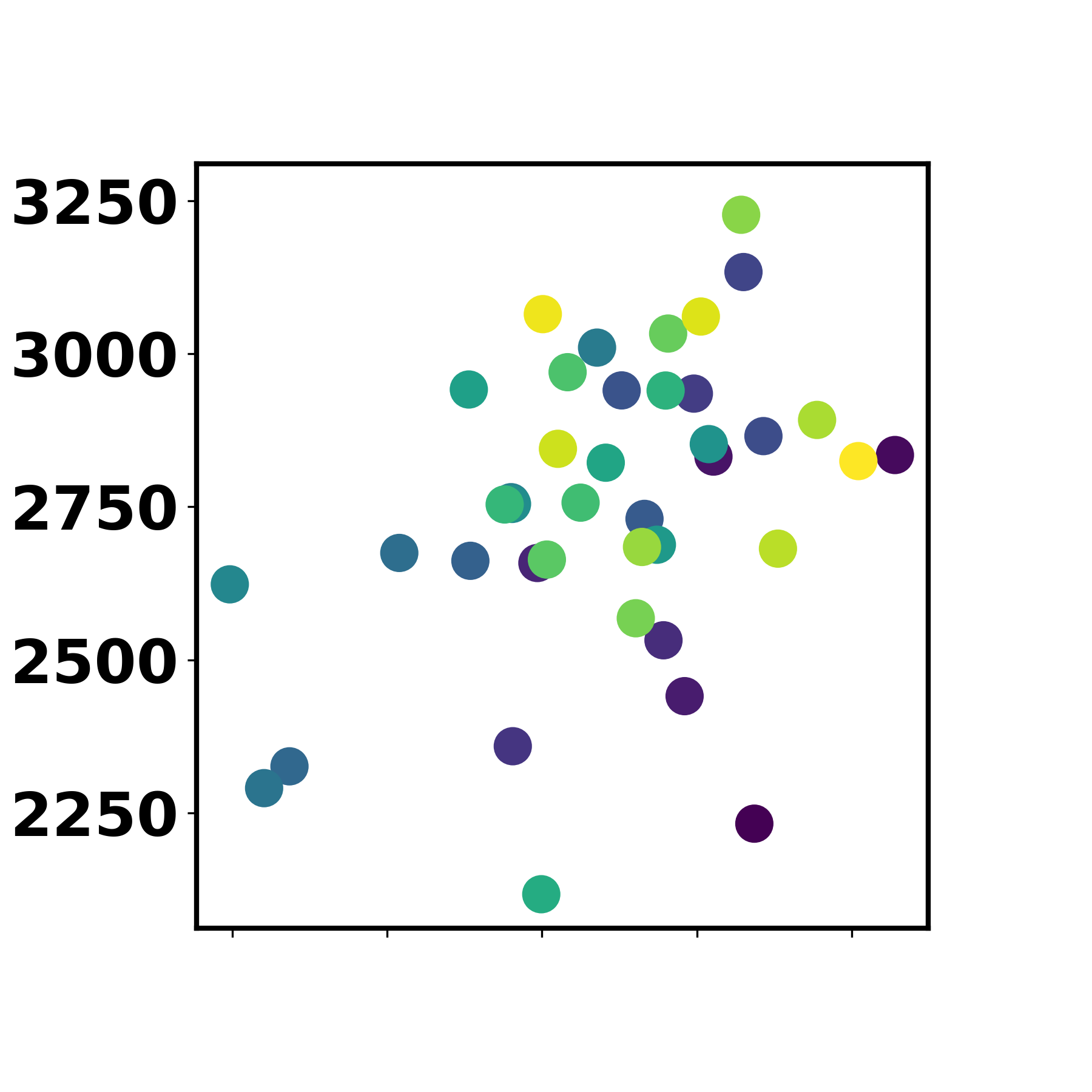}
  \end{subfigure}
  \begin{subfigure}[t]{.18\textwidth}
    \includegraphics[trim={0cm 0cm 0cm 2cm}, clip,width=1.1\linewidth]{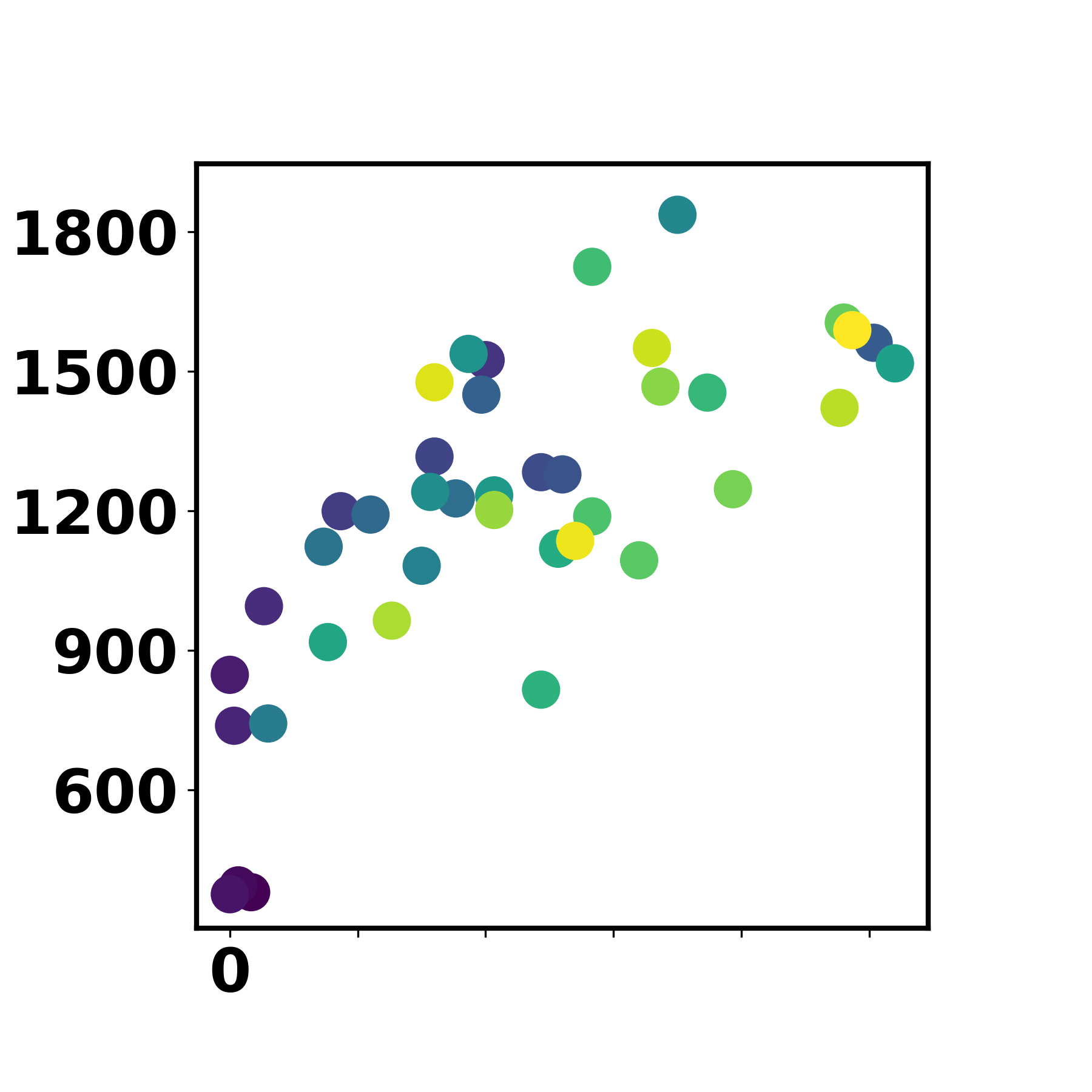}
    \end{subfigure}
  \begin{subfigure}[t]{.18\textwidth}
    \includegraphics[trim={0cm 0cm 0cm 2cm}, clip,width=1.1\linewidth]{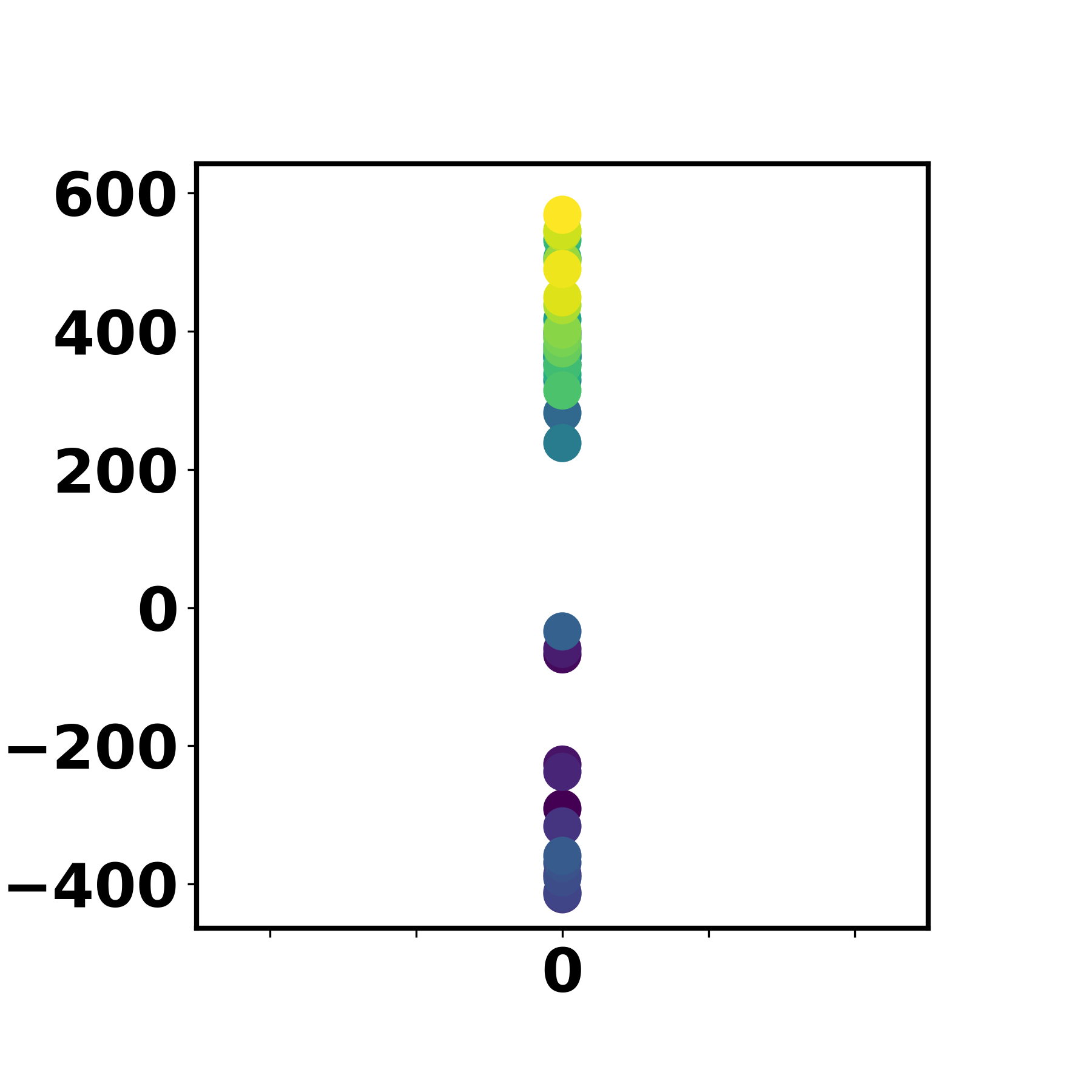}
  \end{subfigure}
  \begin{subfigure}[t]{.18\textwidth}
    \includegraphics[trim={2cm 0cm 0 2cm}, clip,width=1.5\linewidth]{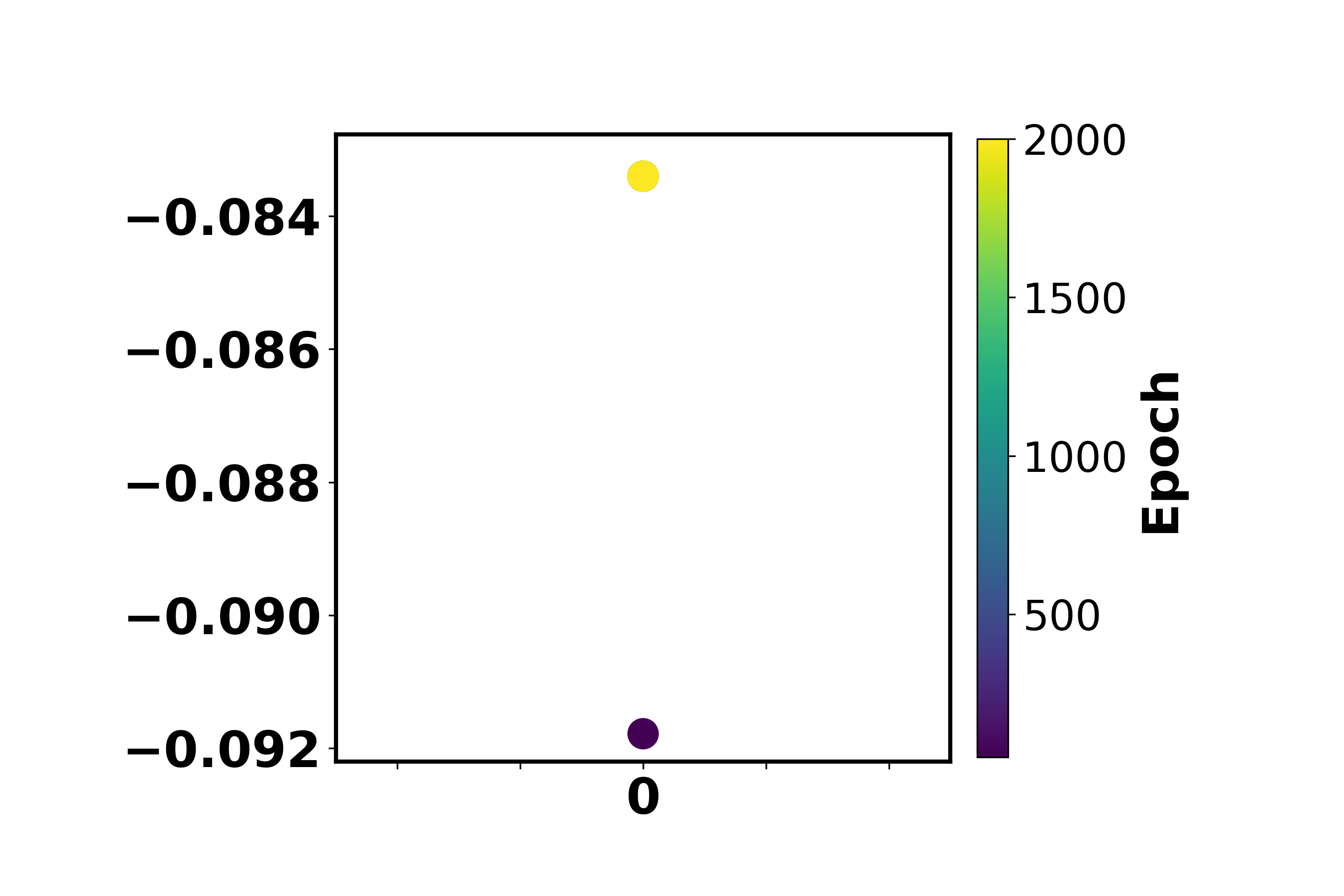}
  \end{subfigure}
  
  \hspace{1.2em}
\begin{subfigure}[t]{.18\textwidth}
  \begin{tikzpicture}
    \node[inner sep=0pt, outer sep=0pt] (img)
      {\includegraphics[trim={0cm 0cm 0 2cm}, clip, width=1.1\linewidth]{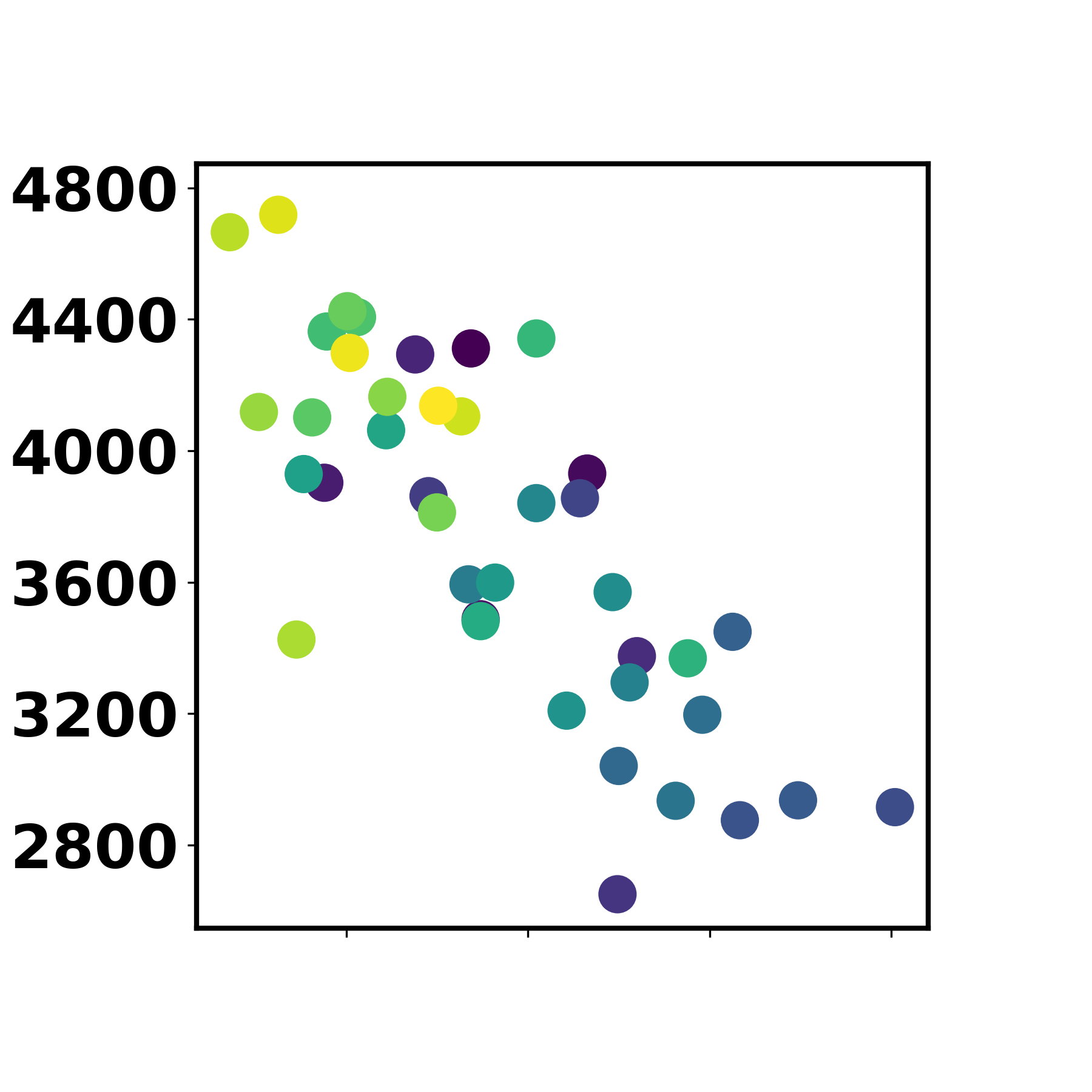}};
    \node[font=\scriptsize, anchor=north, inner sep=0pt, outer sep=0pt,
          text height=1ex, text depth=0pt]
      at ($(img.north)+(0,1.0ex)$) {Walker2d-m-e};
  \end{tikzpicture}
\end{subfigure}
  \begin{subfigure}[t]{.18\textwidth}
    \includegraphics[trim={0cm 0cm 0 2cm}, clip,width=1.1\linewidth]{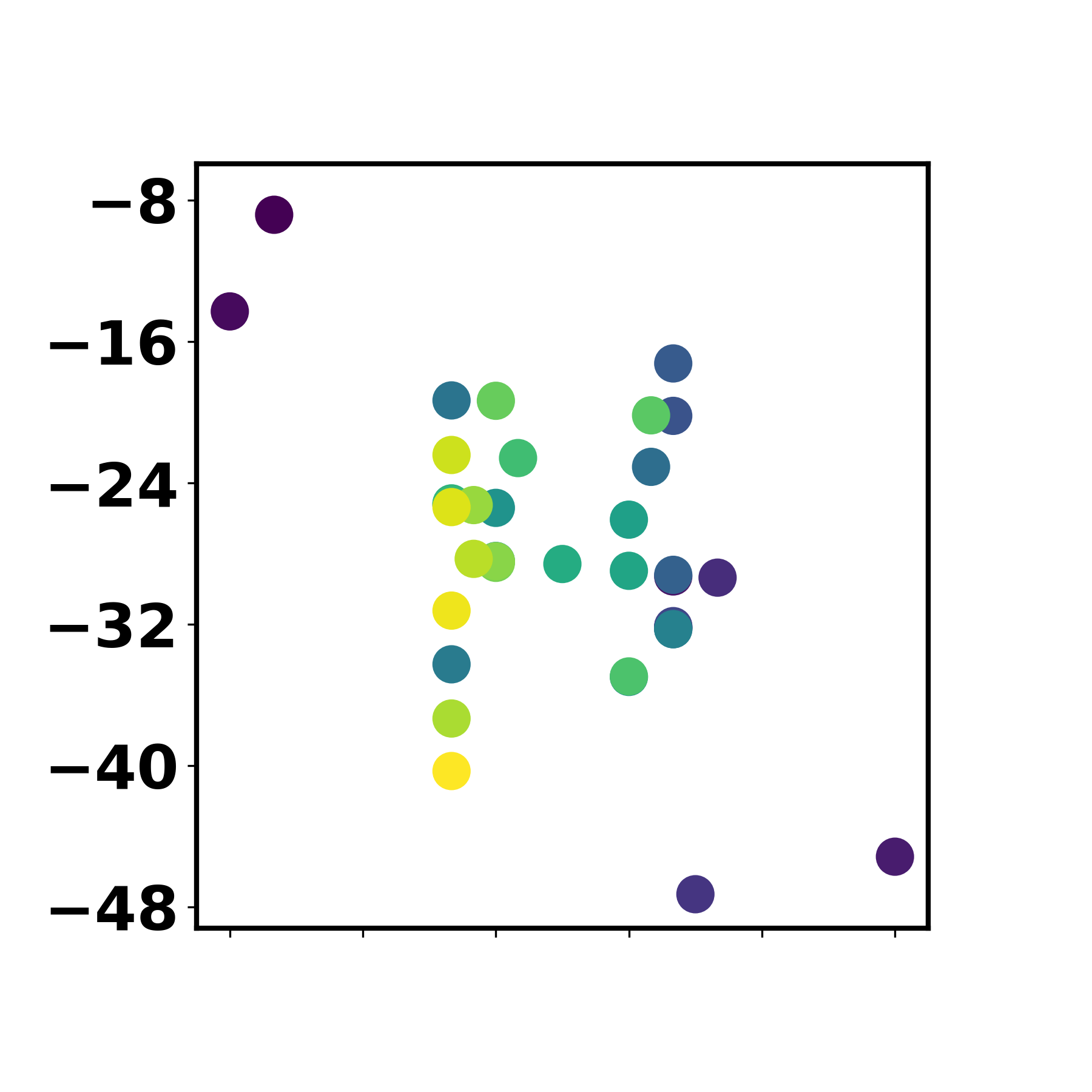}
  \end{subfigure}
  \begin{subfigure}[t]{.18\textwidth}
    \includegraphics[trim={0cm 0cm 0 2cm}, clip,width=1.1\linewidth]{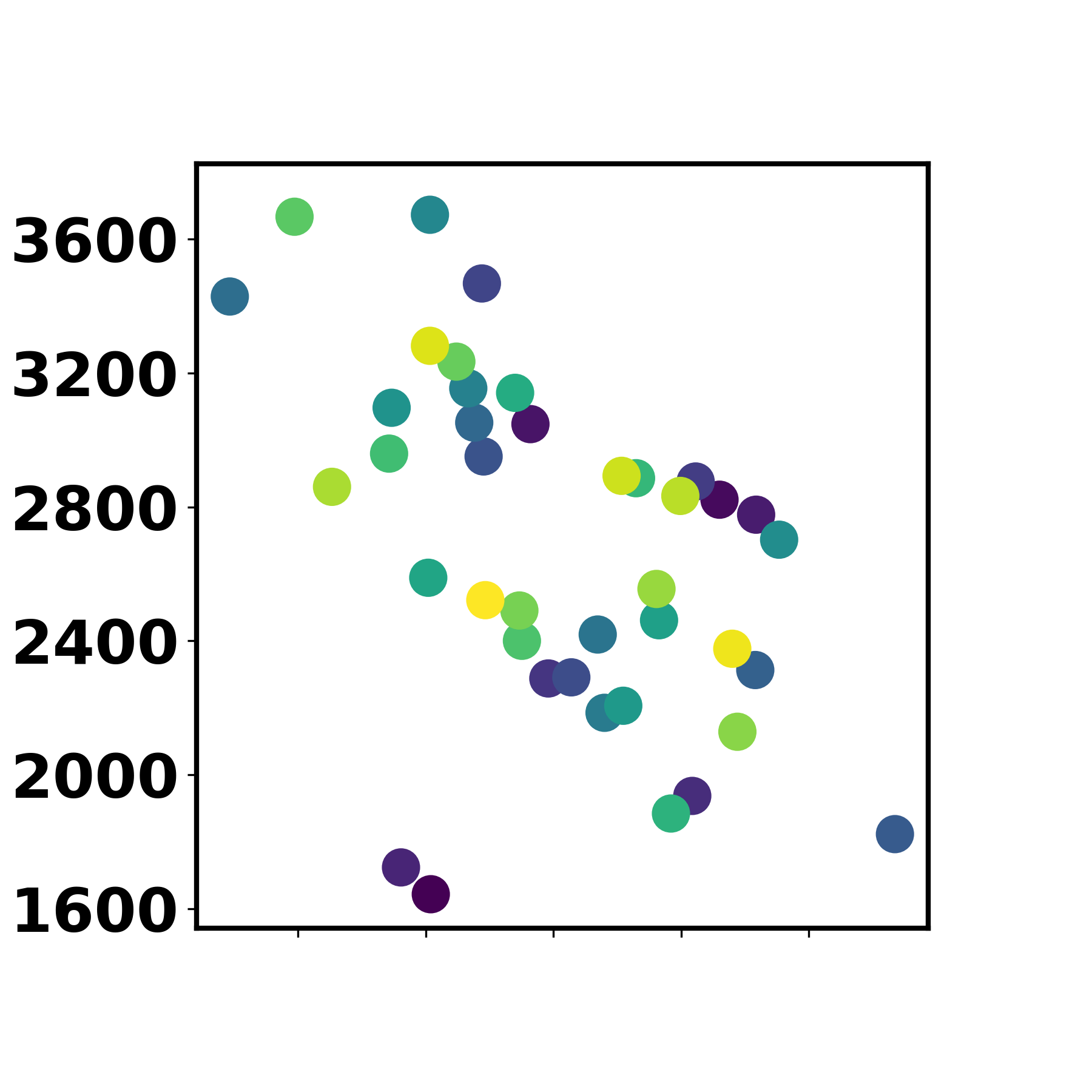}
    \end{subfigure}
  \begin{subfigure}[t]{.18\textwidth}
    \includegraphics[trim={0cm 0cm 0 2cm}, clip,width=1.1\linewidth]{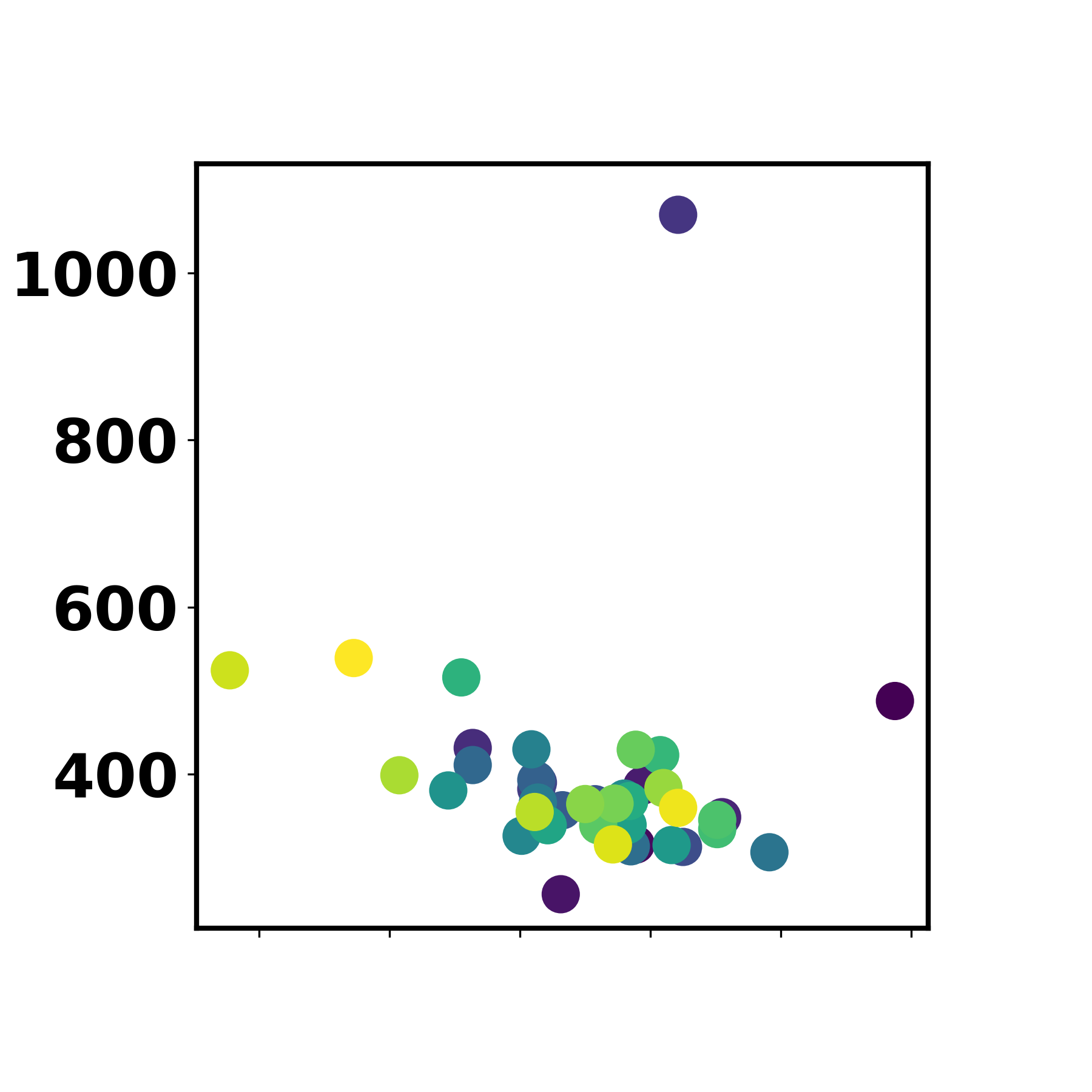}
  \end{subfigure}
    \begin{subfigure}[t]{.18\textwidth}
    \includegraphics[trim={2cm 0cm 0 2cm}, clip,width=1.5\linewidth]{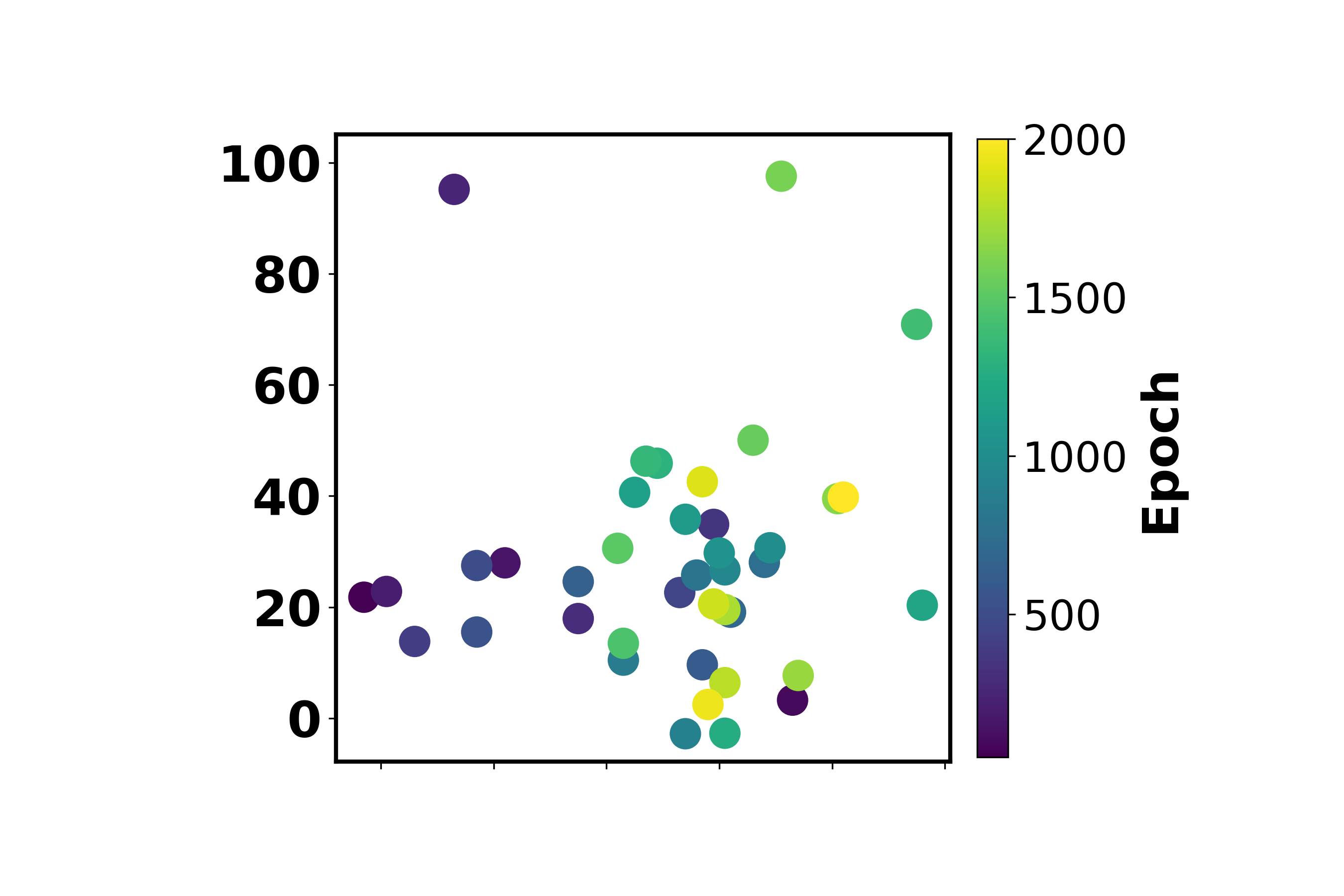}
  \end{subfigure}
  
\hspace{1.2em}
 \begin{subfigure}[t]{.18\textwidth}
  \begin{tikzpicture}
    \node[inner sep=0pt, outer sep=0pt] (img)
      {\includegraphics[trim={0cm 0cm 0 2cm}, clip, width=1.1\linewidth]{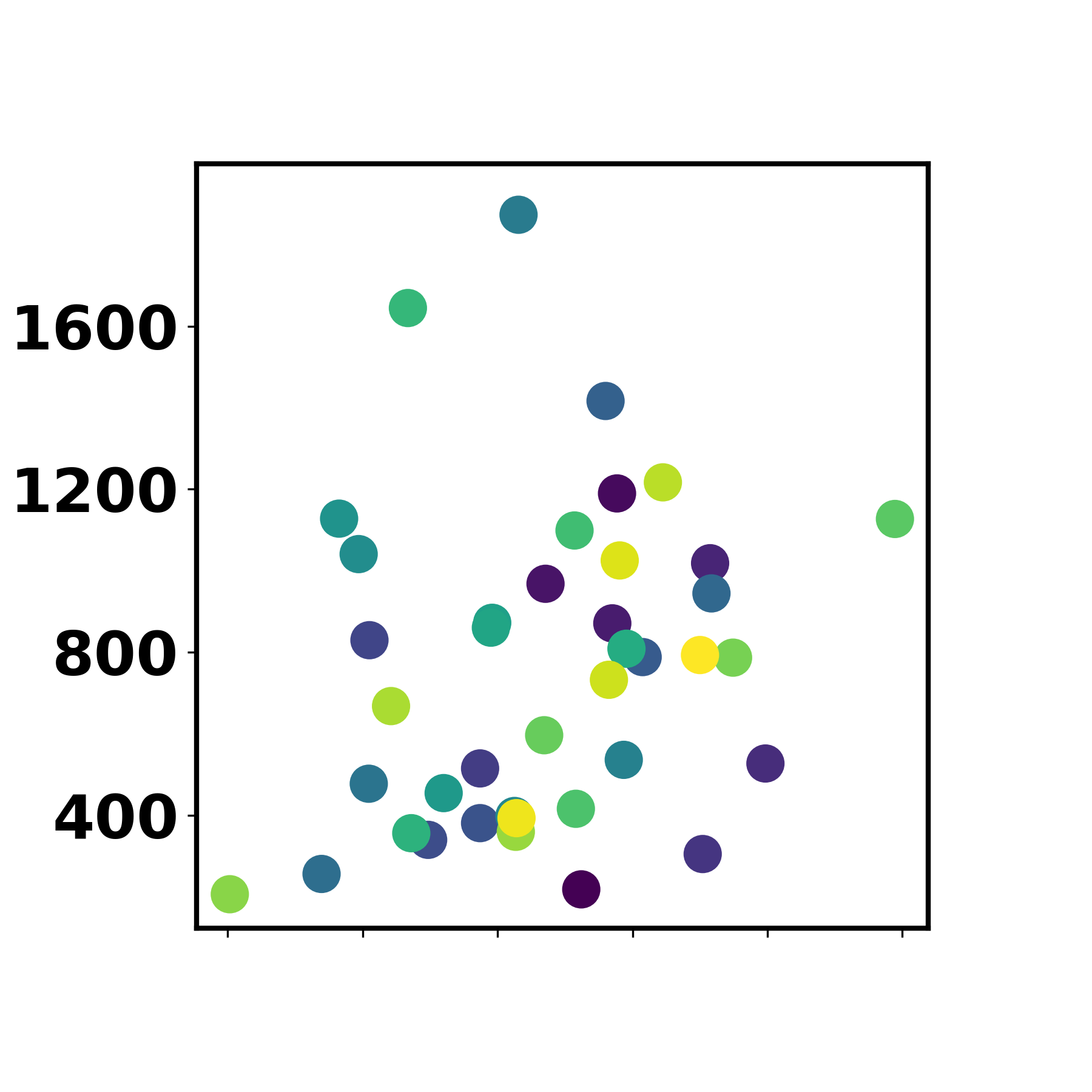}};
    \node[font=\scriptsize, anchor=north, inner sep=0pt, outer sep=0pt,
          text height=1ex, text depth=0pt]
      at ($(img.north)+(0,1.0ex)$) {Walker2d-m-r};
  \end{tikzpicture}
\end{subfigure}
  \begin{subfigure}[t]{.18\textwidth}
    \includegraphics[trim={0cm 0cm 0 2cm}, clip,width=1.1\linewidth]{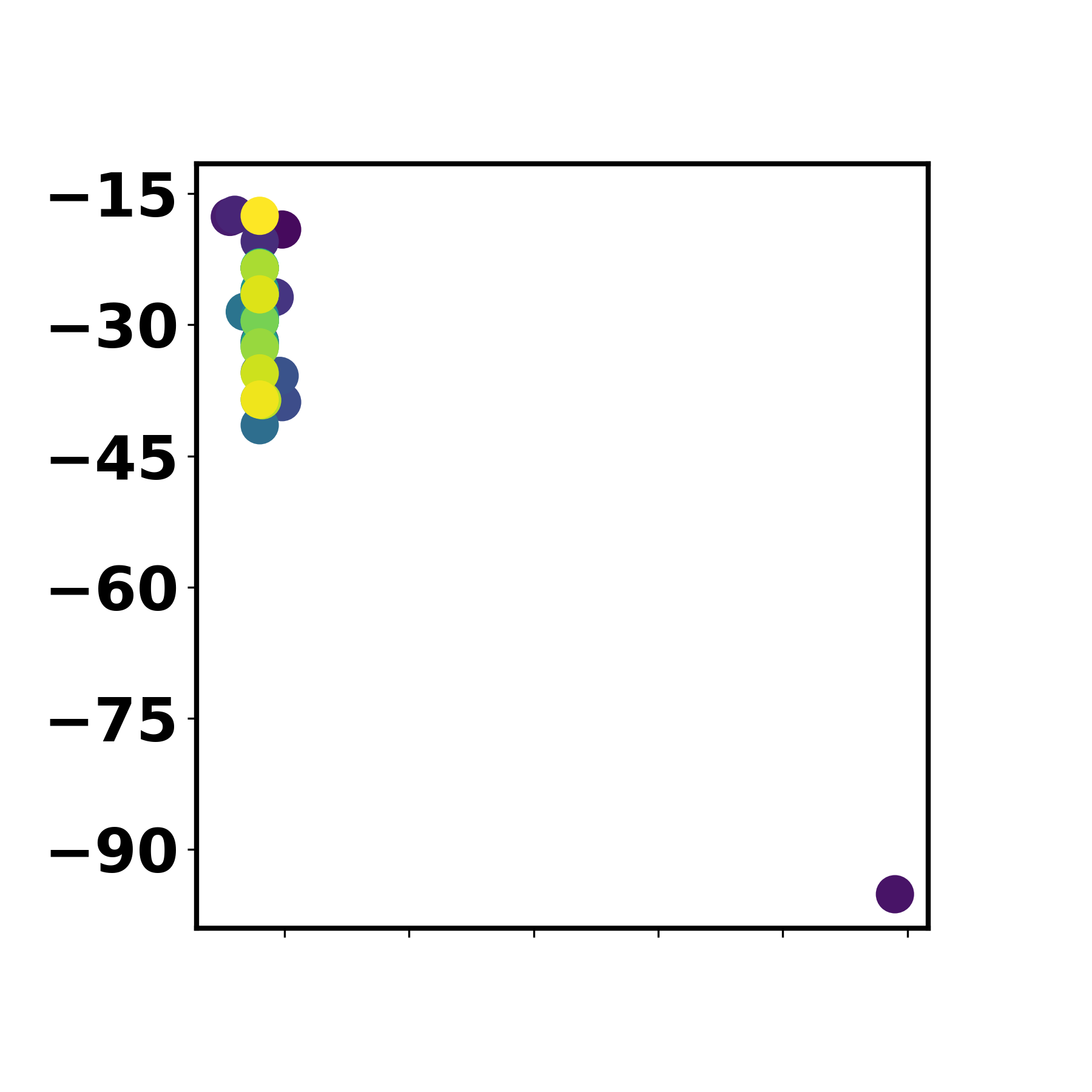}
  \end{subfigure}
  \begin{subfigure}[t]{.18\textwidth}
    \includegraphics[trim={0cm 0cm 0 2cm}, clip,width=1.1\linewidth]{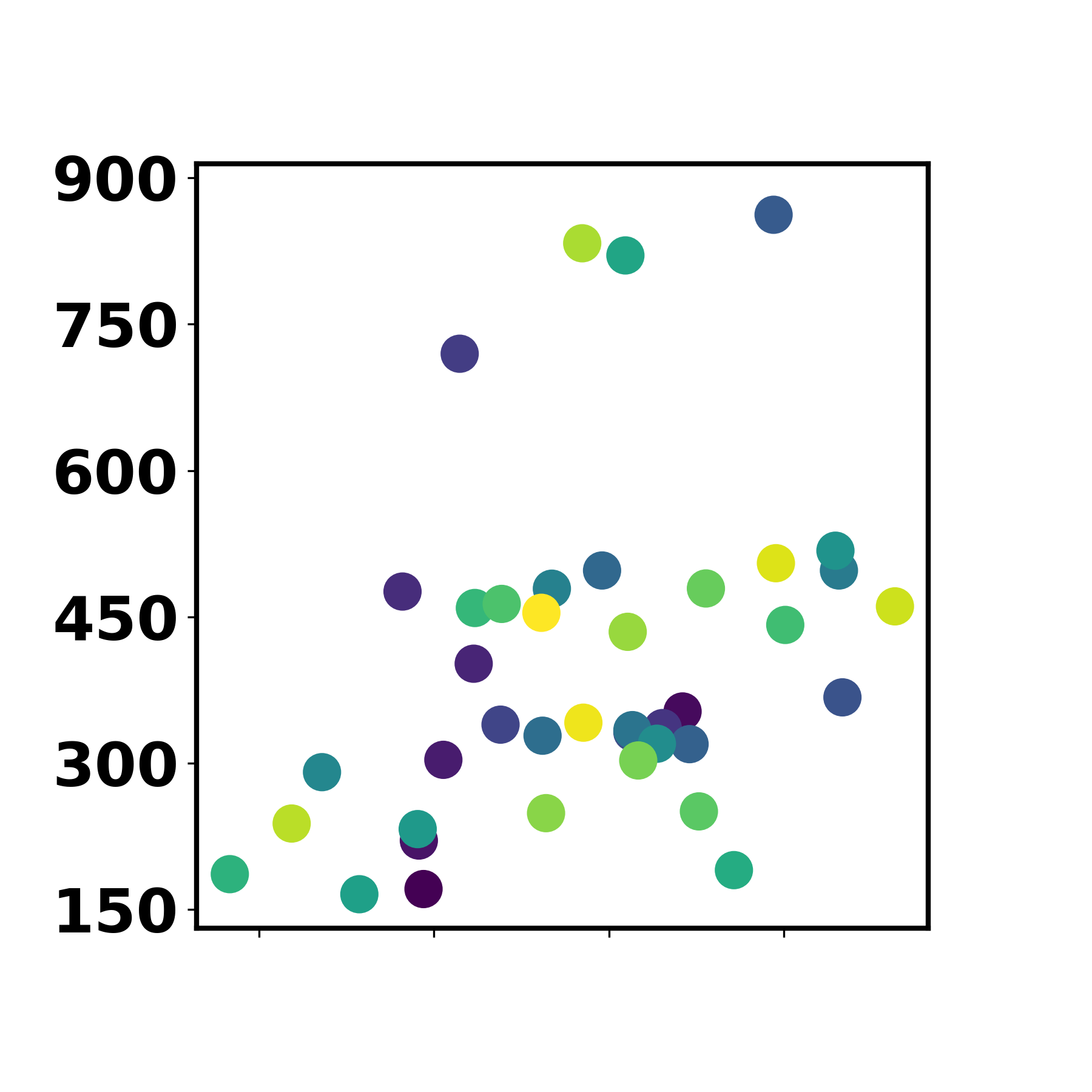}
    \end{subfigure}
  \begin{subfigure}[t]{.18\textwidth}
    \includegraphics[trim={0cm 0cm 0 2cm}, clip,width=1.1\linewidth]{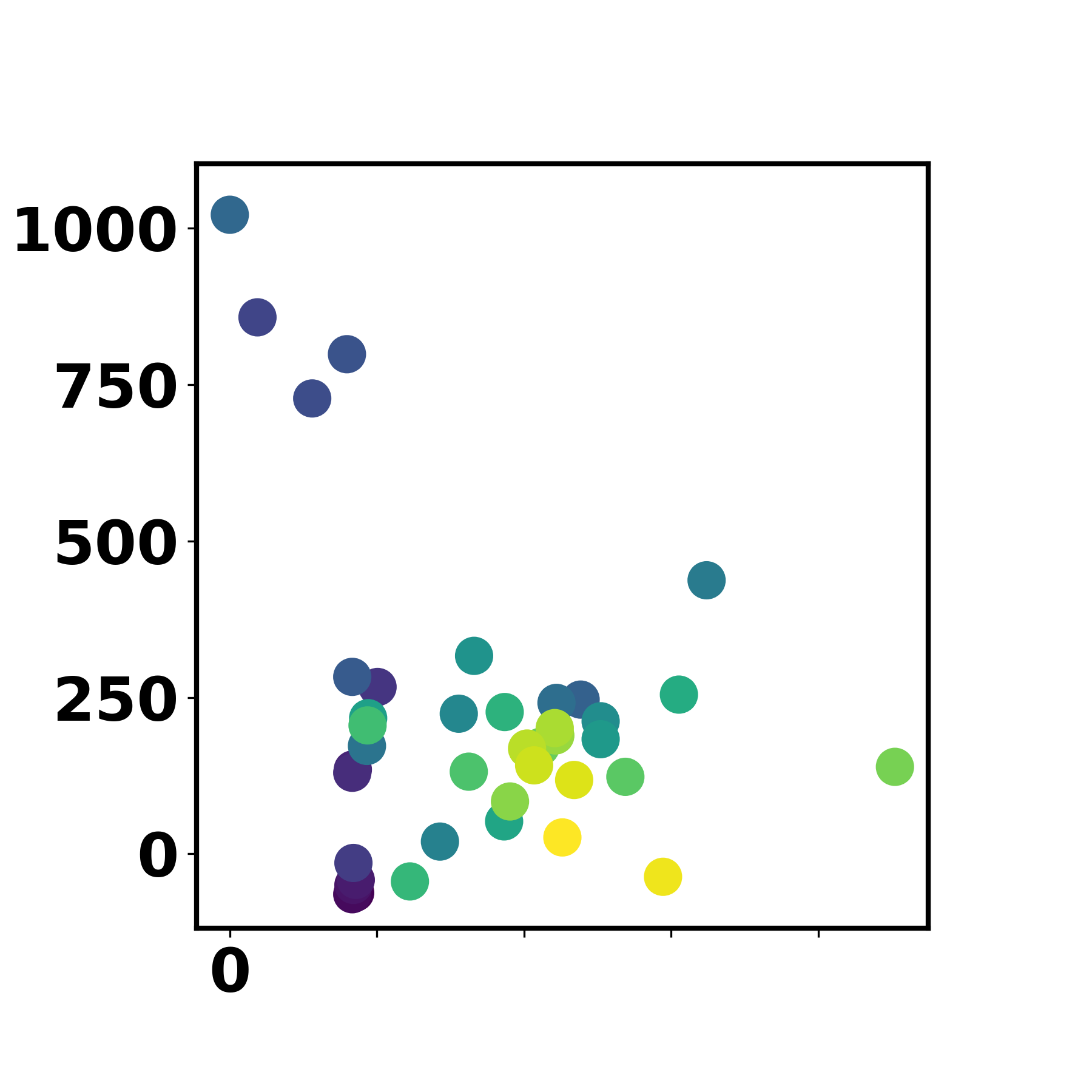}
  \end{subfigure}
  \begin{subfigure}[t]{.18\textwidth}
    \includegraphics[trim={2cm 0cm 0 2cm}, clip,width=1.5\linewidth]{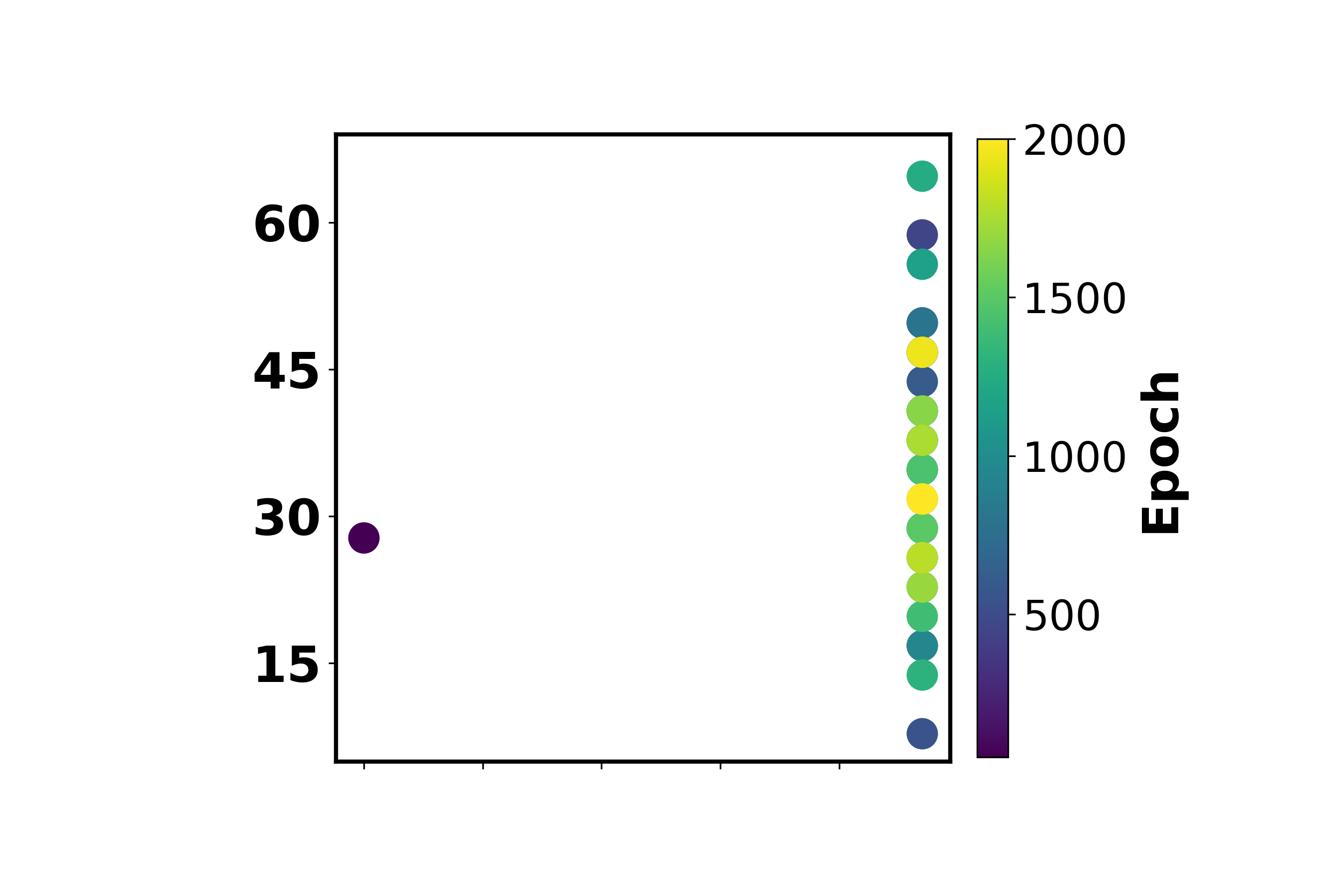}
  \end{subfigure}
  
\hspace{1.2em}
 \begin{subfigure}[t]{.18\textwidth}
  \begin{tikzpicture}
    \node[inner sep=0pt, outer sep=0pt] (img)
      {\includegraphics[trim={0cm 0cm 0 2cm}, clip, width=1.1\linewidth]{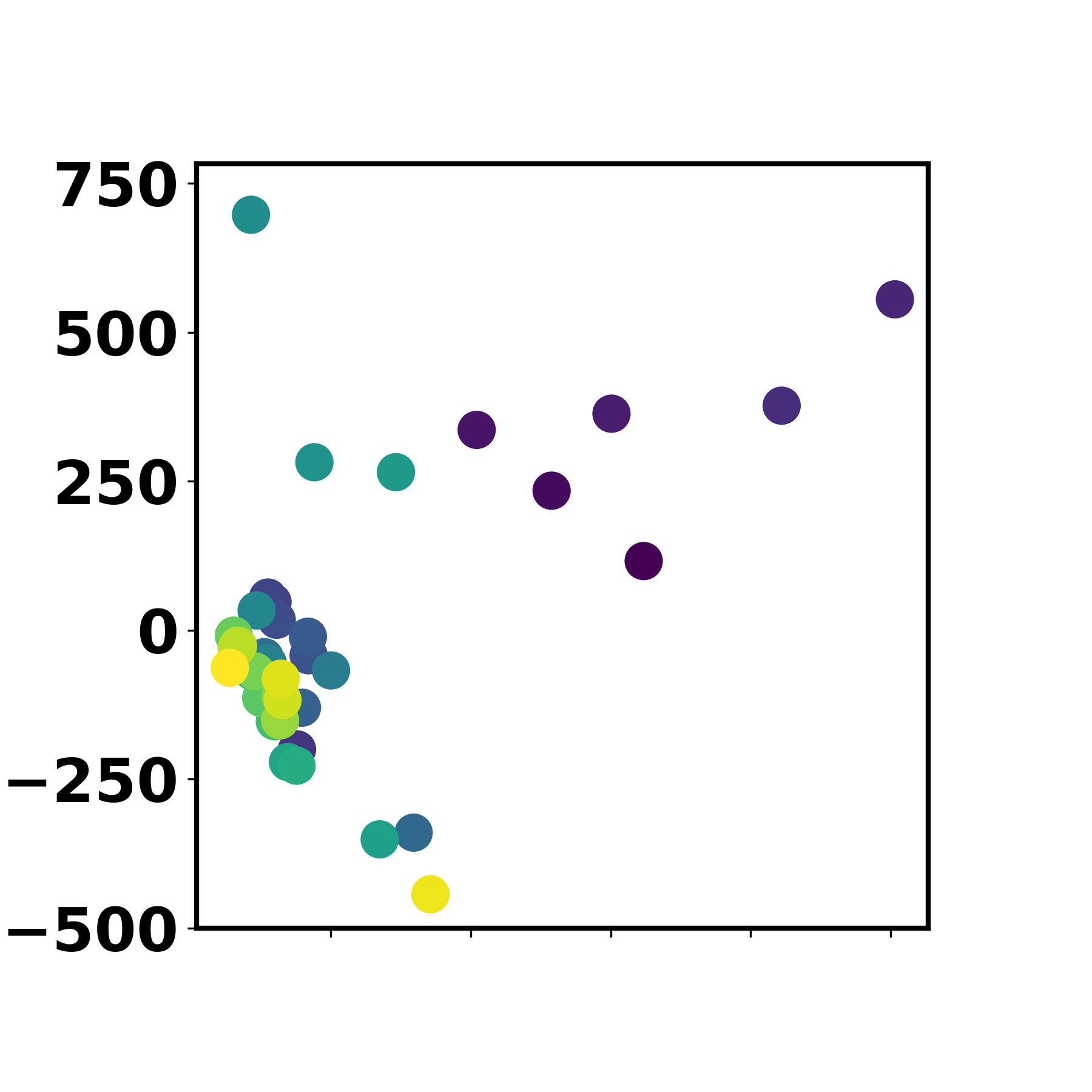}};
    \node[font=\scriptsize, anchor=north, inner sep=0pt, outer sep=0pt,
          text height=1ex, text depth=0pt]
      at ($(img.north)+(0,1.0ex)$) {Hopper-m-e};
  \end{tikzpicture}
\end{subfigure}
  \begin{subfigure}[t]{.18\textwidth}
    \includegraphics[trim={0cm 0cm 0 2cm}, clip,width=1.1\linewidth]{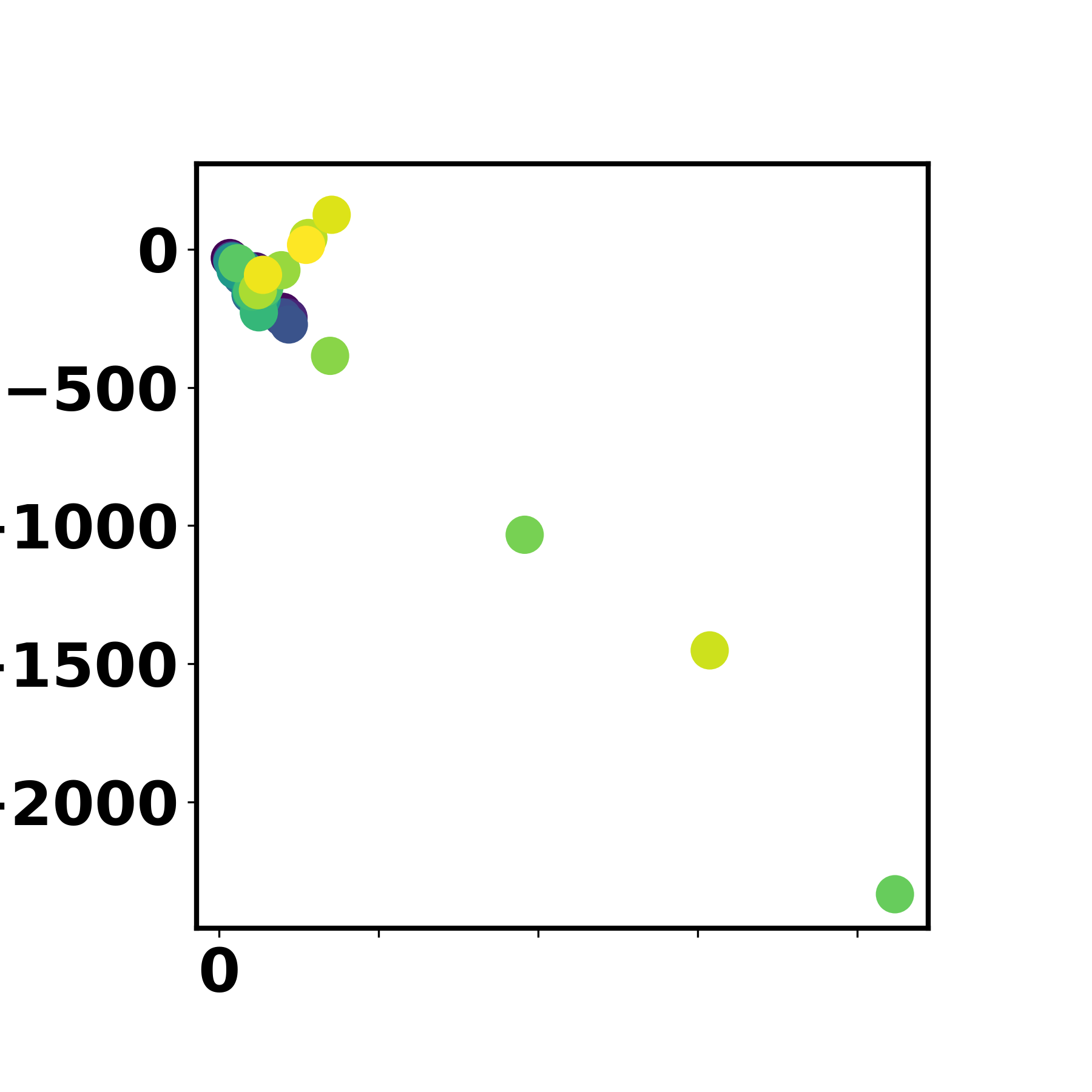}
  \end{subfigure}
  \begin{subfigure}[t]{.18\textwidth}
    \includegraphics[trim={0cm 0cm 0 2cm}, clip,width=1.1\linewidth]{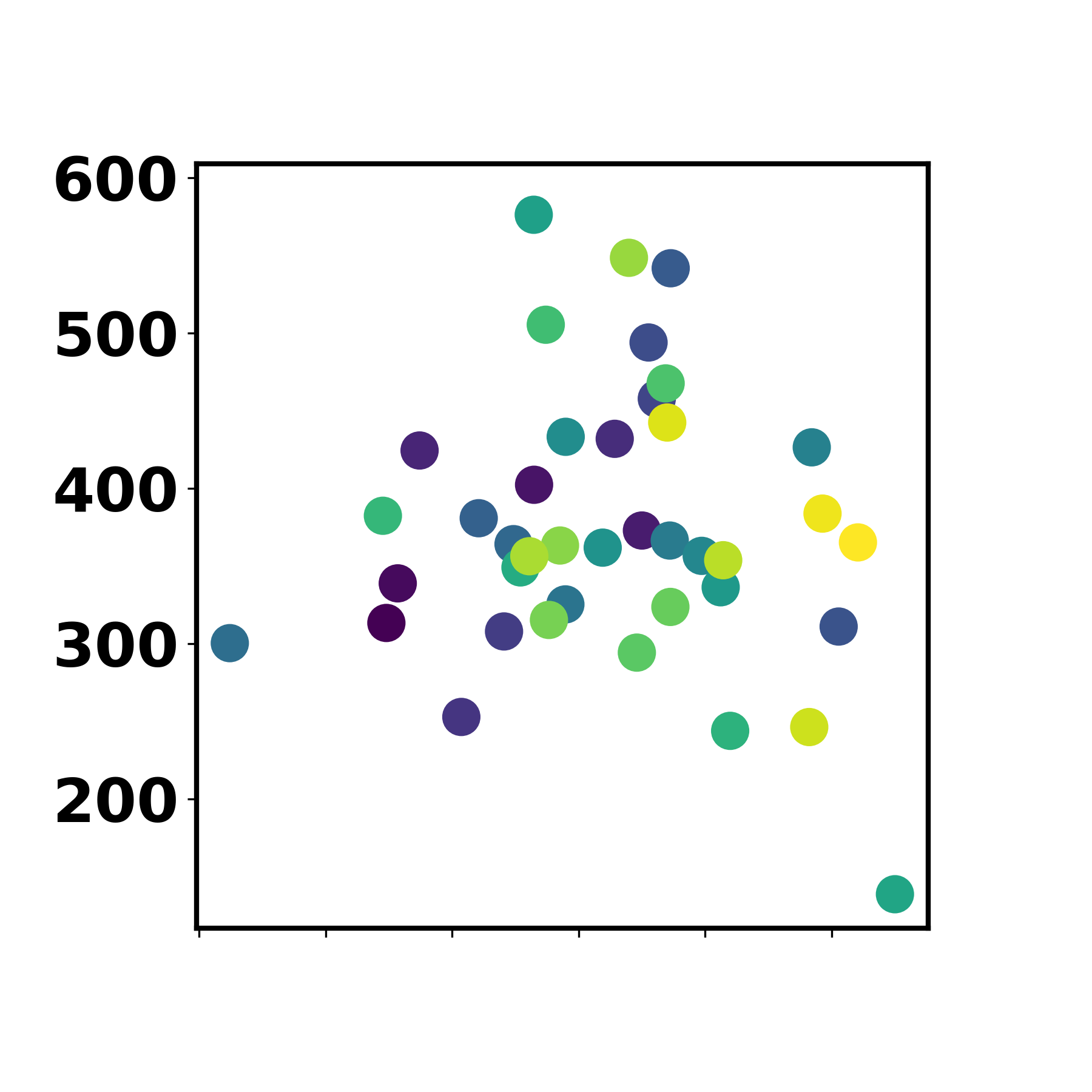}
    \end{subfigure}
  \begin{subfigure}[t]{.18\textwidth}
    \includegraphics[trim={0cm 0cm 0 2cm}, clip,width=1.1\linewidth]{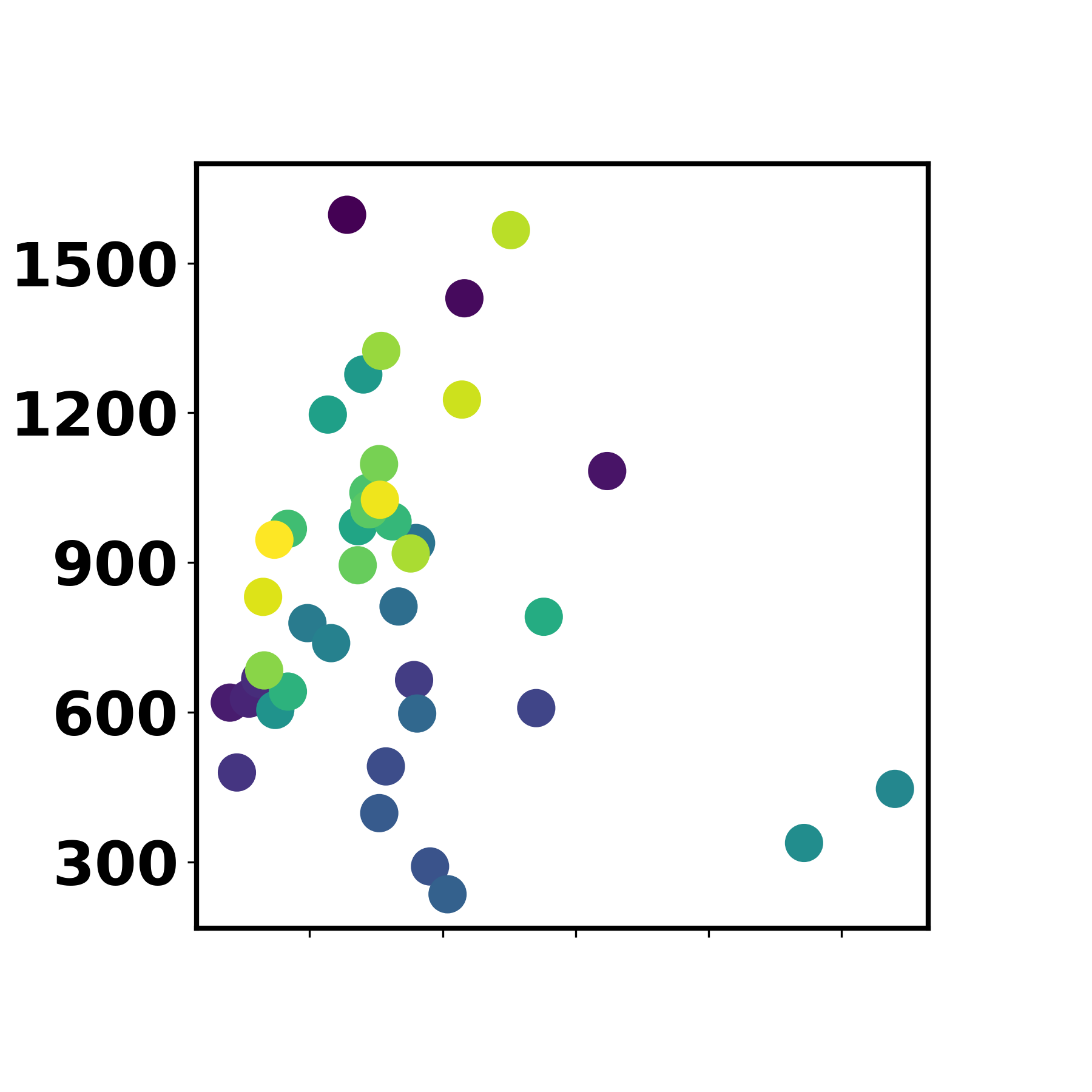}
  \end{subfigure}
  \begin{subfigure}[t]{.18\textwidth}
    \includegraphics[trim={2cm 0cm 0 2cm}, clip,width=1.5\linewidth]{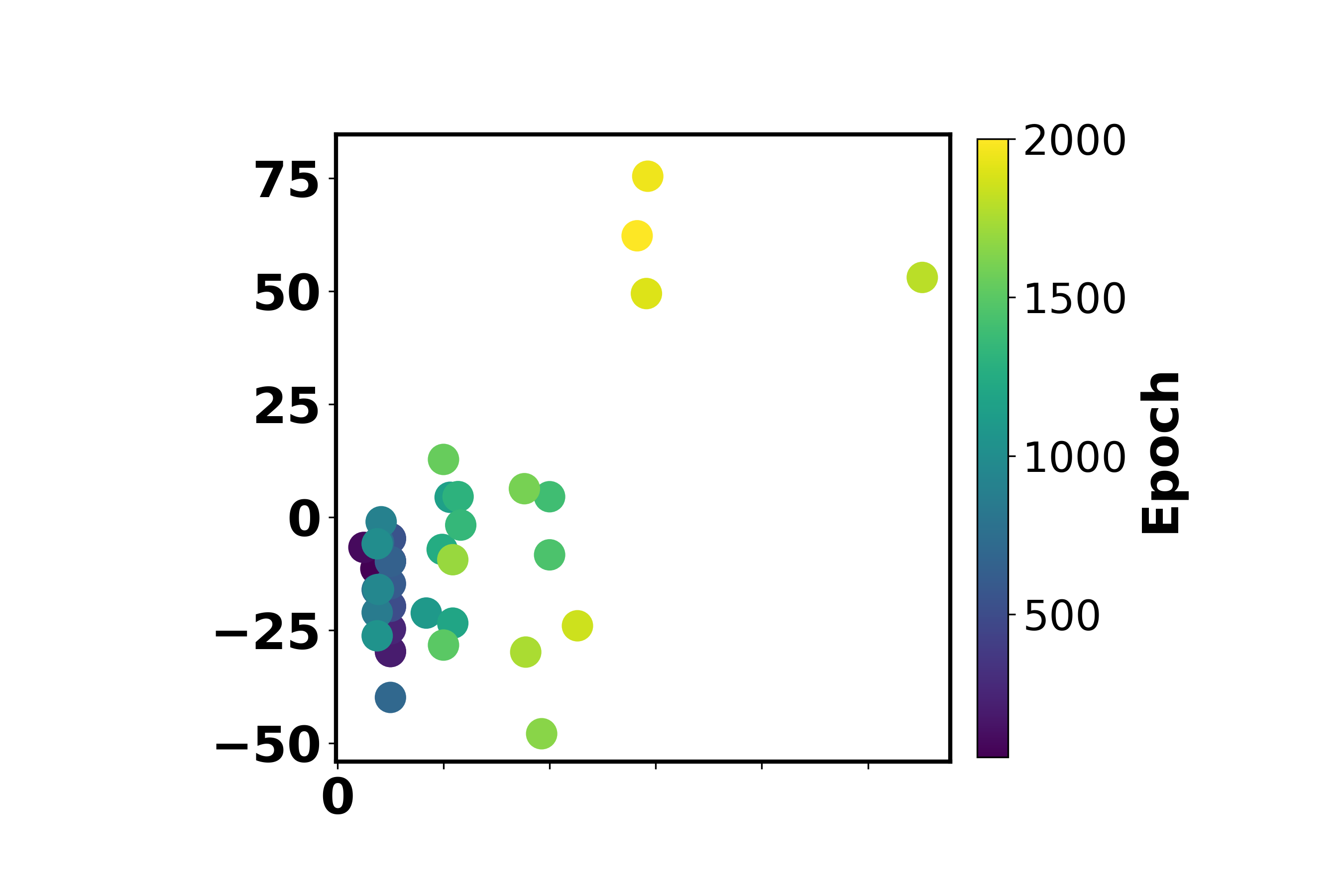}
  \end{subfigure}

\hspace{1.2em}
\begin{subfigure}[t]{.18\textwidth}
  \begin{tikzpicture}
    \node[inner sep=0pt, outer sep=0pt] (img)
      {\includegraphics[trim={0cm 1.5cm 0 2cm}, clip, width=1.1\linewidth]{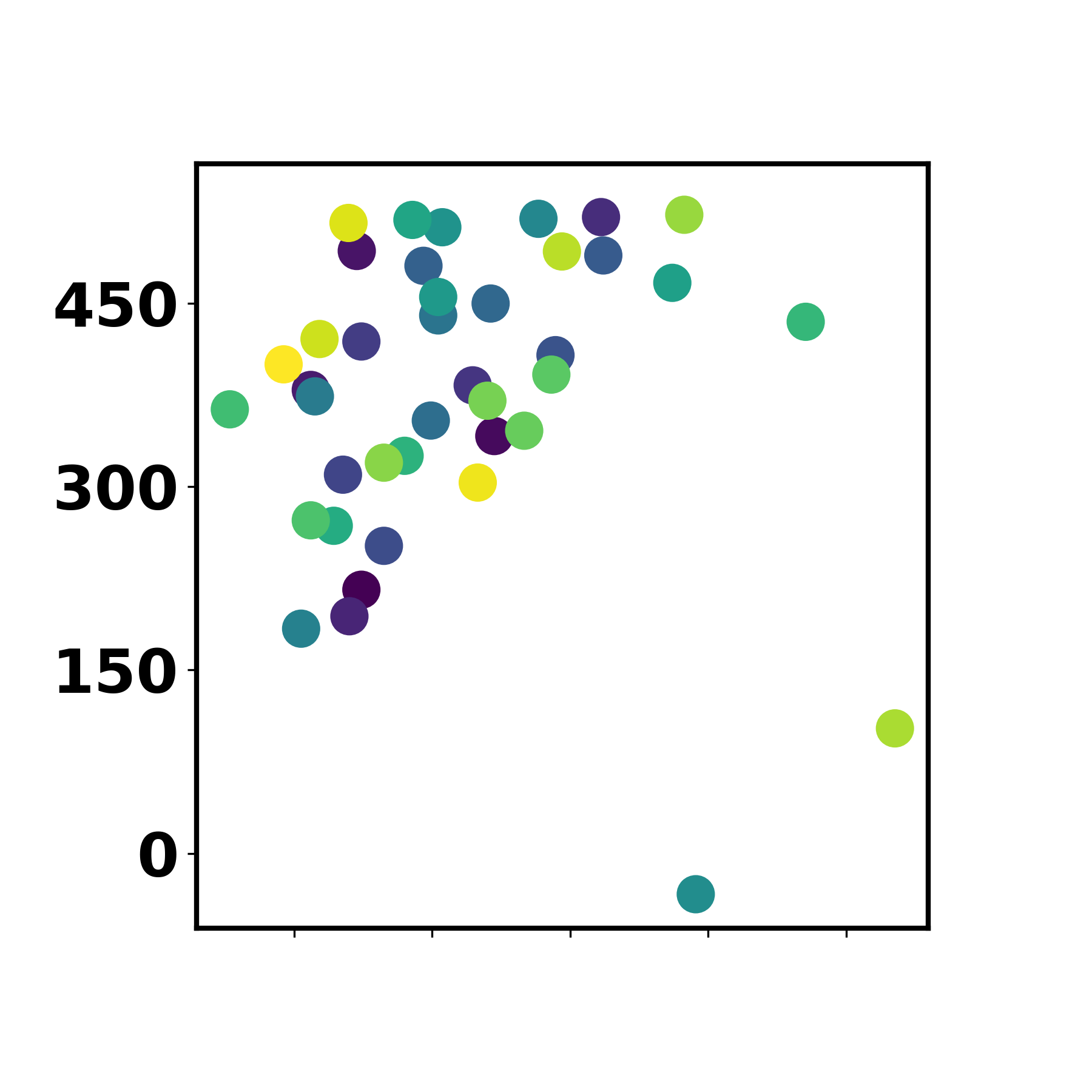}};
    \node[font=\scriptsize, anchor=north, inner sep=0pt, outer sep=0pt,
          text height=1ex, text depth=0pt]
      at ($(img.north)+(0,1.0ex)$) {Hopper-m-r};
  \end{tikzpicture}
  \caption*{\hspace{1em}\footnotesize RADAC} 
\end{subfigure}
  \begin{subfigure}[t]{.18\textwidth}
    \includegraphics[trim={0cm 1.5cm 0 2cm}, clip,width=1.1\linewidth]{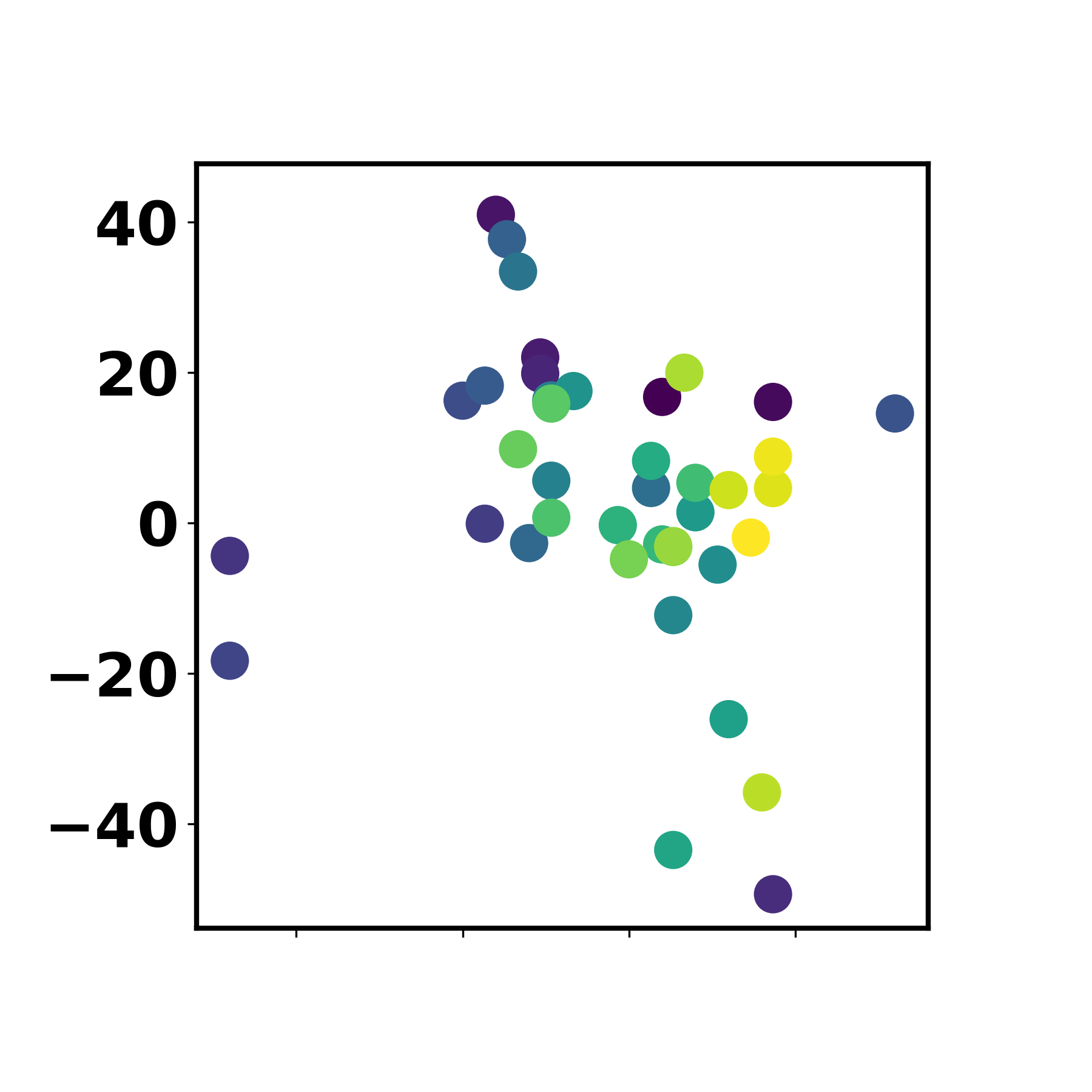}
    \caption*{\hspace{1em}\footnotesize DiffusionQL}
  \end{subfigure}
  \begin{subfigure}[t]{.18\textwidth}
    \includegraphics[trim={0cm 1.5cm 0 2cm}, clip,width=1.1\linewidth]{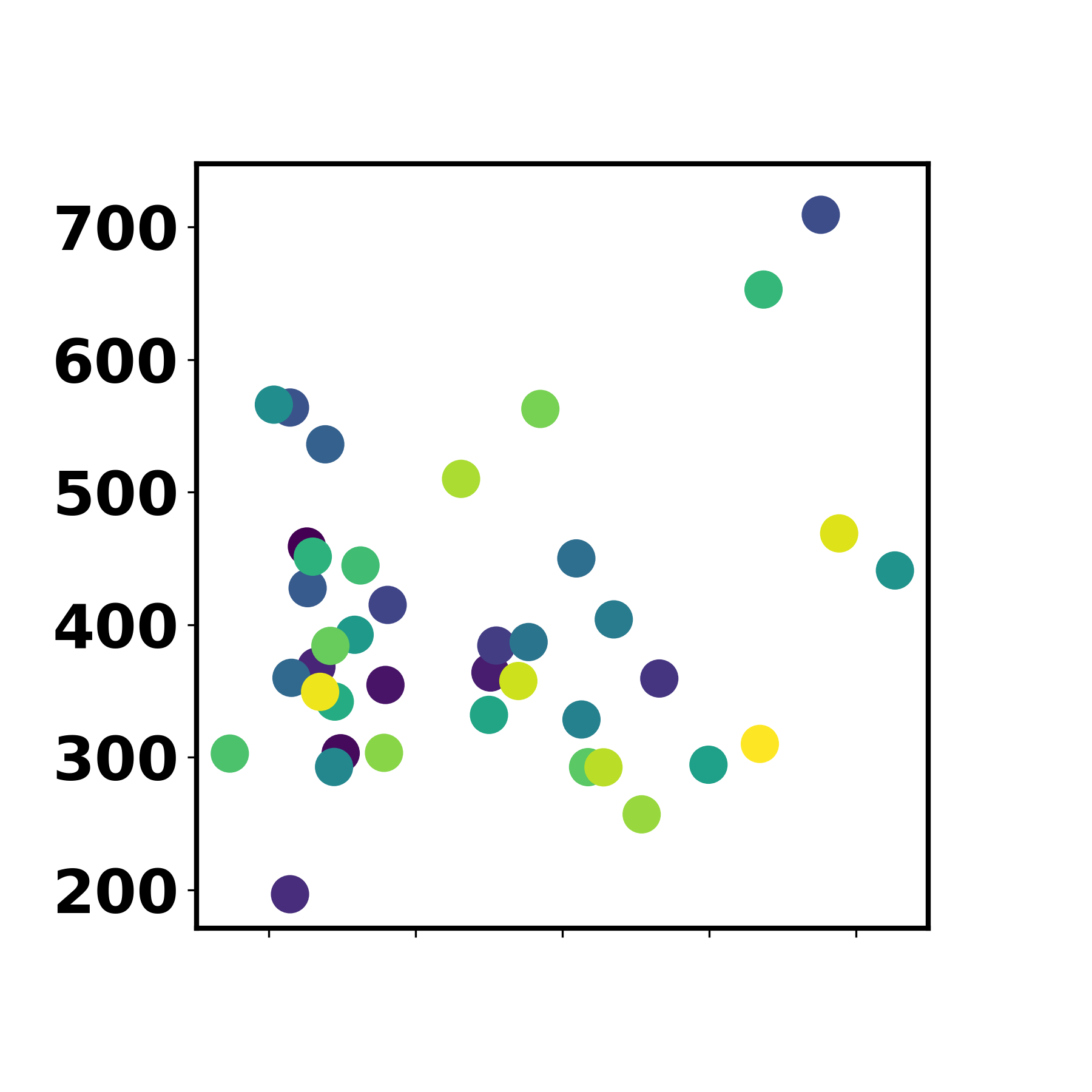}
    \caption*{\hspace{1em}\footnotesize FQL}
    \end{subfigure}
  \begin{subfigure}[t]{.18\textwidth}
    \includegraphics[trim={0cm 1.5cm 0 2cm}, clip,width=1.1\linewidth]{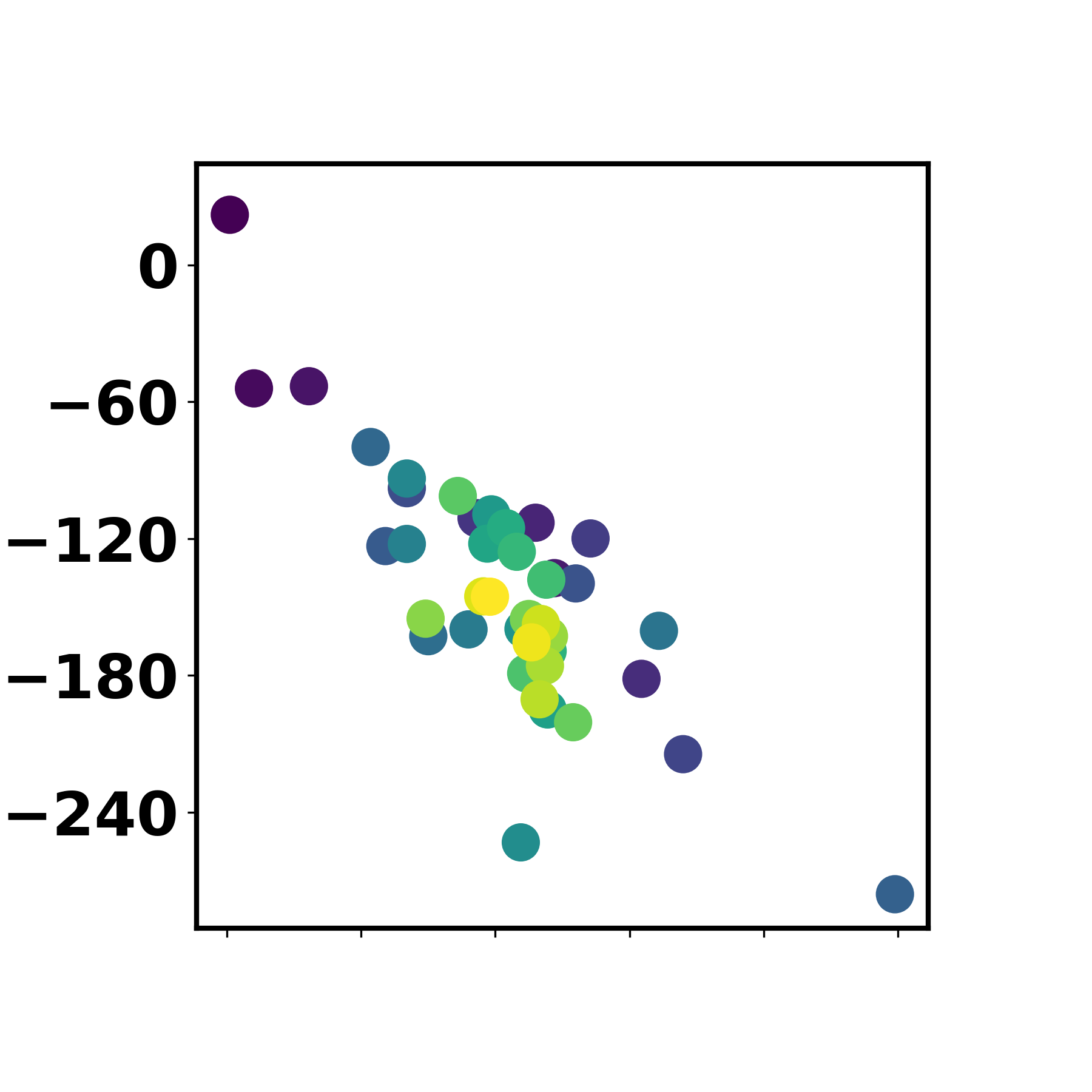}
    \caption*{\hspace{1em}\footnotesize ORAAC}
  \end{subfigure}
  \begin{subfigure}[t]{.18\textwidth}
    \includegraphics[trim={2cm 1.5cm 0 2cm}, clip,width=1.5\linewidth]{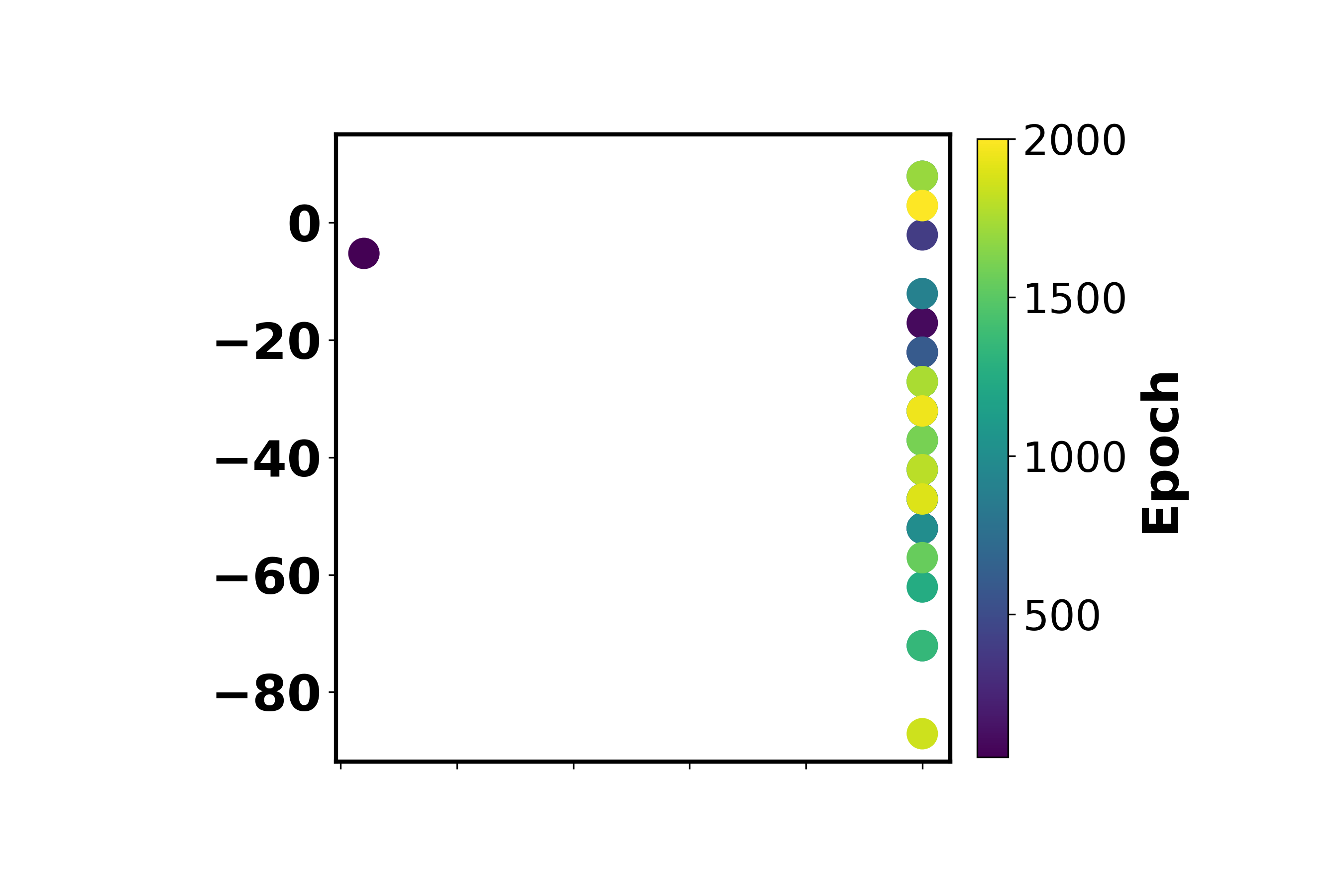}
    \caption*{\hspace{2em}\footnotesize CODAC}
  \end{subfigure}
  \caption{Pareto frontiers of return vs.\ safety violations.
Rows are Stochastic–D4RL tasks (top$\to$bottom: \textsc{HalfCheetah}-m-e, \textsc{HalfCheetah}-m-r, \textsc{Walker2d}-m-e, \textsc{Walker2d}-m-r, \textsc{Hopper}-m-e, \textsc{Hopper}-m-r);
columns are algorithms (left$\to$right: RADAC, DiffusionQL, FQL, ORAAC, CODAC).
Points are evaluation snapshots across training (color encodes epoch; dark$\to$yellow).
$x$–axis: violation count per episode; $y$–axis: mean return (upper–left is better)}
\label{fig:pareto_frontiers}
\end{figure}
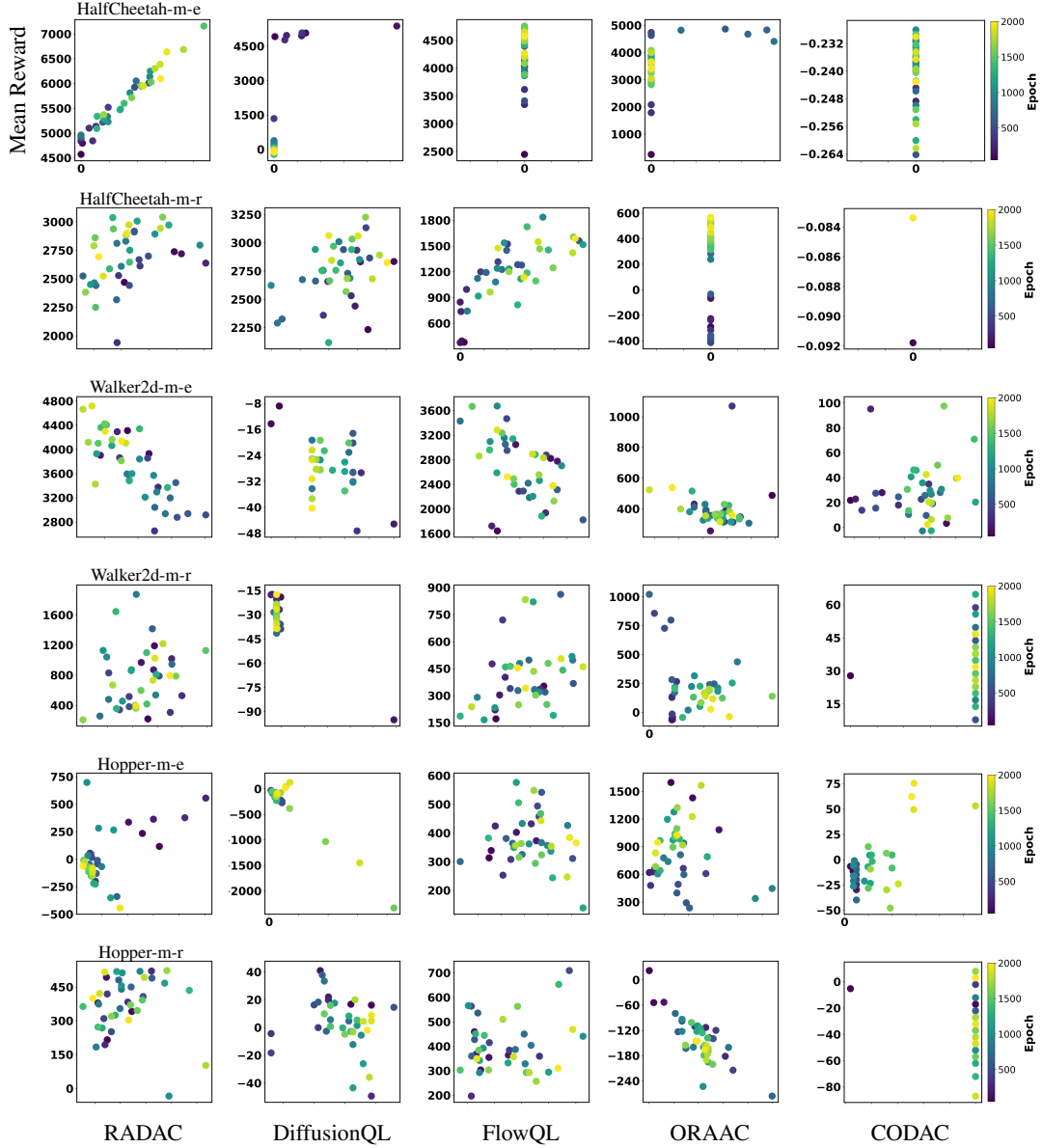
\subsection{Extended Stochastic-D4RL Results}
\label{app:sd4rl_full}
% \paragraph{Protocol}
% To remove post-hoc checkpoint selection and ease reproducibility, we report a full result in Sec.~\ref{sec:} with s.e. in Table~\ref{tab:sd4rl_results_1000_by_dataset} and \emph{fixed 1000-gradient-step evaluation} for every method and task in Table~\ref{tab:sd4rl_results_1000_by_dataset}. 
% Scores are raw returns and episodic $\mathrm{CVaR}_{0.1}$ (mean $\pm$ s.e.\ over 5 seeds), without normalization, matching the stochastic variants used in the main text.
\paragraph{Protocol}
Table~\ref{tab:sd4rl_results_1000_by_dataset} reports an additional
fixed-checkpoint evaluation under a uniform 1000-step rollout horizon for all included
methods and tasks. This evaluation is complementary to the main-text
comparison: it removes checkpoint-selection effects and therefore
should not be interpreted as reproducing the exact ranking in
Table~\ref{tab:d4rl_score}.
\paragraph{Consistency with the Main-Text Trends}
The fixed-checkpoint evaluation preserves several strengths of RADAC,
particularly on \textsc{HalfCheetah}-m-e and \textsc{Hopper}-m-e,
while also revealing settings in which FQL or RAFMAC achieves stronger
mean or lower-tail performance.

\paragraph{Pareto Frontier Analysis: Return vs.\ Safety Violations}
\label{app:pareto_frontiers}
% Figure~\ref{fig:pareto_frontiers} plots mean return (y) against safety-violation counts per episode (x), with color indicating training progress. Unless noted, comparisons refer to the same 1000-step evaluation as in Table~\ref{tab:sd4rl_results_1000_by_dataset}. 
Figure~\ref{fig:pareto_frontiers} plots mean return against the number
of safety violations per episode, with each point corresponding to an
evaluation snapshot collected during training and color indicating the
training epoch. This trajectory-level view complements the fixed
1000-step comparison in Table~\ref{tab:sd4rl_results_1000_by_dataset}. We organize the discussion by algorithm.

\begin{figure*}[t]
  \centering
  %================= Medium--Expert (3 panels) =================
  \begin{subfigure}[t]{.30\textwidth}
    \caption{\small HalfCheetah-Medium-Expert}
    \makebox[0pt][r]{%
      \raisebox{.25\linewidth}{\rotatebox{90}{\normalsize\text{Density}}}%
    }%
    \begin{tikzpicture}
      \node[inner sep=0pt, name=img]
        {\includegraphics[width=\linewidth]{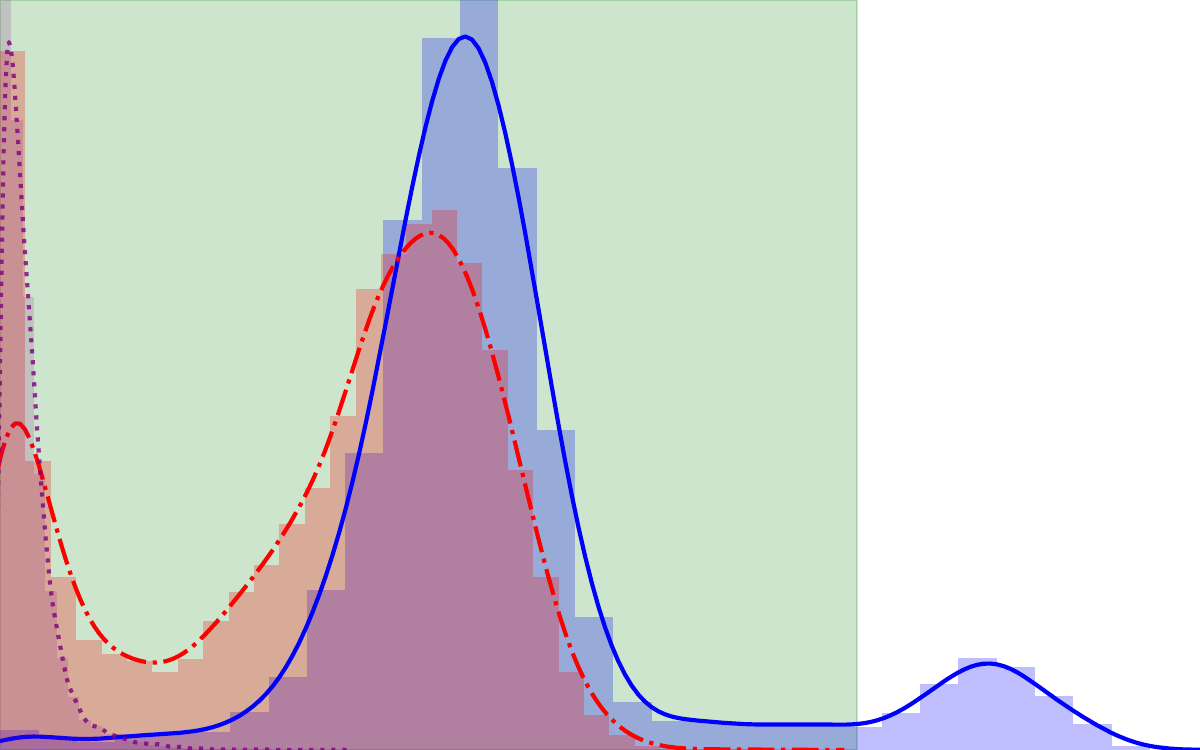}};
      \def\inset{0.6pt}
      \draw[line width=0.35pt]
        ($(img.south west)+(\inset,\inset)$) rectangle
        ($(img.north east)+(-\inset,-\inset)$);
      \coordinate (FSW) at ($(img.south west)+(\inset,\inset)$);
      \coordinate (FSE) at ($(img.south east)+(-\inset,\inset)$);
      \def\ticklen{2.0pt}\def\labelsep{0.7pt}
      \foreach \x/\lab in {0.007/0, 0.350/5, 0.707/10}{
        \coordinate (P) at ($(FSW)!\x!(FSE)$);
        \draw (P) -- ++(0,-\ticklen);
        \node[anchor=north, inner sep=0pt, font=\footnotesize\bfseries]
             at ($(P)+(0,-\ticklen-\labelsep)$) {\lab};
      }
    \end{tikzpicture}
  \end{subfigure}\hspace{0.65em}
  \begin{subfigure}[t]{.30\textwidth}
    \caption{\small Walker2d-Medium-Expert}
    \begin{tikzpicture}
      \node[inner sep=0pt, name=img]
        {\includegraphics[width=\linewidth]{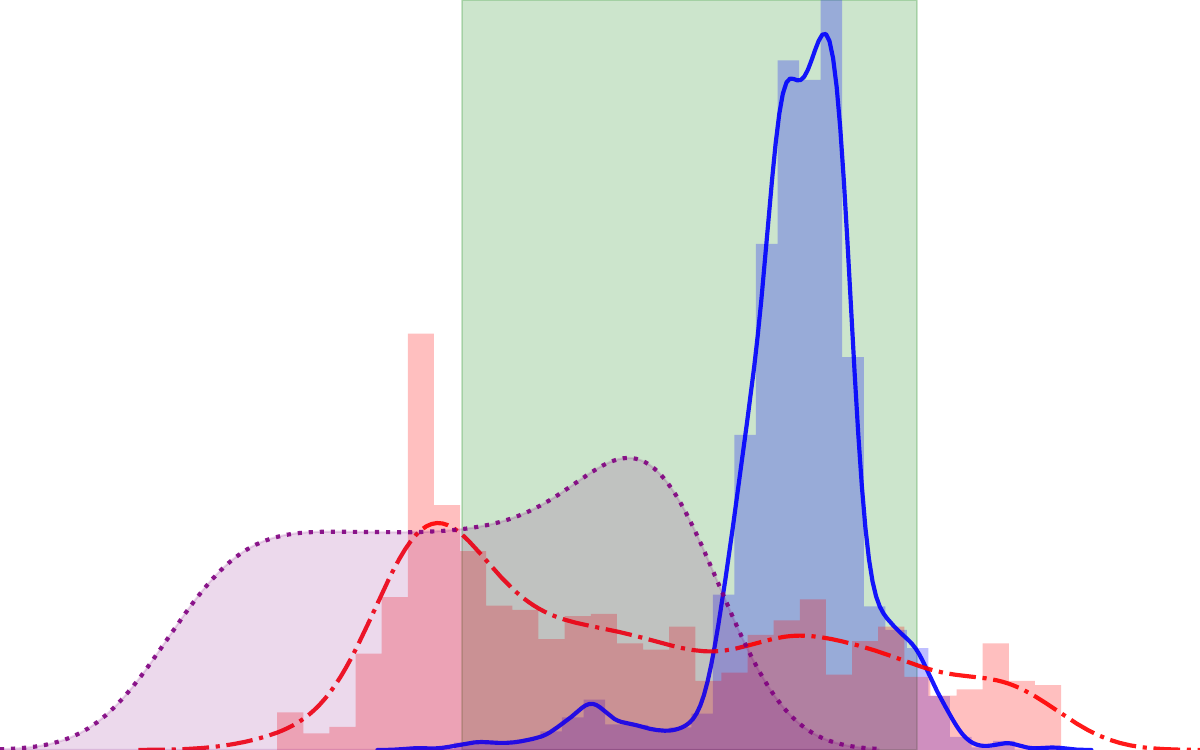}};
      \def\inset{0.6pt}
      \draw[line width=0.35pt]
        ($(img.south west)+(\inset,\inset)$) rectangle
        ($(img.north east)+(-\inset,-\inset)$);
      \coordinate (FSW) at ($(img.south west)+(\inset,\inset)$);
      \coordinate (FSE) at ($(img.south east)+(-\inset,\inset)$);
      \def\ticklen{2.0pt}\def\labelsep{0.7pt}
      \foreach \x/\lab in {0.39/-0.5, 0.575/0, 0.76/0.5}{
        \coordinate (P) at ($(FSW)!\x!(FSE)$);
        \draw (P) -- ++(0,-\ticklen);
        \node[anchor=north, inner sep=0pt, font=\footnotesize\bfseries]
             at ($(P)+(0,-\ticklen-\labelsep)$) {\lab};
      }
    \end{tikzpicture}
  \end{subfigure}\hspace{0.55em}
  \begin{subfigure}[t]{.30\textwidth}
    \caption{\small Hopper-Medium-Expert}
    \begin{tikzpicture}
      \node[inner sep=0pt, name=img]
        {\includegraphics[width=\linewidth]{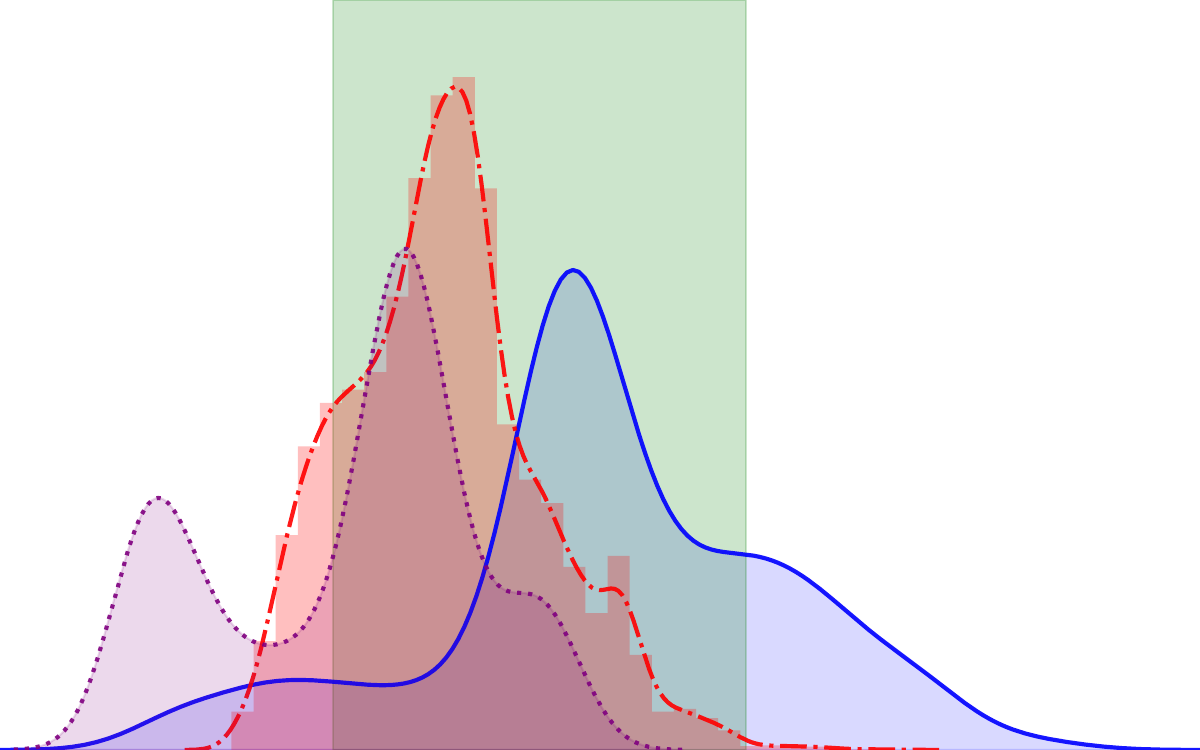}};
      \def\inset{0.6pt}
      \draw[line width=0.35pt]
        ($(img.south west)+(\inset,\inset)$) rectangle
        ($(img.north east)+(-\inset,-\inset)$);
      \coordinate (FSW) at ($(img.south west)+(\inset,\inset)$);
      \coordinate (FSE) at ($(img.south east)+(-\inset,\inset)$);
      \def\ticklen{2.0pt}\def\labelsep{0.7pt}
      \foreach \x/\lab in {0.28/-0.1, 0.45/0, 0.62/0.1}{
        \coordinate (P) at ($(FSW)!\x!(FSE)$);
        \draw (P) -- ++(0,-\ticklen);
        \node[anchor=north, inner sep=0pt, font=\footnotesize\bfseries]
             at ($(P)+(0,-\ticklen-\labelsep)$) {\lab};
      }
    \end{tikzpicture}
  \end{subfigure}

  \caption{Additional qualitative safety heatmaps on \textsc{Medium--Expert}.
  Shaded bands indicate risk-free operating ranges (HalfCheetah: $v\le 10$; Walker2d: $|\theta|\le 0.5$; Hopper: $|\theta|\le 0.1$).
  The qualitative trends match Fig.~\ref{fig:safety_heatmaps} (main text): RADAC shifts mass away from hazardous regions while remaining concentrated on dataset-supported modes.}
  \label{fig:safety_heatmaps_me}
\end{figure*}

Across datasets, RADAC populates the upper-left region of the frontier: for comparable return, it tends to incur fewer violations. Only for \textsc{HalfCheetah}–medium–expert, RADAC sometimes drifts up-right (higher return with slightly more violations) because the penalty is light and non-terminating, so near-threshold speed pays off, consistent with its best Mean/CVaR. Mechanistically, diffusion with CVaR guidance enables fine-grained reweighting away from safety thresholds while BC keeps samples on-manifold, so trajectories in the plot drift left (fewer violations) without sacrificing return.
ORAAC forms the frontier on \textsc{Hopper}-m-e with few violations and strong returns, matching its leading scores under terminating pose hazards. In other settings it remains reliably conservative (low violations) at the cost of mean on some tasks, consistent with anchor-based regularization.
FQL often achieves high-mean points but with comparatively higher violation counts in the Pareto plot. Without tail-aware guidance, safety depends on the expected-value critic and task smoothness, explaining the variability across datasets.
DiffusionQL exhibits wider scatter: runs either reach moderate returns with elevated violations or collapse to low-return, near-zero violation regions. This variability is consistent with value-only guidance under stochastic penalties and matches its weaker CVaR.
CODAC clusters in the low-return/low-violation corner across tasks, as expected from conservative critics.
% =========================
\paragraph{Additional Qualitative Safety Analysis on \textsc{Medium--Expert}}
\label{app:safety_plots}
% =========================
Fig.~\ref{fig:safety_heatmaps} in the main text reports qualitative safety heatmaps on the
\textsc{Medium-Replay} datasets. Here we provide the corresponding plots on
\textsc{Medium-Expert} to verify that the same qualitative trends hold under a different data regime.
As in the main text, shaded bands indicate the risk-free operational ranges
(HalfCheetah: $v\le 10$; Walker2d: $|\theta|\le 0.5$; Hopper: $|\theta|\le 0.1$). Across tasks in Fig.~\ref{fig:safety_heatmaps_me}, \textbf{RADAC} concentrates probability mass
inside or near the risk-free boundary while preserving high-reward behavior supported by the dataset.
In contrast, \textbf{DiffusionQL} tends to under-explore near the boundary due to risk-blind bootstrapping under rare hazards, often yielding overly conservative action distributions.
Finally, \textbf{ORAAC} (prior-anchored perturbation) can allocate nontrivial mass near low-density regions induced by the anchored geometry, consistent with the leakage intuition in Sec.~\ref{sec:risk_pitfalls}.
Overall, the \textsc{Medium--Expert} results are consistent with the main-text conclusion.

\subsection{Ablation Study}
\label{app:ablation_study}
\subsubsection{Objective Ablation: Roles of BC and CVaR}
\label{app:objective_ablation}

To isolate the roles of behavior regularization and tail-risk optimization, 
we compare the full RADAC objective against BC-only and CVaR-only variants 
on two representative Stochastic-D4RL tasks. We additionally report the 
state-conditioned 1-NN OOD action rate used in Sec.~\ref{subsec:epsilon_act}. 
The results support a two-stage interpretation: BC substantially suppresses 
OOD leakage, while the CVaR term improves lower-tail performance within this 
low-OOD regime.

\begin{table}[t]
\centering
\small
\setlength{\tabcolsep}{5pt}
\renewcommand{\arraystretch}{1.08}
\caption{Objective ablation on representative Stochastic-D4RL tasks over 3 seeds. 
OOD denotes the state-conditioned 1-NN action rate (\%); smaller is better.}
\label{tab:objective_ablation}
\begin{tabular}{llccc}
\toprule
\textbf{Dataset} & \textbf{Variant} & \textbf{Mean} & \textbf{CVaR$_{0.1}$} & \textbf{OOD} \\
\midrule
\multirow{3}{*}{HalfCheetah-m-r}
& RADAC (full) & $518.7$ & $341.0$ & $2.08$ \\
& BC-only      & $485.9$ & $269.5$ & $2.42$ \\
& CVaR-only    & $2.9$   & $-28.3$ & $12.97$ \\
\midrule
\multirow{3}{*}{Walker2d-m-r}
& RADAC (full) & $585.9$ & $128.3$ & $0.49$ \\
& BC-only      & $617.3$ & $-95.6$ & $0.46$ \\
& CVaR-only    & $-14.4$ & $-53.0$ & $62.87$ \\
\bottomrule
\end{tabular}
\end{table}
\subsubsection{Choice of Risk Distortion}
\label{app:risk_ablation}
We evaluate RADAC and RAFMAC with three risk distortions CVaR, Wang, and CPW under the same 1000‑step evaluation protocol used above.
 Across seeds, Wang generally tilts updates toward higher means and weaker tails; CPW sits between CVaR and Wang but shows higher variance across seeds. Overall, CVaR is the most reliable choice for lower‑tail control at comparable mean. 
\begin{table}[t]
\centering
\scriptsize
\setlength{\tabcolsep}{3pt}
\caption{Ablation (1000-step evaluation). RADAC/RAFMAC with CVaR, Wang, and CPW on \textsc{HalfCheetah}-Medium-Replay and \textsc{Walker2d}-Medium-Replay. Scores are mean $\pm$ s.e.\ over 3 seeds.}
\label{tab:ablation_mr_radac_rafmac}
\begin{tabular}{llcccc}
\toprule
\multirow{2}{*}{Method} & \multirow{2}{*}{Distortion} & \multicolumn{2}{c}{HalfCheetah–m–r} & \multicolumn{2}{c}{Walker2d–m–r} \\
 & & Mean & CVaR$_{0.1}$ & Mean & CVaR$_{0.1}$ \\
\midrule
RADAC  & CVaR & $2758.5 \pm 84.1$  & $1759.5 \pm 71.5$   & $681.3 \pm 409.3$ & $-395.1 \pm 438.3$ \\
RADAC  & Wang & $2653.5 \pm 86.5$  & $310.8 \pm 92.6$    & $417.3 \pm 397.0$ & $-52.1 \pm 11.4$   \\
RADAC  & CPW  & $2777.9 \pm 93.7$  & $1061.6 \pm 731.7$  & $64.3 \pm 149.3$  & $-203.6 \pm 69.8$  \\
\midrule
RAFMAC & CVaR & $2835.8 \pm 116.3$ & $1981.2 \pm 405.3$  & $698.8 \pm 215.5$ & $5.6 \pm 60.8$     \\
RAFMAC & Wang & $2625.6 \pm 113.8$ & $462.5 \pm 427.6$   & $552.2 \pm 134.8$ & $-706.4 \pm 687.5$ \\
RAFMAC & CPW  & $2539.2 \pm 31.1$ & $95.9 \pm 92.3$   & $360.7 \pm 49.6$ & $-71.6 \pm 22.1$ \\
\bottomrule
\end{tabular}
\end{table}
\subsubsection{Effect of the Tail Sample Size $K$ and $N$ on the CVaR Estimator}
\label{app:cvar-k}

To clarify the role of the tail sample size $K$ in our IQN-based CVaR
estimator, we run an experiment on a subset of D4RL, 
\textsc{HalfCheetah-Medium-Replay-v2} and
\textsc{Walker2d-Medium-Replay-v2}.
During training we randomly sample $2{,}000$ states from the replay
buffer and store them as a fixed evaluation set
$\{s_j\}_{j=1}^{2000}$.  For each state $s_j$, each
$K\in\{2,4,8,16\}$, and each risk level
$\alpha\in\{0.05,0.1,0.2\}$ we estimate $\mathrm{CVaR}_\alpha$ using the offline-selected RADAC policy (trained with $\alpha_{\text{train}}=0.1$).

Given a state $s$, we draw $K$ i.i.d. pairs
$(a_k,\tau_k)$ with $a_k \sim \pi_\theta(\cdot \mid s)$ and
$\tau_k \sim \mathrm{Unif}(0,\alpha)$, evaluate both distributional
critics $Z_{\phi_1}, Z_{\phi_2}$, and form the tail-sampling CVaR
estimator
\begin{equation}
  \widehat{\mathrm{CVaR}}_\alpha(s)
  =
  \frac{1}{K}\sum_{k=1}^K
  \min_{i\in\{1,2\}} Z_{\phi_i}\bigl(s, a_k;\tau_k\bigr).
  \label{eq:app-cvar-tail-estimator}
\end{equation}
For each $(K,\alpha)$ we repeat this procedure $100$ times per state
(using new $(a_k,\tau_k)$ draws each time), compute the variance of
$\widehat{\mathrm{CVaR}}_\alpha(s_j)$ across the $100$ repetitions for
each state $s_j$, and then aggregate these per-state variances by
their mean and standard deviation over $j=1,\dots,2000$.

\begin{figure}[t]
  \centering
  \includegraphics[width=0.48\linewidth]{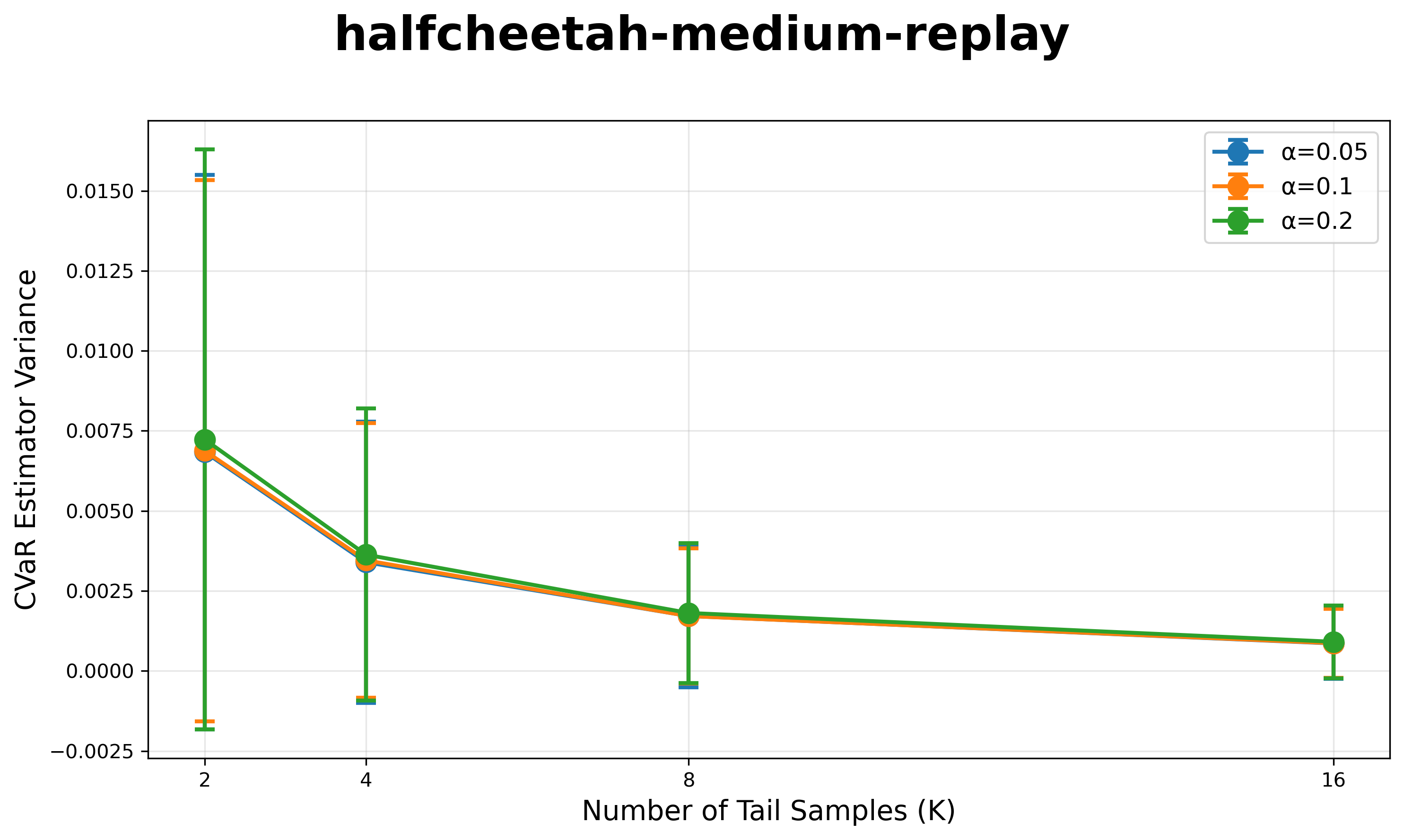}
  \hfill
  \includegraphics[width=0.48\linewidth]{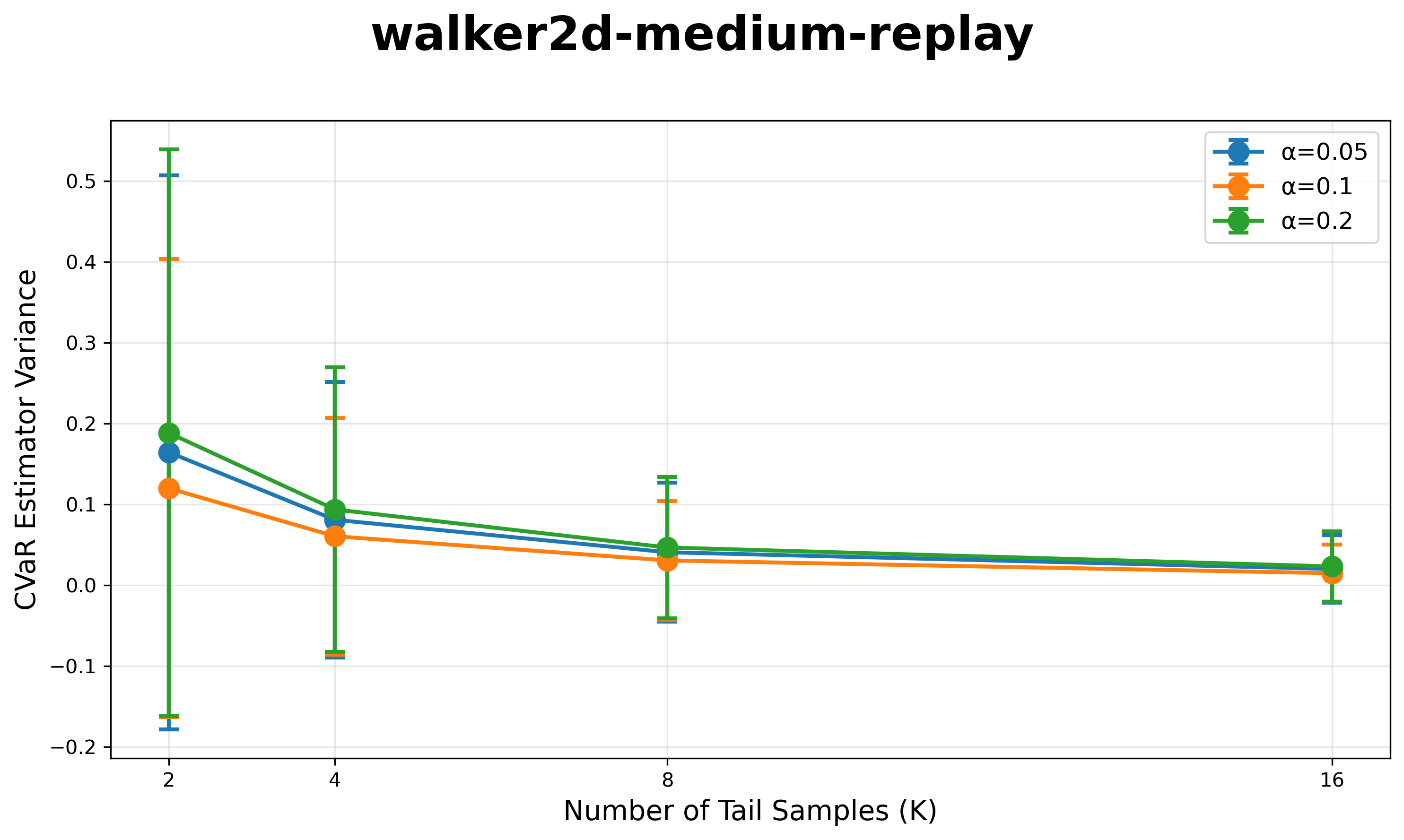}
  \caption{
    Empirical variance of the IQN-based $\mathrm{CVaR}_{0.1}$ estimator
    as a function of the tail sample size $K$ for the offline-selected
    RADAC policies on
    \textsc{HalfCheetah-Medium-Replay-v2} (left) and
    \textsc{Walker2d-Medium-Replay-v2} (right).
    For each $K\in\{2,4,8,16\}$ we compute
    $\widehat{\mathrm{CVaR}}_\alpha$; the curves show the
    mean per-state estimator variance with error bars indicating one
    standard deviation across states for
    $\alpha\in\{0.05,0.1,0.2\}$; all three risk levels exhibit a
    similar $1/K$-like decay.}
  \label{fig:app-cvar-variance-k}
\end{figure}

Figure~\ref{fig:app-cvar-variance-k} reports the resulting
\emph{estimator variance} as a function of $K$ for
$\alpha\in\{0.05,0.1,0.2\}$.
Across both tasks  and all three risk levels the variance decreases approximately at a $1/K$
rate: using only $K=2$ tail samples yields noisy CVaR estimates,
while increasing $K$ to $4$ and $8$ substantially reduces the
variance.  The marginal improvement from $K=8$ to $K=16$ is much
smaller, despite doubling the number of tail samples.
This supports our choice of a moderate tail size (with $K$ in the range $8$–$16$ across tasks in our experiments), which provides a good trade-off between estimator noise
and computational cost.
Note that the absolute variance scale differs between environments.
\paragraph{Role of the number of quantiles $N$.}
The tail estimator in Eq.~\ref{eq:app-cvar-tail-estimator} depends on the
number of tail samples $K$, while the IQN critic itself is trained with
$N$ quantile samples per state–action pair.  In all experiments we use
moderate values $N\in\{16,32\}$ (see Table~\ref{tab:ramac_key_hparams}), which are standard in
prior IQN-based work and make the critic updates stable.
Increasing $N$ primarily reduces the variance of the critic update and
smooths the learned value distribution, but once $N$ is in this
moderate range we do not observe qualitative changes in the CVaR
estimates or policy performance, while the computational cost grows
roughly linearly in $N$.  Hence we treat $N$ as a fixed architectural
hyperparameter.
\subsubsection{OOD Detector Robustness}
\label{app:ood_detectors}
To check that the RADAC $<$ ORAAC trend in Table~\ref{tab:ood_lb_compact} is not an artifact of the 1-NN score, we repeated the analysis with two additional detectors: a Local Outlier Factor (LOF) and a simple Mahalanobis detector based on a single Gaussian fit to the dataset joint $(s,a)$ points. Table~\ref{tab:ood_detectors_seed} reports the three Stochastic-D4RL medium-expert tasks with these detectors. On all datasets,  all three detectors agree with the main-text result and rank RADAC as having substantially lower OOD action rates than ORAAC variants, indicating that the RADAC $<$ ORAAC ordering is robust to the detector choice. Implementation details of these detectors are provided in App.~\ref{app:ood_measurement}

\begin{table}[t]
\centering
\caption{Detector–robust OOD action rates $\varepsilon_{\text{act}}$ (\%, mean $\pm$ std over 3 seeds) on Stochastic–D4RL medium–expert tasks. Smaller is better.}
\label{tab:ood_detectors_seed}
\begin{tabular}{l l c c c}
\toprule
Env & Algo & 1-NN (state-cond.)  & LOF $(s,a)$ & Mahalanobis $(s,a)$ \\
\midrule
HalfCheetah-m.e. & RADAC & 1.22 $\pm$ 0.21 & 0.37 $\pm$ 0.16 & 2.08 $\pm$ 0.62 \\
HalfCheetah-m.e. & ORAAC & 29.27 $\pm$ 12.76 & 13.33 $\pm$ 20.37 & 19.06 $\pm$ 19.57  \\
HalfCheetah-m.e. & ORAAC-Diff. & 35.44 $\pm$ 25.93 & 3.28 $\pm$ 0.883 & 30.32 $\pm$ 23.35\\
Hopper-m.e.      & RADAC &1.21 $\pm$ 0.21 & 0.38 $\pm$ 0.50 & 5.58 $\pm$ 0.26 \\
Hopper-m.e.      & ORAAC & 11.60 $\pm$ 3.95 & 36.66 $\pm$ 1.47 & 18.44 $\pm$ 11.41 \\
Hopper-m.e.      & ORAAC-Diff. & 22.75 $\pm$ 6.56 & 39.64 $\pm$ 12.43 & 15.19 $\pm$ 6.10 \\
Walker2d-m.e.    & RADAC & 0.96 $\pm$ 0.45 & 0.63 $\pm$ 0.26 & 3.12 $\pm$ 0.80 \\
Walker2d-m.e.    & ORAAC & 5.57 $\pm$ 3.31 & 6.99 $\pm$ 2.97 & 7.76 $\pm$ 1.92 \\
Walker2d-m.e.    & ORAAC-Diff. & 18.54 $\pm$ 2.71 & 16.22 $\pm$ 3.86 & 15.89 $\pm$ 3.19 \\
\bottomrule
\end{tabular}
\end{table}

\subsection{Runtime and Inference Latency}
\label{app:runtime}

We compare wall-clock training time and per-action inference latency for
DiffusionQL, FQL, and RADAC on \textsc{Hopper-Medium-Expert-v2}. All methods use the same A6000 GPU, PyTorch/CUDA stack, and training budget. Training time is measured between the first and last gradient update; latency is the average GPU time to produce a single action for
$10^4$ replay states with \texttt{torch.no\_grad}.

\begin{figure}[h]
  \centering
  \includegraphics[width=0.8\linewidth]{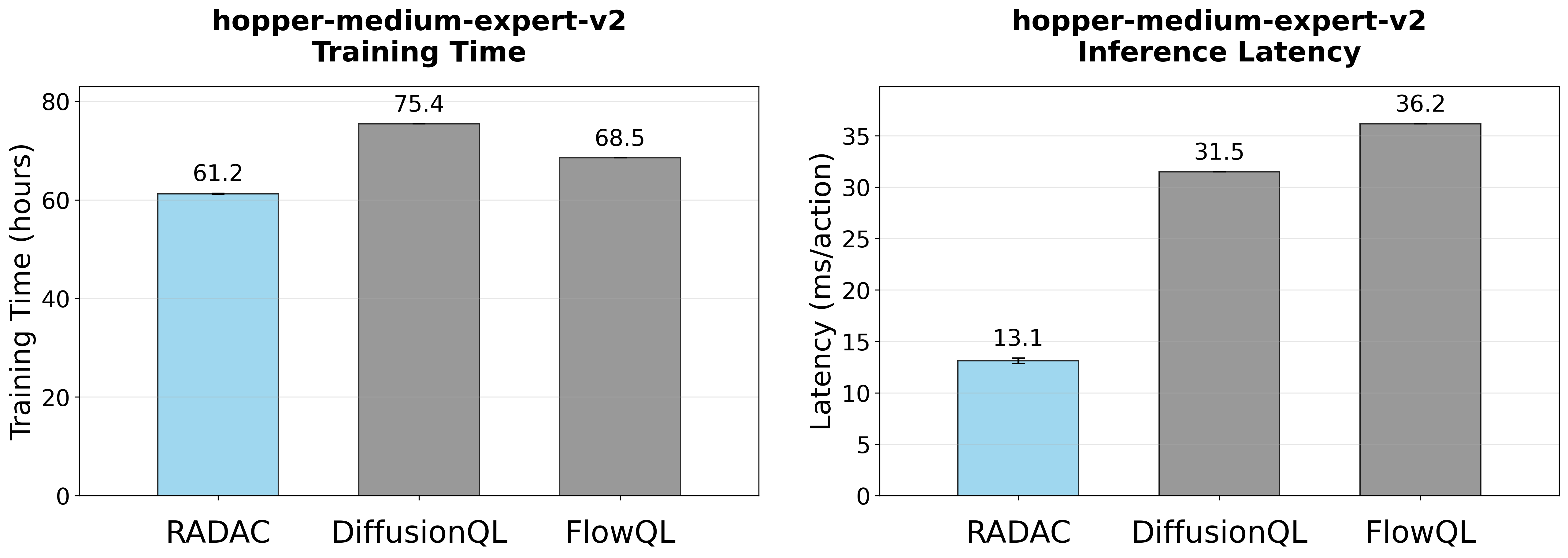}
  \caption{Wall-clock training time (left) and per-action inference
  latency (right) on \textsc{Hopper-Medium-Expert-v2}.}
  \label{fig:runtime_latency}
\end{figure}

RADAC replaces the scalar critic with a distributional IQN and optimizes
a CVaR objective, but this does not dominate runtime. In these methods,
most of the computational cost comes from the expressive generative
actor (diffusion or flow), not from the critic. The extra work required
by IQN, evaluating a small number of quantiles and averaging the bottom
$\alpha$-fraction for CVaR, adds only minor overhead relative to a full
diffusion / flow pass.

Conversely, FQL carries its own overhead by training both a
flow-matching prior and a distilled one-step policy head, and
the DiffusionQL implementation uses slightly heavier hyperparameters
(e.g., a larger actor) than RADAC. As a
result, under our implementation and hyperparameter choices, RADAC ends
up slightly faster in wall-clock time on
\textsc{Hopper-Medium-Expert-v2}. We do not claim that RADAC is
intrinsically faster than DiffusionQL or FQL; these measurements
simply show that the CVaR + distributional critic extension does
\emph{not} introduce an order-of-magnitude runtime penalty. Inference
latency is likewise dominated by the shared diffusion/flow backbone, and
the RADAC actor achieves comparable or lower per-action latency than the
risk-neutral expressive baselines when using similar numbers of
denoising / flow steps.

\subsection{Return Distributions of Rollout Trajectories}
\label{app:return-dists}

\begin{figure}[t]
  \centering
  \includegraphics[width=0.9\linewidth]{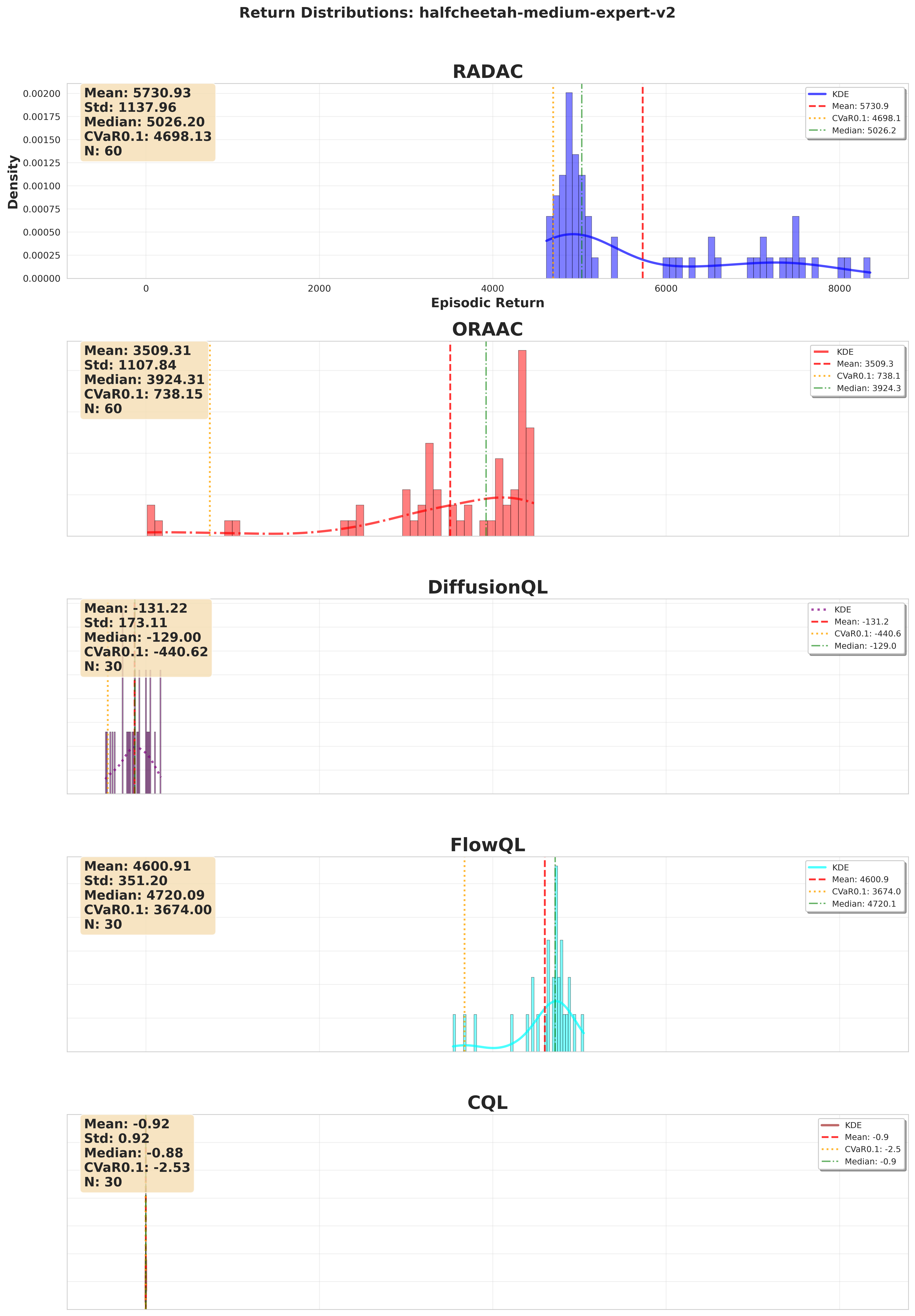}
  \caption{
  Empirical return distributions on
  \textsc{HalfCheetah-Medium-Expert-v2} under the stochastic hazard
  wrapper.
  Each subplot shows a histogram and KDE of episodic returns across
  evaluation rollouts, with vertical lines for the mean, median, and
  $\mathrm{CVaR}_{0.1}$.}
  \label{fig:return-dists}
\end{figure}

We visualize the empirical return distributions under the
stochastic hazard wrapper for \textsc{HalfCheetah-Medium-Expert-v2}.
For each method we aggregate $30$--$60$ evaluation episodes across three seeds and plot histograms and kernel-density estimates of the rollout returns, together with vertical markers for the mean, median, and $\mathrm{CVaR}_{0.1}$ (Fig.~\ref{fig:return-dists}). On \textsc{HalfCheetah-Medium-Expert-v2}, for example, RADAC
concentrates mass in a high-return band
($\mathrm{CVaR}_{0.1}\!\approx\! 4.7\!\times\!10^3$) while ORAAC
exhibits a multimodal distribution with occasional near-zero or
negative episodes, and DiffusionQL/CQL collapse near zero.
Since hazard events in our wrapper correspond to large negative penalties without early termination, catastrophic hazard-inducing episodes populate
the extreme left tail; the reduced tail mass for RADAC thus reflects a
lower frequency of hazard-heavy trajectories rather than mere
over-conservatism.

\subsection{Risk-Return Frontier under CVaR Level}
\label{app:risk_return_frontier}

\begin{figure}[t]
    \centering
    \includegraphics[width=0.48\linewidth]{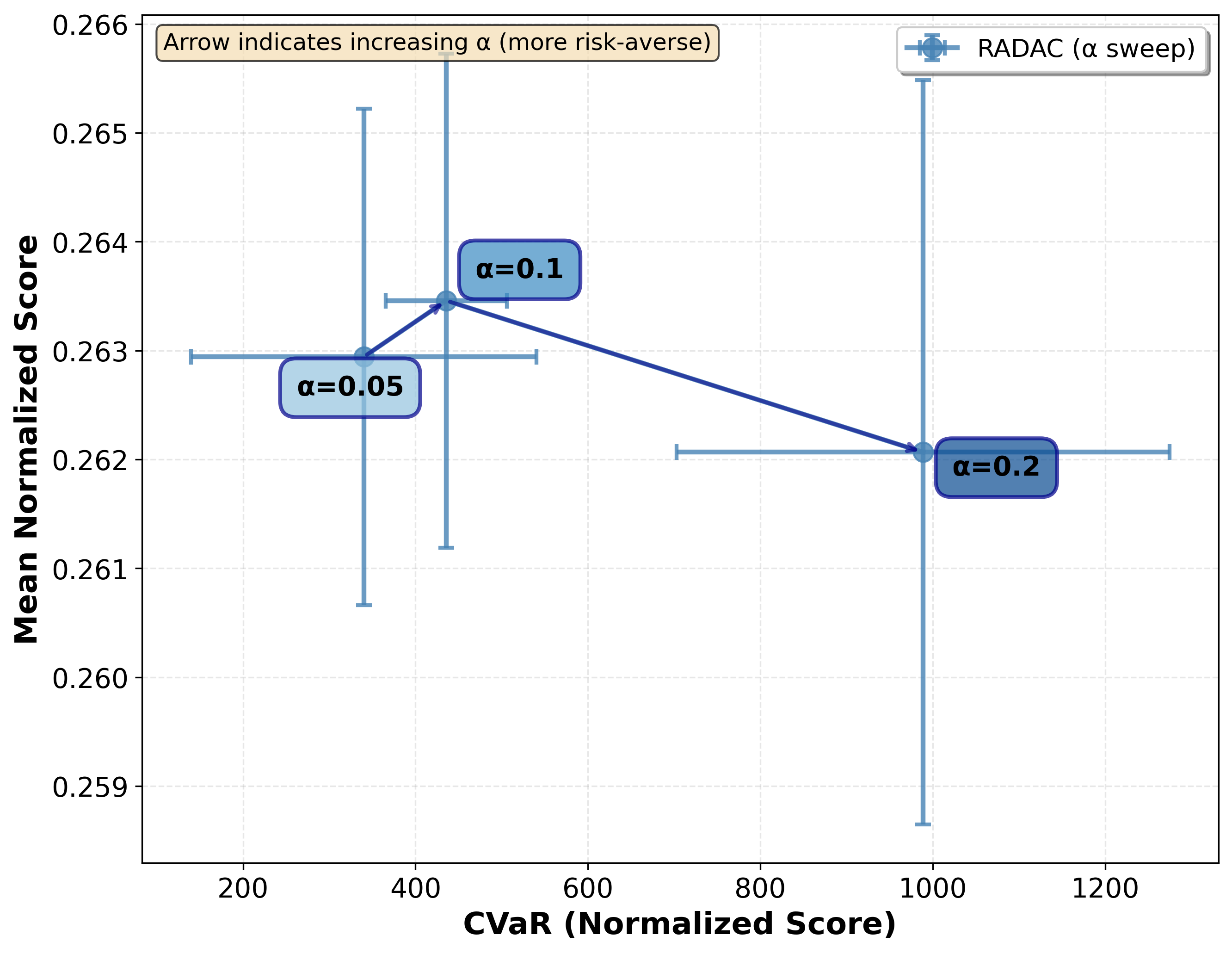}
    \includegraphics[width=0.48\linewidth]{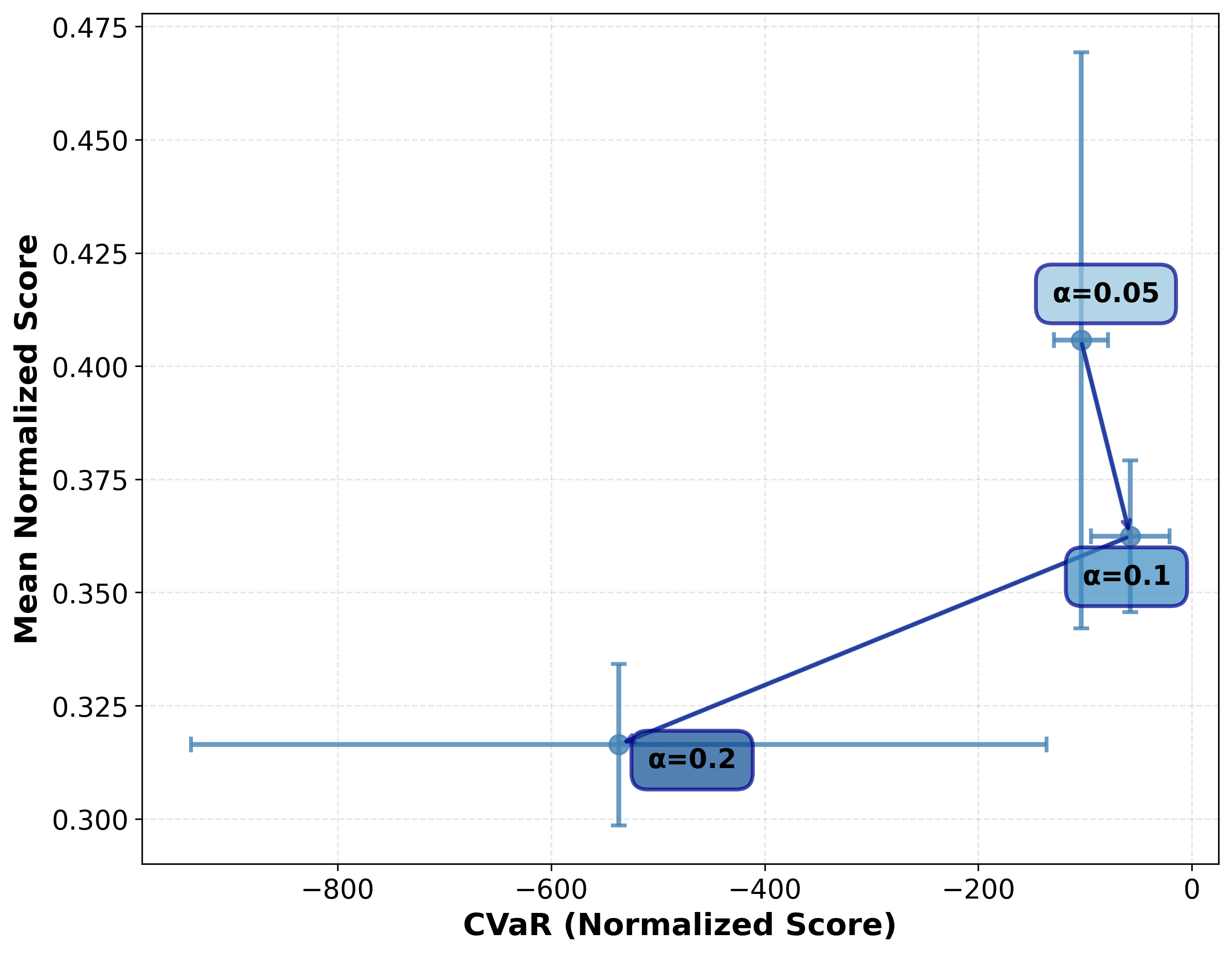}
    \caption{\textbf{Risk-return frontier for RADAC under different CVaR levels.}
    Each point shows the seed-averaged mean normalized score and
    $\mathrm{CVaR}_\alpha$ for RADAC trained with a fixed $\alpha \in
    \{0.05,0.10,0.20\}$ on \textsc{HalfCheetah-Medium-Replay-v2} (left) and
    \textsc{Walker2d-Medium-Replay-v2} (right). Error bars denote the standard
    error of the mean across seeds.}
    \label{fig:app_alpha_frontier}
\end{figure}

To provide a clearer view of the safety-return trade-off, we
visualize how the CVaR level $\alpha$ affects RADAC's behavior on \textsc{HalfCheetah-Medium-Replay-v2} and
\textsc{Walker2d-Medium-Replay-v2}. For each environment, we train RADAC with
$\alpha \in \{0.05, 0.10, 0.20\}$ using the same hyperparameters as in the main
Stochastic-D4RL experiments, and aggregate results across multiple seeds. For
each run, we select the checkpoint with the highest normalized score and compute
the mean normalized return and empirical $\mathrm{CVaR}_\alpha$ from eval
rollouts under the risky wrapper. We then report the seed-averaged mean
normalized score versus $\mathrm{CVaR}_\alpha$, with error bars denoting the
standard error of the mean across seeds
(Fig.~\ref{fig:app_alpha_frontier}).

On \textsc{HalfCheetah-Medium-Replay-v2}, varying $\alpha$ across
$\{0.05, 0.10, 0.20\}$ primarily affects the tail: the points move noticeably
along the horizontal axis in $\mathrm{CVaR}_\alpha$, while the mean normalized
score stays in a very tight band, indicating that RADAC can reshape the lower
tail of the return distribution with only a minor impact on average
performance. In contrast, on \textsc{Walker2d-Medium-Replay-v2} the frontier
moves up and to the right as we adjust $\alpha$: the more tail-sensitive
settings simultaneously improve $\mathrm{CVaR}_\alpha$ and the mean normalized
score, suggesting that on this task catastrophic low-return trajectories are
frequent enough that suppressing them not only reduces tail risk but also
raises average returns. Overall, these frontiers confirm that the BC+CVaR
objective in RADAC exposes a smooth knob to trade off tail-risk and mean
return, and that in some regimes increasing tail sensitivity can strictly
improve both safety and performance.

% =========================
% Appendix: Experimental Details
% =========================
\section{Experimental Details}
\label{app:experimental_details}

\subsection{2D Synthetic Task Details}
\label{app:synthetic_details}

\paragraph{Risky–Bandit dataset}
We generate $N=10^4$ state–action–reward tuples with dummy zero states. Actions come from two modes:
(i) Ring (80\%): radius $0.9\pm0.04$; base reward $\mathcal N(9,\,0.3^2)$; with probability $0.05$ a trap penalty $-40$ is applied (heavy lower tail).
(ii) Center (20\%): $\mathcal N(\mathbf{0},\,0.1^2\mathbf I)$; reward $\mathcal N(5,\,0.3^2)$.
Actions are clipped to $[-1,1]^2$.

All methods train on the same static dataset; when a BC regularizer is required we use the standard loss of the underlying generator. RADAC adds the CVaR term from Eq.~\ref{eq:policy_loss_full} to the diffusion/flow BC objective and backpropagates.
For each trained policy we draw $1{,}000$ action samples for visualization in Fig.~\ref{fig:toy_rl}. 
\subsection{Stochastic-D4RL MuJoCo Suite}
\label{app:sd4rl_details}

\textbf{Datasets} We adopt the \emph{stochastic MuJoCo} protocol for risk-sensitive offline RL, following  \citep{urpi2021risk}. Policies are evaluated on
\[
\{\textsc{Hopper},\ \textsc{Walker2d},\ \textsc{HalfCheetah}\}
\ \times\
\{\textsc{medium-expert},\ \textsc{medium-replay}\},
\]
Compared to prior work, we prefer \textsc{medium-expert} and \textsc{medium-replay} to validate both \emph{risk sensitivity} and \emph{policy expressiveness} under multimodal action distributions.
For training, we relabel per-transition rewards in the offline datasets to inject stochastic hazards
(velocity or torso-pitch thresholds with Bernoulli penalties and early termination);
\emph{the same hazard model is used at evaluation.}
All algorithms (CQL, CODAC, ORAAC, DiffusionQL/FQL, RADAC, etc.) are trained on these relabeled rewards; the hazard indicator is never provided as an input feature or mask, so no method receives privileged information about hazard locations.
This ensures the critic and the policy are trained on the risk-aware rewards rather than only being tested under hazards.

\textbf{Settings} Each task defines a monitored signal and an additive Bernoulli penalty when a safety condition is violated; pose-based tasks also include an early-termination threshold.
\begin{itemize}[leftmargin=*]
  \item \textbf{\textsc{HalfCheetah}}: monitor forward velocity. Apply a penalty with probability $p=0.05$ if the threshold is exceeded. Thresholds/penalties: \textsc{medium-expert}/\textsc{medium-replay} uses $v>10.0$/ $v>5.0$ with penalty $-70.0$. No early termination. Max episode steps: $200$.
  \item \textbf{\textsc{Hopper} / \textsc{Walker2d}}:
monitor torso pitch angle. When $|\theta|$ leaves the healthy range,
add a penalty with probability $p=0.10$; terminate early if
$|\theta|>2|\tilde{\theta}|$. Max episode steps: $500$.
  \begin{itemize}[leftmargin=1.25em]
    \item \textsc{Hopper}: healthy range $[-0.1,0.1]$ rad; penalty $-50.0$ when $|\tilde{\theta}|>0.1$; early termination if $|\theta|>0.2$.
    \item \textsc{Walker2d}: healthy range $[-0.5,0.5]$ rad; penalty $-30.0$ when $|\tilde{\theta}|>0.5$; early termination if $|\theta|>1.0$.
  \end{itemize}
\end{itemize}
% =========================
% Appendix: Baselines (Implementation & Hyperparameters)
% =========================
\subsection{Baselines: Implementation \& Hyperparameters}
\label{app:baselines}

We include six representative offline-RL baselines:
\begin{itemize}[leftmargin=*]
 \item \textbf{CQL}~\citep{kumar2020conservative} (value pessimism). Non-distributional conservative Q-learning baseline.
  \item \textbf{CODAC}~\citep{ma2021conservative} (distributional conservative learning). We primarily use the CVaR-optimizing specification (``CODAC-C'', $\mathrm{CVaR}_{0.1}$ objective). 
  % \textcolor{blue}{In all experiments, CODAC is configured with risk level $\alpha=0.1$, so it serves as a ``conservative + distributional CVaR'' baseline against which RAMAC is compared.}
  \item \textbf{ORAAC}~\citep{urpi2021risk} (offline risk-averse actor–critic). A distributional critic with imitation-regularized policy optimizing a coherent risk objective.
  % \textcolor{blue}{A hypothetical ``CQL--CVaR'' variant would combine CQL's conservative penalty with the same type of distributional CVaR head used in CODAC, making it conceptually very close to CODAC; we therefore treat CODAC as the representative conservative+CVaR baseline and do not report a separate CQL--CVaR instantiation.}
  \item \textbf{DiffusionQL}~\citep{wang2022diffusion} (expressive risk-neutral diffusion policy).
  \item \textbf{FQL}~\citep{park2025flow} (expressive risk-neutral flow-matching policy).
  \item \textbf{ORAAC-Diffusion (ORAAC-Diff.)}~\citep{chen2025diffusion} (offline risk-averse prior-anchored perturbation with a diffusion behavior prior)
\end{itemize}

\paragraph{Hyperparameter selection \& tuning}
For each of baselines, we run all baselines ourselves and tune the following parameters or adopt authors’ recommended settings, mirroring the practice in \citep{kumar2020conservative, ma2021conservative,park2025flow,urpi2021risk,wang2022diffusion}. 
\begin{itemize}[leftmargin=*]
  \item \textbf{FQL}~\citep{park2025flow}: we sweep the policy weight $\alpha \in \{1,10,30,100,1000\}$ per task and report the best-performing setting (selection by $\mathrm{CVaR}_{0.1}$ unless noted).
  \item \textbf{ORAAC-Diff.}~\citep{chen2025diffusion}: we tune the mixing coefficient 
$\lambda \in \{0.1,0.2,0.4,0.6\}$ per task and report the best-performing setting, following the role of $\lambda$ as the main environment-dependent trade-off coefficient in the original method.
  \item \textbf{DiffusionQL}~\citep{wang2022diffusion}: we consider $\eta \in \{0.1, 0.5,1.0\}$ for the BC coefficient. We use authors’ recommended configuration for other parameters without retuning. We also used the best checkpoint of their model on each benchmark by following their protocol.
  % \item \textbf{ORAAC}~\citep{urpi2021risk}: use the paper’s recommended configuration (distributional critic, risk level $\alpha=0.1$, anchor/prior regularization) without additional sweeps.
  \item \textbf{ORAAC}~\citep{urpi2021risk}: use the paper's recommended configuration, including its distributional critic,
risk level $\alpha=0.1$, and anchor/prior regularization setting.
  \item \textbf{CODAC}~\citep{ma2021conservative}: use the paper’s tuned settings for D4RL (risk level $\alpha=0.1$) without further tuning.
  \item \textbf{CQL}~\citep{kumar2020conservative}: use the standard conservative coefficient and implementation defaults for MuJoCo locomotion.
\end{itemize}

\subsection{Estimating OOD Action Rates and Detectors}
\label{app:ood_measurement}

At evaluation time we measure the fraction of evaluation \emph{state--action} pairs produced by a policy that fall outside the empirical support of the offline dataset. Let $\mathcal{D}=\{(s_i,a_i)\}_{i=1}^{N}$ denote the offline dataset for a given task, and let $\{(s_t^{(\mathrm{eval})},a_t^{(\mathrm{eval})})\}_{t=1}^{T}$ be all state--action pairs emitted across evaluation rollouts.

Actions are scaled to $[-1,1]$ per dimension in MuJoCo, so distances in action space are computed directly in $\ell_2$. For detectors operating on joint $(s,a)$ features, we standardize each \emph{state} dimension using dataset statistics (z-scoring), while actions remain in their native $[-1,1]$ scale.

Given a detector that assigns an OOD indicator
\[
\mathbf{1}_{\mathrm{OOD}}\!\bigl(s_t^{(\mathrm{eval})},a_t^{(\mathrm{eval})}\bigr)\in\{0,1\},
\]
we define the OOD rate
\[
\varepsilon_{\mathrm{act}}
\;=\;
\frac{1}{T}\sum_{t=1}^{T}\mathbf{1}_{\mathrm{OOD}}\!\bigl(s_t^{(\mathrm{eval})},a_t^{(\mathrm{eval})}\bigr),
\]
and report the mean and standard deviation over seeds (mean $\pm$ std), matching Table~\ref{tab:ood_lb_compact} and Table~\ref{tab:ood_detectors_seed}.

\paragraph{State-conditioned 1-NN detector.}
We use a local notion of action support conditioned on the evaluation state. For an evaluation state $s$, define the $k$-NN state neighborhood in the dataset
$\mathcal{N}_k(s)\subset\{s_i\}_{i=1}^{N}$ (we use $k{=}10$), and the corresponding local action set
\[
\mathcal{A}_{\mathcal{D}}(s) \;=\; \{a_i \;|\; s_i \in \mathcal{N}_k(s)\}.
\]
For each evaluation pair $(s_t^{(\mathrm{eval})},a_t^{(\mathrm{eval})})$, we compute the local 1-NN action distance
\[
d_t^{(\mathrm{eval})} \;=\; \min_{a_i \in \mathcal{A}_{\mathcal{D}}(s_t^{(\mathrm{eval})})} \|a_t^{(\mathrm{eval})}-a_i\|_2.
\]
We set the threshold
$\tau=\kappa\cdot \mathrm{median}\{d_i\}$ with $\kappa{=}3$,
where each dataset reference distance $d_i$ is computed in the same
manner using a leave-one-out construction that excludes the query
transition itself from the candidate neighbor set. The OOD indicator is
\[
\mathbf{1}_{\mathrm{OOD}}\!\bigl(s_t^{(\mathrm{eval})},a_t^{(\mathrm{eval})}\bigr)
=\mathbb{I}\!\left\{d_t^{(\mathrm{eval})}>\tau\right\}.
\]
This state-conditioned 1-NN rate is the quantity reported in Sec.~\ref{subsec:epsilon_act}.
Nearest-neighbor queries are implemented via a KD--tree over states to retrieve $\mathcal{N}_k(s)$, followed by a local 1-NN query in action space over $\mathcal{A}_{\mathcal{D}}(s)$.

\paragraph{Alternative detectors (joint $(s,a)$ space).}
To check that the RADAC $<$ ORAAC trend is not an artifact of the 1-NN score, we also evaluate two additional detectors on the joint $(s,a)$ space. We concatenate standardized state features with raw actions and fit the detectors on $\{(s_i,a_i)\}_{i=1}^{N}$, then evaluate on $\{(s_t^{(\mathrm{eval})},a_t^{(\mathrm{eval})})\}_{t=1}^{T}$:

\begin{itemize}[leftmargin=*]
\item \textbf{Local Outlier Factor (LOF).}
We use LOF with $20$ neighbors and contamination level $0.01$ in the joint $(s,a)$ space, and flag an evaluation point as OOD when LOF predicts an outlier label.

\item \textbf{Single-Gaussian Mahalanobis distance.}
We fit a single multivariate Gaussian $\mathcal{N}(\mu,\Sigma)$ to the joint $(s,a)$ data (with a small diagonal jitter added to $\Sigma$ for numerical stability). For any point $x=[s;a]$ we compute the squared Mahalanobis distance
\[
d^2_{\mathrm{Mah}}(x) \;=\; (x-\mu)^\top \Sigma^{-1} (x-\mu).
\]
We set the OOD threshold $\tau$ as the upper $95\%$ quantile of $\{d^2_{\mathrm{Mah}}(x_i)\}_{i=1}^{N}$ estimated from up to $5\times 10^{4}$ subsampled dataset points, and declare $(s_t^{(\mathrm{eval})},a_t^{(\mathrm{eval})})$ OOD if $d^2_{\mathrm{Mah}}(x_t^{(\mathrm{eval})})>\tau$.
\end{itemize}

\begin{table}[t]
  \centering
  \small
  \setlength{\tabcolsep}{6pt}
  \renewcommand{\arraystretch}{1.08}
  \caption{\textbf{RAMAC: hyperparameters.}
  We keep only the knobs that materially affect performance and stability. Values are our defaults; brackets show typical sweep ranges.}
  \label{tab:ramac_key_hparams}
  \begin{tabular}{ll}
    \toprule
    \multicolumn{2}{l}{\textbf{Global}}\\
    \midrule
    Discount $\gamma$ & $0.99$ \\
    Batch size $B$ & $256$ \\
    Target update $\tau_{\text{target}}$ & $0.005$ \\
    Risk level $\alpha$ & $0.1$ \\
    \midrule
    \multicolumn{2}{l}{\textbf{Critic (Deterministic IQN)}}\\
    \midrule
    \#Quantiles $N$ & $32$ \\
    Grid $\mathcal T_N$ & $\{(i-\tfrac12)/N\}_{i=1}^{N}$ (fixed) \\
    Embedding dim & $128$ \\
    Critic LR & $3\times 10^{-4}$ \\
    Huber $\kappa$ & $1$ (fixed) \\
    Double IQN & enabled \\
    \midrule
    \multicolumn{2}{l}{\textbf{Actor (shared)}}\\
    \midrule
    Actor LR & $3\times 10^{-4}$ \\
    BC weight $\lambda_{\mathrm{BC}}$ & $1.0$ \\
    Risk weight $\eta$ & \textbf{RADAC}: $0.05$ \; [$0.02$–$0.1$], \quad \textbf{RAFMAC}: $1000$ \; [$100$–$1000$] \\
    Double critic clipping  & \textbf{RADAC}:$[-150, 150]$ or $[-300, 300]$, \quad \textbf{RAFMAC}:$[-300, 300]$ \\
    \midrule
    \multicolumn{2}{l}{\textbf{RADAC-specific}}\\
    \midrule
    Reverse diffusion steps $T$ & $5$ (VP schedule) \\
    \midrule
    \multicolumn{2}{l}{\textbf{RAFMAC-specific}}\\
    \midrule
    Flow steps $K$ & $10$ (Euler, $\Delta t{=}1/K$) \\
    \bottomrule
  \end{tabular}
\end{table}
%%%%%%%%%%%%%%%%%%%%%%%%%%%%%%%%%%%%%%%%%%%%%%%%%%%%%%%%%%%%%%%%%%%%%%%%%%%%%%%
%%%%%%%%%%%%%%%%%%%%%%%%%%%%%%%%%%%%%%%%%%%%%%%%%%%%%%%%%%%%%%%%%%%%%%%%%%%%%%%

\end{document}